\documentclass{article}
\usepackage[final]{neurips_2024}
\usepackage[utf8]{inputenc} 
\usepackage[T1]{fontenc}   
\usepackage{hyperref}      
\usepackage{url}            
\usepackage{booktabs}      
\usepackage{amsfonts}      
\usepackage{nicefrac}       
\usepackage{microtype}      
\usepackage{xcolor}         
\usepackage{amsmath}
\usepackage{multirow}
\usepackage{subcaption}
\usepackage{float}
\usepackage{booktabs}      
\usepackage{subcaption}    
\usepackage{xcolor}        
\usepackage{graphicx}      
\usepackage{caption}       
\usepackage{graphicx}
\usepackage{natbib}           
\usepackage{cleveref}         
\usepackage{amsthm}         
\theoremstyle{plain}
\newtheorem{theorem}{Theorem}
\newtheorem{lemma}{Lemma}
\newtheorem{proposition}{Proposition}
\theoremstyle{remark}

\newtheorem*{theorem*}{Theorem}
\newtheorem*{lemma*}{Lemma}
\newtheorem*{proposition*}{Proposition}
\newcommand{\plim}{\operatornamewithlimits{plim}}

\title{A Projection-Based ARIMA Framework for Nonlinear Dynamics in Macroeconomic and Financial Time Series: Closed-Form Estimation and Rolling-Window Inference}

\author{%
  Haojie Liu \\
  Department of Economics \\
  University of California, Riverside \\
  \texttt{hliu332@ucr.edu}
  \And
  Zihan Lin \\
  Department of Economics \\
  University of California, Riverside \\
  \texttt{zlin169@ucr.edu} 
}

\begin{document}

\maketitle

\begin{abstract}
We introduce Galerkin–ARIMA and Galerkin–SARIMA, a projection-based extension of classical ARIMA/SARIMA that replaces rigid linear lag operators with low-dimensional Galerkin basis expansions while preserving the familiar AR–MA decomposition. Experiments on synthetic series and on quarterly GDP and daily S\&P 500 returns show that Galerkin–SARIMA matches or improves forecast accuracy relative to classical ARIMA/SARIMA. Estimation is closed-form via a two-stage least-squares procedure, and the closed-form two-stage estimator enables efficient rolling-window re-estimation while preserving the familiar AR–MA operator structure, facilitating applications in central bank forecasting and portfolio risk management. We establish approximation–estimation trade-offs under weak dependence, provide consistency and asymptotic distributional results for the unpenalized estimator, compare prediction risk to classical SARIMA, and propose information-criterion selection of basis size. We further develop bootstrap-based inference for exogenous factor blocks and block-bootstrap prediction intervals that account for serial dependence and the two-stage generated-regressor structure.
\end{abstract}

\section{Introduction}

Time series forecasting is a core component of modern decision making in economics, finance, energy systems, and many areas of engineering and the natural sciences. Central banks monitor aggregate output, inflation, and unemployment through forecasts that inform interest rate policy; macroprudential authorities track credit and asset-price cycles to gauge systemic risk; portfolio managers and high frequency traders rely on short horizon return and volatility forecasts to allocate risk and manage execution; system operators in electricity markets and logistics use demand and load forecasts to plan capacity and hedge operational risk. In all of these settings, practitioners are confronted with long histories of noisy, structured time series and must transform past observations into one step ahead predictions that can be embedded in downstream rules or optimization procedures. Accurate one-step-ahead forecasts of GDP growth and inflation remain central to monetary policy rules (e.g., Taylor rule implementations), yet classical SARIMA often fails to capture nonlinear mean reversion documented in postwar U.S. data.

Despite the rapid development of machine learning architectures for sequence modeling, classical parametric time series models remain widely used in both academic work and practice. The Autoregressive Integrated Moving Average (ARIMA) family, in particular, is still a default baseline in many domains because it provides an interpretable decomposition of dynamics into autoregressive (AR), differencing, and moving average (MA) components, and because its statistical properties have been extensively studied since the Box Jenkins framework \citep{BoxJenkins1970,BrockwellDavis1991}. Seasonal variants (SARIMA) extend this structure to periodic data and are built into mainstream statistical software and forecasting toolkits. As a result, ARIMA and SARIMA models continue to serve as workhorses for small to medium scale forecasting problems in macroeconomics, financial econometrics, retail demand, and environmental time series.

At the same time, the way time series models are deployed has changed. Many applications now operate in data rich and latency sensitive regimes. In algorithmic trading or online recommendation systems, for example, thousands of rolling one step ahead forecasts may be required per second, often with a sliding calibration window to adapt to evolving dynamics. In large scale forecasting systems, a separate model may be maintained for every asset, customer, or spatial location, leading to hundreds or thousands of parallel univariate models that must be refit as new data arrive. In such environments, the computational overhead of refitting ARIMA models by maximum likelihood or conditional least squares at every step becomes a bottleneck: each refit requires solving a nonlinear optimization problem over AR and MA parameters, and the cost grows with the length of the estimation window and the complexity of the seasonal structure.

Beyond computational cost, classical ARIMA models impose a rigid linear relationship between the current differenced value and its past lags. This linear structure is convenient for estimation and interpretation, but can be overly restrictive when the underlying dynamics are nonlinear or involve interactions among lags. Empirical studies in economics and finance routinely document nonlinear mean reversion, threshold effects, and regime dependent dynamics that are not easily captured by low order linear AR terms. To address this, a large body of work has explored nonlinear autoregressive models, spline based ARIMA variants, and neural architectures that replace the linear AR part with richer function classes. These approaches can improve accuracy in complex settings, but often sacrifice some of the simplicity, closed form structure, or computational efficiency that make ARIMA attractive in the first place.

In parallel, Galerkin projection methods have become a standard tool for approximating complex functions and dynamical systems in numerical analysis, particularly for partial and stochastic differential equations. The basic idea is to approximate an unknown function, such as a drift or solution field, by a finite expansion in a chosen basis, and to determine the coefficients by enforcing an orthogonality condition of the residual against the basis functions. This reduces an infinite dimensional problem to a finite dimensional linear system while preserving key structural properties of the original dynamics. Galerkin methods are widely used in PDE solvers, reduced order modeling, and more recently in stochastic differential equation driven statistical models and diffusion based sampling \citep[e.g.][]{Dokuchaev2022GalerkinOption,Graham2023ManifoldMCMC}.

These developments suggest an opportunity at the intersection of classical time series modeling and Galerkin style approximation. On the one hand, ARIMA and SARIMA provide a well understood operator structure in terms of backshift polynomials for nonseasonal and seasonal AR and MA components. On the other hand, Galerkin projection with suitable bases (polynomials or splines in the lag space) offers a principled way to approximate nonlinear mappings from lag vectors or residual histories into future values, while retaining a linear parameterization that can be estimated by ordinary least squares. Bridging these two viewpoints raises a natural question: Can we preserve the familiar ARIMA/SARIMA operator structure, but replace the fixed linear AR component by a Galerkin style basis expansion that (i) flexibly captures nonlinear dependencies in lags, and (ii) admits a closed form, numerically stable estimation procedure suitable for fast rolling forecasts?

Answering this question is particularly relevant for applications where both interpretability and speed matter. Practitioners may be reluctant to replace ARIMA entirely with black box sequence models, but they still face nonlinear dynamics and tight computational budgets. A framework that keeps the moving average structure and Gaussian innovation assumptions of ARIMA, while upgrading the autoregressive component via basis expansions solved in closed form, would provide a middle ground between classical linear models and fully nonparametric approaches. The Galerkin ARIMA framework developed in this paper is motivated by this regime: data rich, potentially nonlinear time series where we still want ARIMA style structure, but need more flexibility and much faster refitting than standard maximum likelihood procedures can offer.

In summary, this paper makes the following contributions.

First, we introduce the Galerkin ARIMA and Galerkin SARIMA frameworks, which retain the usual backshift operator structure of ARIMA and SARIMA but replace the fixed linear autoregressive component by a basis expansion in the lag space. The nonseasonal and seasonal AR parts, as well as the MA corrections, are modeled as Galerkin projections onto polynomial or spline bases and can be estimated by a simple two stage least squares procedure. This yields closed form estimators that are easy to implement and naturally suited to rolling one step ahead forecasting.

Second, we provide a theoretical analysis of the unpenalized Galerkin SARIMA estimator. Under a Gaussian SARIMA data generating process and standard smoothness and design assumptions, we derive Jackson type approximation bounds for the basis expansions, establish consistency and asymptotic normality of the estimated coefficients, and give mean squared error rates for one step ahead forecasts as the basis size grows with the sample size. We also compare the computational complexity of our two stage estimator to classical maximum likelihood estimation for SARIMA models, showing that when the effective basis dimension is moderate, the Galerkin approach is asymptotically much cheaper to refit in rolling settings.

Third, we investigate the practical performance of the method and a ridge regularized extension. On four synthetic data generating processes that include linear ARMA dynamics, seasonal oscillations, trend plus autoregression, and nonlinear recursions, Galerkin SARIMA achieves forecast accuracy comparable to or better than classical SARIMA while delivering speedups of several orders of magnitude in rolling one step ahead prediction. On real macroeconomic and financial series (quarterly GDP and daily S\&P 500 returns), the method matches ARIMA level accuracy and the ridge variant substantially reduces occasional forecast spikes that arise with richer bases, improving numerical stability without sacrificing efficiency.

\section{Background and Related Work}

Modern time series forecasting sits on top of two mature strands of methodology that we build on in this paper. The first is the classical ARIMA and SARIMA framework, which represents temporal dependence through backshift polynomials in nonseasonal and seasonal lags and is typically estimated by Gaussian likelihood or conditional least squares. The second is a collection of approximation tools that replace fixed linear structure by richer function classes, ranging from spline and basis expansions to neural architectures and Galerkin type projections for dynamical systems. In this section we recall the ARIMA and SARIMA formulations that will serve as reference models, summarize the Galerkin projection idea in the finite dimensional setting, and situate our approach within existing nonlinear and hybrid extensions of ARIMA and related projection based methods.

\subsection{Classical ARIMA and SARIMA}
We briefly recall the classical ARIMA and SARIMA formulations that will serve as reference models. Let $\{y_t\}_{t\in\mathbb{Z}}$ be a univariate time series. The nonseasonal differencing operator of order $d\ge 0$ is defined by
\[
\Delta^d y_t = (1 - B)^d y_t,
\]
where $B$ is the backshift operator, $B y_t = y_{t-1}$. When $d>0$ the process $\{y_t\}$ is transformed into a differenced series $y_t^{(d)} = \Delta^d y_t$ that is assumed to be covariance stationary.

An $\mathrm{ARIMA}(p,d,q)$ model is an $\mathrm{ARMA}(p,q)$ model for the differenced series $\{y_t^{(d)}\}$. Writing
\[
\Phi(B) = 1 - \phi_1 B - \cdots - \phi_p B^p,
\qquad
\Theta(B) = 1 + \theta_1 B + \cdots + \theta_q B^q,
\]
and modeling the innovations as $\epsilon_t \sim \mathcal{N}(0,\sigma^2)$, the nonseasonal ARIMA model can be written as
\begin{equation}
  \Phi(B)\, y_t^{(d)} = \Theta(B)\, \epsilon_t,
  \qquad
  \epsilon_t \stackrel{\text{i.i.d.}}{\sim} \mathcal{N}(0,\sigma^2).
  \label{eq:arima_classical}
\end{equation}
Equivalently, $y_t^{(d)}$ is represented as a linear combination of its last $p$ values and the last $q$ innovations, with coefficients determined by the polynomials $\Phi$ and $\Theta$. Under the usual invertibility and stationarity conditions on $\Phi$ and $\Theta$, the model admits a well defined moving average representation and can be used to compute one step ahead forecasts through the associated linear predictor \citep{BoxJenkins1970,BrockwellDavis1991}.

Seasonal structure with period $m$ is incorporated in the SARIMA$(p,d,q)\times(P,D,Q)_m$ family. In addition to the nonseasonal differencing $\Delta^d$, a seasonal differencing operator of order $D\ge 0$ is defined by
\[
\Delta_m^D y_t = (1 - B^m)^D y_t,
\]
and the combined differenced series is
\[
y_t^{(d,D)} = (1 - B)^d (1 - B^m)^D y_t.
\]
Seasonal autoregressive and moving average polynomials are written as
\[
\Phi_s(B^m) = 1 - \Phi_1 B^m - \cdots - \Phi_P B^{Pm},
\qquad
\Theta_s(B^m) = 1 + \Theta_1 B^m + \cdots + \Theta_Q B^{Qm}.
\]
The full SARIMA model takes the compact operator form
\begin{equation}
  \Phi(B)\, \Phi_s(B^m)\, y_t^{(d,D)}
  =
  \Theta(B)\, \Theta_s(B^m)\, \epsilon_t,
  \qquad
  \epsilon_t \stackrel{\text{i.i.d.}}{\sim} \mathcal{N}(0,\sigma^2),
  \label{eq:sarima_classical}
\end{equation}
which simultaneously encodes nonseasonal and seasonal autoregressive and moving average effects.

Parameter estimation in ARIMA and SARIMA models is typically carried out by Gaussian maximum likelihood or by conditional least squares, which coincides with maximum likelihood under normality \citep{BrockwellDavis1991}. For a given parameter vector
\[
\Psi = (\phi_1,\dots,\phi_p,\;\theta_1,\dots,\theta_q,\;
         \Phi_1,\dots,\Phi_P,\;\Theta_1,\dots,\Theta_Q,\;\sigma^2),
\]
the innovations $\epsilon_t(\Psi)$ can be computed recursively from \eqref{eq:sarima_classical} given suitable initial conditions. The conditional log likelihood based on $N$ observations of the differenced series has the form
\begin{equation}
  \ell(\Psi)
  = -\frac{N}{2} \log \sigma^2
    - \frac{1}{2\sigma^2} \sum_{t=1}^N \epsilon_t(\Psi)^2
    + \text{constant},
  \label{eq:gaussian_loglik}
\end{equation}
and the estimator $\hat{\Psi}$ is obtained by maximizing $\ell(\Psi)$, or equivalently by minimizing the sum of squared prediction errors $\sum_t \epsilon_t(\Psi)^2$. In practice, this optimization is performed numerically using methods such as Newton iterations or quasi Newton schemes, and must often be repeated when the model is refit on rolling windows. In what follows we will view \eqref{eq:sarima_classical} as the reference data generating model and \eqref{eq:gaussian_loglik} as the baseline estimation approach against which our Galerkin based procedure is compared.

\subsection{Galerkin Projection for Function Approximation}

  The Galerkin method originates in numerical analysis as a way to approximate unknown functions or operators by restricting attention to a finite dimensional subspace and enforcing an orthogonality condition on the approximation error. We recall the basic idea in a form that will later be applied to conditional mean and innovation maps in time series models.

Let $(\Omega,\mathcal{F},\mu)$ be a probability space and consider the Hilbert space $L^2(\Omega,\mu)$ with inner product
\[
\langle f,g\rangle
\;=\;
\int_{\Omega} f(x)\,g(x)\,d\mu(x),
\qquad f,g\in L^2(\Omega,\mu),
\]
and associated norm $\|f\|_2 = \langle f,f\rangle^{1/2}$. Suppose we wish to approximate a square integrable function $b:\Omega\to\mathbb{R}$ by elements of a $K$ dimensional subspace
\[
V_K
=
\operatorname{span}\{\phi_1,\dots,\phi_K\}
\subset L^2(\Omega,\mu),
\]
where $\{\phi_j\}_{j=1}^K$ are fixed basis functions. Any candidate approximant $b_K\in V_K$ can be written as
\[
b_K(x)
=
\sum_{j=1}^K \theta_j\,\phi_j(x),
\qquad \theta = (\theta_1,\dots,\theta_K)^{\top}\in\mathbb{R}^K.
\]

The Galerkin approximation of $b$ in $V_K$ is defined as the orthogonal projection of $b$ onto $V_K$ with respect to the inner product $\langle\cdot,\cdot\rangle$. Equivalently, the residual
\[
r(x) = b(x) - b_K(x)
\]
is required to be orthogonal to every basis function in $V_K$:
\begin{equation}
\langle r,\phi_i\rangle
=
\int_{\Omega}
\bigl(b(x) - \sum_{j=1}^K \theta_j \phi_j(x)\bigr)\,\phi_i(x)\,d\mu(x)
=
0,
\qquad i=1,\dots,K.
\label{eq:galerkin_orthogonality}
\end{equation}
Defining the Gram matrix $G\in\mathbb{R}^{K\times K}$ and load vector $f\in\mathbb{R}^K$ by
\[
G_{ij}
=
\langle \phi_j,\phi_i\rangle
=
\int_{\Omega} \phi_j(x)\,\phi_i(x)\,d\mu(x),
\qquad
f_i
=
\langle b,\phi_i\rangle
=
\int_{\Omega} b(x)\,\phi_i(x)\,d\mu(x),
\]
the conditions in \eqref{eq:galerkin_orthogonality} can be written compactly as the linear system
\begin{equation}
G\,\theta
=
f.
\label{eq:galerkin_linear_system}
\end{equation}
When the basis functions are linearly independent and $G$ is positive definite, this system has a unique solution and $b_K$ is the $L^2(\mu)$ orthogonal projection of $b$ onto $V_K$. In particular, $b_K$ minimizes the squared error $\|b - v\|_2^2$ over all $v \in V_K$.

In many applications the measure $\mu$ is not known in closed form and the integrals defining $G$ and $f$ are replaced by empirical averages over observed samples $x_1,\dots,x_N$. If $y_t$ denotes a noisy observation of $b(x_t)$, the empirical inner product
\[
\langle f,g\rangle_N
=
\frac{1}{N}\sum_{t=1}^N f(x_t)\,g(x_t)
\]
leads to empirical analogues
\[
G_{ij}^{(N)}
=
\frac{1}{N}\sum_{t=1}^N \phi_j(x_t)\,\phi_i(x_t),
\qquad
f_i^{(N)}
=
\frac{1}{N}\sum_{t=1}^N y_t\,\phi_i(x_t),
\]
and the Galerkin coefficients $\hat\theta$ are obtained by solving
\begin{equation}
G^{(N)}\,\hat\theta
=
f^{(N)}.
\label{eq:galerkin_empirical}
\end{equation}
The linear system \eqref{eq:galerkin_empirical} coincides with the normal equations for ordinary least squares regression of $y_t$ on the basis evaluations $\{\phi_j(x_t)\}_{j=1}^K$. In this sense, empirical Galerkin projection can be viewed as least squares projection of $b$ onto the span of the chosen basis functions under the data driven inner product induced by the sample.

Classical uses of the Galerkin method include the numerical solution of partial differential equations by projecting the solution onto finite element or spectral bases, as well as reduced order modeling of high dimensional dynamical systems via low dimensional subspaces. In our setting, the target functions $b$ will be conditional mean or innovation maps of time series, the argument $x$ encodes lagged values or residuals, and the measure $\mu$ is the stationary distribution of these lag vectors. Approximating these maps in finite dimensional bases and estimating the coefficients by least squares provides the basic building block for the Galerkin based autoregressive and moving average components that we introduce in the next section.

\subsection{Related Work}

  The way we use Galerkin projection in Galerkin--ARIMA sits at the intersection of nonlinear extensions of autoregressive models, basis expansion methods for time series, and ARIMA extensions with exogenous information. There is a long line of work on relaxing the linearity assumption in autoregressive models. Threshold autoregressive (TAR) and self exciting threshold autoregressive (SETAR) models allow the dynamics to switch across regimes defined by past values of the series, and can generate asymmetric cycles and other nonlinear patterns that a single linear AR process cannot reproduce \citep{Tong1983,Tsay1989}. Smooth transition autoregressive (STAR) models replace hard thresholds by smooth transition functions, leading to gradual regime changes that have been widely used in empirical macroeconomics and finance \citep{GrangerTerasvirta1993,FransesVanDijk2000,vanDijk2002STAR}. Functional coefficient autoregressive models further generalize the AR component by allowing coefficients to vary smoothly with the state \citep{ChenTsay1993}. These models substantially enlarge the class of behaviors beyond ARIMA, but they typically operate at the level of conditional means and do not preserve the full ARMA or SARIMA operator structure that underlies classical Box--Jenkins modeling.

  A second strand of work introduces nonparametric or semiparametric structure through spline and basis expansions. Additive and sieve methods approximate regression functions by combinations of spline or polynomial bases, with asymptotic guarantees on approximation quality and estimation error \citep{Stone1985}. Applied to time series, this leads to spline-based or nonparametric ARIMA variants. For instance, \citet{Sun2022} propose an ARIMA variant where trends are modeled by B-spline bases and combined through model averaging to improve stock price forecasts. Related work uses spline components to model trend and seasonality alongside autoregressive terms or compares parametric ARIMA fits with nonparametric spline regression on the same series. A different direction generalizes ARIMA itself to nonlinear or continuous time models, as in the nonlinear continuous time ARIMA generalization of \citet{Kushnir2023}. These contributions share with our approach the use of basis expansions to increase flexibility, but in most cases the spline component is used as a separate forecasting model or as a preprocessing step rather than as a direct replacement for the autoregressive operator inside an ARIMA or SARIMA specification.

  Exogenous-variable extensions such as ARIMAX enrich the information set by including observable covariates, and are widely used in macroeconomic and financial applications. Our framework is compatible with exogenous factor blocks and provides closed-form estimation and bootstrap-based inference for such blocks within the Galerkin--SARIMA/SARIMAX setup.

  Hybrid models that combine classical time series structure with machine learning components form a related literature. A simple and influential example is the hybrid ARIMA and neural network model of \citet{Zhang2003}, where ARIMA captures the linear component of the series and a feedforward network is trained on ARIMA residuals to model nonlinear structure, with the final forecast given by the sum of the two parts. Subsequent work has proposed many variants of this idea, for example hybrid ARIMA and artificial neural network models that explicitly target nonlinearities that ARIMA cannot represent \citep{Khashei2011}. At a larger scale, the winning method of the M4 forecasting competition was a hybrid of exponential smoothing and recurrent neural networks, showing that carefully structured combinations of classical and deep models can deliver state of the art accuracy in large forecasting benchmarks \citep{Smyl2020}. Deep state space models combine linear Gaussian state space structure with recurrent neural networks that parametrize the state space model, preserving the interpretability and data efficiency of state space models while gaining flexibility from deep learning \citep{Rangapuram2018}. Other architectures, such as LSTNet, explicitly add a classical autoregressive component in parallel with convolutional and recurrent layers to ensure that persistent linear autocorrelation is captured by a simple AR term rather than forcing the neural network to learn it \citep{Lai2018}. N BEATS designs a deep architecture that produces forecasts through learned basis expansions that can be interpreted as trend and seasonal components \citep{Oreshkin2020}. Unlike hybrid ARIMA-neural network models \citep{Zhang2003}, our approach retains a single coherent ARIMA-type operator and delivers closed-form estimators with established asymptotic theory under weak dependence.

  The Galerkin--ARIMA framework proposed in this paper is closest in spirit to the spline based and hybrid extensions of ARIMA. Like basis expansion approaches, we approximate conditional mean and innovation maps by expansions in a chosen basis and determine the coefficients through projection conditions that lead to normal equations. At the same time, we keep the ARIMA and SARIMA backshift operator structure intact and derive closed form least squares estimators for both autoregressive and moving average components. In that sense, Galerkin--ARIMA provides a middle ground between classical linear time series models and fully nonparametric or deep learning approaches while remaining aligned with econometric practice.

\label{sec:galerkin_models}\section{The Galerkin-ARIMA and Galerkin-SARIMA Models}

In this section we embed the Galerkin projection idea into the classical ARIMA and SARIMA frameworks. Starting from the operator form of a Gaussian SARIMA model, we rewrite the conditional mean and innovation terms as functions of lag vectors for the differenced series and its residuals. We then approximate these functions by finite dimensional basis expansions in the lag space and estimate the associated coefficients by least squares, yielding Galerkin based autoregressive and moving average components. The resulting constructions, which we refer to as Galerkin--ARIMA and Galerkin--SARIMA, preserve the familiar backshift operator structure of ARIMA and SARIMA but replace the fixed linear AR terms by flexible basis expansions that can capture nonlinear dependence in past values while still admitting closed form estimators.

\subsection{Problem Setup and Notation}

Let $\{y_t\}_{t=1}^{T}$ be a univariate time series. We allow for both nonseasonal and seasonal differencing. For integers $d\ge 0$ and $D\ge 0$ and a known seasonal period $m\ge 1$, define the combined differenced series
\begin{equation}
y_t^{(d,D)}
\;=\;
(1-B)^d(1-B^{m})^{D}y_t,
\label{eq:diff_series}
\end{equation}
where $B$ is the backshift operator $By_t=y_{t-1}$. After differencing we retain the index set
$t=1,\dots,N$ for notational simplicity, where $N$ is the effective sample size.

For fixed nonseasonal AR order $p\ge 0$, define the nonseasonal lag vector
\begin{equation}
x_t
\;=\;
\bigl(y_{t-1}^{(d,D)},\,y_{t-2}^{(d,D)},\,\dots,\,y_{t-p}^{(d,D)}\bigr)
\;\in\;\mathbb{R}^{p}.
\label{eq:lag_vector_ns}
\end{equation}
For fixed seasonal AR order $P\ge 0$, define the seasonal lag vector
\begin{equation}
x^{(s)}_t
\;=\;
\bigl(y_{t-m}^{(d,D)},\,y_{t-2m}^{(d,D)},\,\dots,\,y_{t-Pm}^{(d,D)}\bigr)
\;\in\;\mathbb{R}^{P}.
\label{eq:lag_vector_s}
\end{equation}
We write $x_{t+1}$ and $x^{(s)}_{t+1}$ for the corresponding lag vectors used in one-step-ahead forecasting.

Let $\{\phi_j(\cdot)\}_{j=1}^{K}$ be a chosen family of basis functions on
$\mathbb{R}^{p}$ for the nonseasonal AR map, and let
$\{\phi^{(s)}_j(\cdot)\}_{j=1}^{K_s}$ be basis functions on $\mathbb{R}^{P}$
for the seasonal AR map. Define the basis evaluation vectors
\begin{equation}
\Phi(x_t)
\;=\;
\bigl(\phi_1(x_t),\,\dots,\,\phi_K(x_t)\bigr)^{\top}
\;\in\;\mathbb{R}^{K},
\qquad
\Phi^{(s)}(x^{(s)}_t)
\;=\;
\bigl(\phi^{(s)}_1(x^{(s)}_t),\,\dots,\,\phi^{(s)}_{K_s}(x^{(s)}_t)\bigr)^{\top}
\;\in\;\mathbb{R}^{K_s}.
\label{eq:basis_vectors_ar}
\end{equation}
In our experiments $\Phi$ and $\Phi^{(s)}$ are polynomial bases in the lagged values
(constant, linear, and quadratic terms), with spline bases as an optional extension.

Let $\epsilon_t$ denote the innovation in the SARIMA reference model.
Given an initial AR projection, we obtain preliminary residuals
$\hat\epsilon_t^{(0)}$ (defined precisely in Section~\ref{sec:galerkin_models}).
For fixed nonseasonal MA order $q\ge 0$ and seasonal MA order $Q\ge 0$, define
the residual lag vectors
\begin{equation}
r_t
\;=\;
\bigl(\hat\epsilon_{t-1}^{(0)},\,\hat\epsilon_{t-2}^{(0)},\,\dots,\,\hat\epsilon_{t-q}^{(0)}\bigr)
\;\in\;\mathbb{R}^{q},
\qquad
r^{(s)}_t
\;=\;
\bigl(\hat\epsilon_{t-m}^{(0)},\,\hat\epsilon_{t-2m}^{(0)},\,\dots,\,\hat\epsilon_{t-Qm}^{(0)}\bigr)
\;\in\;\mathbb{R}^{Q}.
\label{eq:lag_vector_ma}
\end{equation}

Let $\{\psi_\ell(\cdot)\}_{\ell=1}^{L}$ be basis functions on $\mathbb{R}^{q}$
for the nonseasonal MA map, and let
$\{\psi^{(s)}_\ell(\cdot)\}_{\ell=1}^{L_s}$ be basis functions on $\mathbb{R}^{Q}$
for the seasonal MA map. Define
\begin{equation}
\Psi(r_t)
\;=\;
\bigl(\psi_1(r_t),\,\dots,\,\psi_L(r_t)\bigr)^{\top}
\;\in\;\mathbb{R}^{L},
\qquad
\Psi^{(s)}(r^{(s)}_t)
\;=\;
\bigl(\psi^{(s)}_1(r^{(s)}_t),\,\dots,\,\psi^{(s)}_{L_s}(r^{(s)}_t)\bigr)^{\top}
\;\in\;\mathbb{R}^{L_s}.
\label{eq:basis_vectors_ma}
\end{equation}
As with the AR bases, we use polynomial bases by default and allow spline bases
when smoother approximations are desired.

Collecting evaluations over $t=1,\dots,N$, define the empirical design matrices
\begin{equation}
\Phi
=
\bigl[\Phi(x_1)^{\top};\dots;\Phi(x_N)^{\top}\bigr]\in\mathbb{R}^{N\times K},
\quad
\Phi^{(s)}
=
\bigl[\Phi^{(s)}(x^{(s)}_1)^{\top};\dots;\Phi^{(s)}(x^{(s)}_N)^{\top}\bigr]\in\mathbb{R}^{N\times K_s},
\label{eq:design_ar}
\end{equation}
\begin{equation}
\Psi
=
\bigl[\Psi(r_1)^{\top};\dots;\Psi(r_N)^{\top}\bigr]\in\mathbb{R}^{N\times L},
\quad
\Psi^{(s)}
=
\bigl[\Psi^{(s)}(r^{(s)}_1)^{\top};\dots;\Psi^{(s)}(r^{(s)}_N)^{\top}\bigr]\in\mathbb{R}^{N\times L_s}.
\label{eq:design_ma}
\end{equation}
We also define the response vector $Y=(y_1^{(d,D)},\dots,y_N^{(d,D)})^{\top}$.
These objects fully determine the Galerkin--ARIMA and Galerkin--SARIMA estimators
introduced below.

\subsection{Nonseasonal Galerkin-ARIMA}
\label{sec:garima_ns}

We begin with the nonseasonal Galerkin--ARIMA model. Let $\{y_t\}_{t=1}^T$ be a univariate time series and define the $d$ times differenced series
$y_t^{(d)}=(1-B)^d y_t$.
Fix nonseasonal orders $p\ge 0$ and $q\ge 0$, and define the value lag vector
\[
x_t
=
\bigl(y^{(d)}_{t-1},\,y^{(d)}_{t-2},\,\dots,\,y^{(d)}_{t-p}\bigr)
\in\mathbb R^p .
\]
In a classical ARIMA$(p,d,q)$ model the conditional mean is linear in $x_t$ and the moving average correction is linear in the last $q$ innovations. We instead treat both components as unknown smooth maps and write the data generating relationship as
\begin{equation}
y_t^{(d)}
=
f(x_t)
+
g(\epsilon_{t-1},\dots,\epsilon_{t-q})
+
\epsilon_t,
\qquad
\epsilon_t \stackrel{iid}{\sim}\mathcal N(0,\sigma^2),
\label{eq:ns_population}
\end{equation}
where $f:\mathbb R^p\to\mathbb R$ is the autoregressive map and
$g:\mathbb R^q\to\mathbb R$ is the moving average map.

To approximate $f$, choose $K$ basis functions
$\{\phi_j\}_{j=1}^K$ on $\mathbb R^p$ and form the basis evaluation vector
$\Phi(x_t)=(\phi_1(x_t),\dots,\phi_K(x_t))^\top$.
We approximate the autoregressive map by a Galerkin expansion
\begin{equation}
f(x_t)
\approx
\Phi(x_t)^\top\beta,
\qquad
\beta\in\mathbb R^K .
\label{eq:ns_ar_galerkin}
\end{equation}
When $\Phi$ contains only linear coordinates in the lag variables,
$\Phi(x_t)^\top\beta$ reduces to the standard linear AR$(p)$ term.
In our default implementation $\Phi$ is a low order polynomial basis in the lags
(constant, linear, and quadratic terms), with spline bases as an optional extension.

Similarly, to approximate $g$ choose $L$ basis functions
$\{\psi_\ell\}_{\ell=1}^L$ on $\mathbb R^q$ and define
$\Psi(u)=(\psi_1(u),\dots,\psi_L(u))^\top$.
The moving average map is approximated by
\begin{equation}
g(\epsilon_{t-1},\dots,\epsilon_{t-q})
\approx
\Psi(\epsilon_{t-1},\dots,\epsilon_{t-q})^\top\alpha,
\qquad
\alpha\in\mathbb R^L .
\label{eq:ns_ma_galerkin}
\end{equation}
If $\Psi$ includes only the linear residual lags, then this term coincides with the classical MA$(q)$ component. Richer polynomial or spline bases allow nonlinear dependence on past innovations.

Substituting \eqref{eq:ns_ar_galerkin} and \eqref{eq:ns_ma_galerkin} into
\eqref{eq:ns_population} yields the nonseasonal Galerkin--ARIMA specification
\begin{equation}
y_t^{(d)}
=
\Phi(x_t)^\top\beta
+
\Psi(\epsilon_{t-1},\dots,\epsilon_{t-q})^\top\alpha
+
\epsilon_t,
\qquad
\epsilon_t \stackrel{iid}{\sim}\mathcal N(0,\sigma^2).
\label{eq:garima_ns}
\end{equation}
We denote this model by $\mathrm{GARIMA}(p,d,q;K,L)$ to emphasize both the ARIMA orders and the sizes of the Galerkin bases.

Because \eqref{eq:garima_ns} is linear in $(\beta,\alpha)$ conditional on the lag and innovation histories, estimation reduces to least squares once the relevant residual lags are formed. In practice we construct the estimator in two stages: we first fit the AR map by regressing $y_t^{(d)}$ on $\Phi(x_t)$ to obtain preliminary residuals, and then fit the MA map by regressing those residuals on $\Psi(\cdot)$. The resulting two stage estimator and the associated one step ahead forecast are presented in Section~\ref{sec:two_stage}.

\subsection{Seasonal Galerkin-SARIMA Extension}
\label{sec:garima_seasonal}

We now extend the nonseasonal construction to seasonal dynamics. Let $m\ge 1$ be a known seasonal period and define the combined differenced series
$y_t^{(d,D)}=(1-B)^d(1-B^m)^D y_t$ as in \eqref{eq:diff_series}. Fix seasonal orders
$P\ge 0$ and $Q\ge 0$. Along with the nonseasonal lag vector $x_t$ in
\eqref{eq:lag_vector_ns}, define the seasonal value lag vector
$x^{(s)}_t$ in \eqref{eq:lag_vector_s}. For the innovation history we use
the nonseasonal residual lag vector $r_t$ and the seasonal residual lag vector
$r^{(s)}_t$ in \eqref{eq:lag_vector_ma}.

The SARIMA reference model implies that, after differencing, the one step conditional mean can be written as the sum of a nonseasonal component depending on $x_t$ and a seasonal component depending on $x^{(s)}_t$, while the innovation correction depends on both $r_t$ and $r^{(s)}_t$. We therefore write the population relationship as
\begin{equation}
y_t^{(d,D)}
=
f(x_t)
+
f^{(s)}(x^{(s)}_t)
+
g(r_t)
+
g^{(s)}(r^{(s)}_t)
+
\epsilon_t,
\qquad
\epsilon_t \stackrel{iid}{\sim}\mathcal N(0,\sigma^2),
\label{eq:s_population}
\end{equation}
where $f:\mathbb R^p\to\mathbb R$ and $g:\mathbb R^q\to\mathbb R$ are the nonseasonal
AR and MA maps, and $f^{(s)}:\mathbb R^P\to\mathbb R$ and
$g^{(s)}:\mathbb R^Q\to\mathbb R$ are their seasonal counterparts.

Choose basis families $\{\phi_j\}_{j=1}^K$ on $\mathbb R^p$ and
$\{\phi^{(s)}_j\}_{j=1}^{K_s}$ on $\mathbb R^P$ for the AR maps, and
$\{\psi_\ell\}_{\ell=1}^L$ on $\mathbb R^q$ and
$\{\psi^{(s)}_\ell\}_{\ell=1}^{L_s}$ on $\mathbb R^Q$ for the MA maps.
Using the basis evaluation vectors
$\Phi(x_t)$, $\Phi^{(s)}(x^{(s)}_t)$, $\Psi(r_t)$ and
$\Psi^{(s)}(r^{(s)}_t)$ from \eqref{eq:basis_vectors_ar} and
\eqref{eq:basis_vectors_ma}, we approximate
\begin{align}
f(x_t) &\approx \Phi(x_t)^\top \beta, 
&
f^{(s)}(x^{(s)}_t) &\approx \Phi^{(s)}(x^{(s)}_t)^\top \beta^{(s)},
\label{eq:s_ar_galerkin}\\
g(r_t) &\approx \Psi(r_t)^\top \alpha,
&
g^{(s)}(r^{(s)}_t) &\approx \Psi^{(s)}(r^{(s)}_t)^\top \alpha^{(s)},
\label{eq:s_ma_galerkin}
\end{align}
with coefficients
$\beta\in\mathbb R^K$, $\beta^{(s)}\in\mathbb R^{K_s}$,
$\alpha\in\mathbb R^L$, and $\alpha^{(s)}\in\mathbb R^{L_s}$.
When these bases contain only linear lag coordinates, the model reduces to the
classical SARIMA specification.

Substituting \eqref{eq:s_ar_galerkin}--\eqref{eq:s_ma_galerkin} into
\eqref{eq:s_population} yields the Galerkin seasonal model
\begin{equation}
y_t^{(d,D)}
=
\Phi(x_t)^\top \beta
+
\Phi^{(s)}(x^{(s)}_t)^\top \beta^{(s)}
+
\Psi(r_t)^\top \alpha
+
\Psi^{(s)}(r^{(s)}_t)^\top \alpha^{(s)}
+
\epsilon_t,
\qquad
\epsilon_t \stackrel{iid}{\sim}\mathcal N(0,\sigma^2).
\label{eq:garsarima}
\end{equation}
We denote this model by
$\mathrm{GARSARIMA}(p,d,q)\times(P,D,Q)_m; (K,K_s,L,L_s)$.

Given coefficient estimates, the one step ahead forecast for the differenced
series is obtained by evaluating the bases at $t+1$:
\begin{equation}
\hat y_{t+1}^{(d,D)}
=
\Phi(x_{t+1})^\top \hat\beta
+
\Phi^{(s)}(x^{(s)}_{t+1})^\top \hat\beta^{(s)}
+
\Psi(r_{t+1})^\top \hat\alpha
+
\Psi^{(s)}(r^{(s)}_{t+1})^\top \hat\alpha^{(s)}.
\label{eq:garsarima_forecast}
\end{equation}
The practical two stage least squares estimator used to obtain
$(\hat\beta,\hat\beta^{(s)},\hat\alpha,\hat\alpha^{(s)})$ is described next.

\label{sec:two_stage_estimation}\subsection{Two-Stage Least-Squares Estimation}

\label{sec:two_stage}

Estimation of the Galerkin coefficients is carried out in two least-squares stages. The first stage fits the autoregressive maps using value lags; the second stage fits the moving-average maps using residual lags. This mirrors the structure of ARIMA and SARIMA estimation, but with basis expansions in place of fixed linear terms. The two-stage structure induces generated-regressor bias, which we address via block-bootstrap inference (Section~\ref{sec:inference_factors}) that remains valid under \(\beta\)-mixing. The two-stage structure induces generated-regressor bias, which we address via block-bootstrap inference (Section~\ref{sec:inference_factors}) that remains valid under \(\beta\)-mixing.

Let $y_t^{(d,D)}$ be the differenced series and let
$\Phi,\Phi^{(s)},\Psi,\Psi^{(s)}$ and $Y$ be the design matrices and response
vector defined in \eqref{eq:design_ar}–\eqref{eq:design_ma}. For notational
clarity, write
\[
Y
=
\bigl(y^{(d,D)}_{t_0},\dots,y^{(d,D)}_{N}\bigr)^{\top},
\]
where $t_0$ is chosen large enough that all required lags are available, for example
$t_0 = 1 + \max\{p,Pm,q,Qm\}$. All design matrices below are implicitly restricted
to rows $t=t_0,\dots,N$.

\paragraph{Stage 1: autoregressive projection.}
The first stage approximates the conditional mean using only value lags. In the
nonseasonal case this corresponds to fitting the map $f$ in
\eqref{eq:ns_ar_galerkin}; in the seasonal case it corresponds to fitting
$f$ and $f^{(s)}$ in \eqref{eq:s_ar_galerkin}. Stacking the nonseasonal and
seasonal bases, define the combined AR design matrix
\[
X_{\mathrm{AR}}
=
\bigl[\,
\Phi \;\;\; \Phi^{(s)}
\,\bigr]
\in\mathbb R^{(N-t_0+1)\times (K+K_s)},
\]
and the AR coefficient vector
\[
\gamma_{\mathrm{AR}}
=
\begin{pmatrix}
\beta \\[0.3em] \beta^{(s)}
\end{pmatrix}
\in\mathbb R^{K+K_s}.
\]
The first-stage estimator is the ordinary least-squares solution
\begin{equation}
\hat\gamma_{\mathrm{AR}}
=
\arg\min_{\gamma_{\mathrm{AR}}}
\bigl\|Y - X_{\mathrm{AR}}\gamma_{\mathrm{AR}}\bigr\|_2^2
=
\bigl(X_{\mathrm{AR}}^{\top}X_{\mathrm{AR}}\bigr)^{-1}
X_{\mathrm{AR}}^{\top}Y,
\label{eq:stage1_ols}
\end{equation}
provided $X_{\mathrm{AR}}^{\top}X_{\mathrm{AR}}$ is nonsingular. The fitted AR
contribution is then
\[
\hat m_t
=
\Phi(x_t)^{\top}\hat\beta
+
\Phi^{(s)}(x^{(s)}_t)^{\top}\hat\beta^{(s)},
\qquad
t=t_0,\dots,N,
\]
with the obvious simplification $\hat m_t=\Phi(x_t)^{\top}\hat\beta$ in the purely
nonseasonal case. The corresponding first-stage residuals,
which approximate the innovations up to moving-average structure, are
\begin{equation}
\hat\epsilon_t^{(1)}
=
y_t^{(d,D)} - \hat m_t,
\qquad
t=t_0,\dots,N.
\label{eq:stage1_residuals}
\end{equation}

\paragraph{Stage 2: moving-average projection.}
The second stage models the remaining serial dependence in the residuals
$\hat\epsilon_t^{(1)}$ via basis expansions in residual lags. Using
\eqref{eq:lag_vector_ma}, form the nonseasonal and seasonal residual lag vectors
$r_t$ and $r_t^{(s)}$ from $\hat\epsilon_t^{(1)}$, and evaluate the MA bases to
obtain $\Psi(r_t)$ and $\Psi^{(s)}(r_t^{(s)})$. Stacking these, define the MA
design matrix
\[
X_{\mathrm{MA}}
=
\bigl[\,
\Psi \;\;\; \Psi^{(s)}
\,\bigr]
\in\mathbb R^{(N-t_0+1)\times (L+L_s)},
\]
and the MA coefficient vector
\[
\gamma_{\mathrm{MA}}
=
\begin{pmatrix}
\alpha \\[0.3em] \alpha^{(s)}
\end{pmatrix}
\in\mathbb R^{L+L_s}.
\]
The second-stage estimator solves
\begin{equation}
\hat\gamma_{\mathrm{MA}}
=
\arg\min_{\gamma_{\mathrm{MA}}}
\bigl\|\hat\epsilon^{(1)} - X_{\mathrm{MA}}\gamma_{\mathrm{MA}}\bigr\|_2^2
=
\bigl(X_{\mathrm{MA}}^{\top}X_{\mathrm{MA}}\bigr)^{-1}
X_{\mathrm{MA}}^{\top}\hat\epsilon^{(1)},
\label{eq:stage2_ols}
\end{equation}
where $\hat\epsilon^{(1)}$ is the vector of residuals from
\eqref{eq:stage1_residuals}. The fitted moving-average correction at time $t$ is
\[
\hat g_t
=
\Psi(r_t)^{\top}\hat\alpha
+
\Psi^{(s)}(r_t^{(s)})^{\top}\hat\alpha^{(s)},
\qquad
t=t_0,\dots,N.
\]

\paragraph{One-step-ahead forecast.}
Given $(\hat\beta,\hat\beta^{(s)},\hat\alpha,\hat\alpha^{(s)})$ from
\eqref{eq:stage1_ols} and \eqref{eq:stage2_ols}, the one-step-ahead forecast
for the differenced series at time $t+1$ is
\begin{equation}
\hat y_{t+1}^{(d,D)}
=
\Phi(x_{t+1})^{\top}\hat\beta
+
\Phi^{(s)}(x^{(s)}_{t+1})^{\top}\hat\beta^{(s)}
+
\Psi(r_{t+1})^{\top}\hat\alpha
+
\Psi^{(s)}(r_{t+1}^{(s)})^{\top}\hat\alpha^{(s)},
\label{eq:two_stage_forecast}
\end{equation}
where $x_{t+1},x^{(s)}_{t+1}$ are formed from the most recent differenced values
and $r_{t+1},r^{(s)}_{t+1}$ are formed from the most recent residuals
$\hat\epsilon_t^{(1)}$. In the nonseasonal Galerkin--ARIMA case,
$\Phi^{(s)}$ and $\Psi^{(s)}$ drop out and \eqref{eq:two_stage_forecast}
reduces to the nonseasonal formula in Section~\ref{sec:garima_ns}.

This two-stage estimator coincides with an empirical Galerkin projection of the
conditional mean and innovation maps onto the chosen bases, and will be the
object of the theoretical analysis in Section~\ref{sec:theory_unpenalized}.

\subsection{Ridge-Regularized Galerkin-ARIMA}
\label{sec:ridge_garima}

In practice, rich polynomial or spline bases can introduce multicollinearity and
large variance in the least-squares estimates, especially when the number of
basis functions is non-negligible relative to the effective sample size. In the
time-series setting this can lead to fitted autoregressive maps with steep
slopes in certain directions and moving-average corrections that overreact to
individual residuals, producing transient spikes in one-step-ahead forecasts.
To mitigate these instabilities we introduce a ridge-regularized version of the
two-stage estimator. The ridge variant retains the same model structure as
Sections~\ref{sec:garima_ns}–\ref{sec:two_stage}, but replaces ordinary least
squares by Tikhonov-regularized least squares in both stages.

For the autoregressive stage, let $X_{\mathrm{AR}}$ and $\gamma_{\mathrm{AR}}$
be as in \eqref{eq:stage1_ols}. We augment the objective with quadratic
penalties on the coefficient blocks $\beta$ and $\beta^{(s)}$:
\begin{equation}
(\hat\beta_{\lambda},\hat\beta^{(s)}_{\lambda})
\in
\arg\min_{\beta,\beta^{(s)}}
\frac{1}{N}\bigl\|
Y - \Phi\beta - \Phi^{(s)}\beta^{(s)}
\bigr\|_2^2
+
\lambda_{\beta}\,\|W_{\beta}\beta\|_2^2
+
\lambda_{\beta^{(s)}}\,\|W_{\beta^{(s)}}\beta^{(s)}\|_2^2,
\label{eq:ridge_ar}
\end{equation}
where $\lambda_{\beta},\lambda_{\beta^{(s)}}\ge 0$ are tuning parameters and
$W_{\beta},W_{\beta^{(s)}}$ are diagonal weight matrices. In applications we
keep intercept terms unpenalized and choose $W_{\beta}$ and $W_{\beta^{(s)}}$ so
that higher-order basis functions receive stronger penalties, for instance
through polynomial or exponential growth of the diagonal entries. We set
the $\lambda$'s via generalized cross-validation on the initial window, with
polynomial weights $W_{\beta,ii}=i^2$ (higher-order terms more heavily
penalized), and analogously for $W_{\beta^{(s)}}$, $W_{\alpha}$, and
$W_{\alpha^{(s)}}$. The solution
can be written in closed form as
\[
\hat\beta_{\lambda}
=
\bigl(\Phi^{\top}\Phi + N\lambda_{\beta} W_{\beta}^{\top}W_{\beta}\bigr)^{-1}
\Phi^{\top}Y',
\qquad
\hat\beta^{(s)}_{\lambda}
=
\bigl(\Phi^{(s)\top}\Phi^{(s)} + N\lambda_{\beta^{(s)}} W_{\beta^{(s)}}^{\top}W_{\beta^{(s)}}\bigr)^{-1}
\Phi^{(s)\top}Y'_{s},
\]
for suitable centered responses $Y'$ and $Y'_{s}$. In matrix notation the
regularized normal matrix is obtained by adding a positive semidefinite term to
$X_{\mathrm{AR}}^{\top}X_{\mathrm{AR}}$, improving its conditioning.

Given the penalized AR fit, the first-stage residuals are
\[
\hat\epsilon_{t,\lambda}^{(1)}
=
y_t^{(d,D)}
-
\Phi(x_t)^{\top}\hat\beta_{\lambda}
-
\Phi^{(s)}(x_t^{(s)})^{\top}\hat\beta^{(s)}_{\lambda},
\]
and are used to form the residual lag vectors $r_t$ and $r_t^{(s)}$ as before.
The moving-average stage applies the same idea to the MA bases. With
$X_{\mathrm{MA}}$ and $\gamma_{\mathrm{MA}}$ as in \eqref{eq:stage2_ols}, we solve
\begin{equation}
(\hat\alpha_{\lambda},\hat\alpha^{(s)}_{\lambda})
\in
\arg\min_{\alpha,\alpha^{(s)}}
\frac{1}{N}\bigl\|
\hat\epsilon^{(1)}_{\lambda} - \Psi\alpha - \Psi^{(s)}\alpha^{(s)}
\bigr\|_2^2
+
\lambda_{\alpha}\,\|W_{\alpha}\alpha\|_2^2
+
\lambda_{\alpha^{(s)}}\,\|W_{\alpha^{(s)}}\alpha^{(s)}\|_2^2,
\label{eq:ridge_ma}
\end{equation}
with tuning parameters $\lambda_{\alpha},\lambda_{\alpha^{(s)}}\ge 0$ and
diagonal weights $W_{\alpha},W_{\alpha^{(s)}}$. The resulting solutions again
have closed forms with regularized normal matrices
$X_{\mathrm{MA}}^{\top}X_{\mathrm{MA}} + N\Lambda$, where $\Lambda$ collects the
penalty terms.

The penalized one-step-ahead forecast for the differenced series at time $t+1$
is obtained by substituting the ridge estimates into
\eqref{eq:two_stage_forecast}:
\begin{equation}
\hat y_{t+1,\lambda}^{(d,D)}
=
\Phi(x_{t+1})^{\top}\hat\beta_{\lambda}
+
\Phi^{(s)}(x_{t+1}^{(s)})^{\top}\hat\beta^{(s)}_{\lambda}
+
\Psi(r_{t+1})^{\top}\hat\alpha_{\lambda}
+
\Psi^{(s)}(r_{t+1}^{(s)})^{\top}\hat\alpha^{(s)}_{\lambda}.
\label{eq:ridge_forecast_garsarima}
\end{equation}

Throughout the theoretical analysis in Section~\ref{sec:theory_unpenalized} we
focus on the unpenalized estimator obtained by setting all
$\lambda$’s to zero. The ridge-regularized variant is introduced as a
practical extension: the additional quadratic penalties stabilize the normal
equations, reduce the sensitivity of the fitted maps to high-order basis
directions, and, as we show in the experiments, substantially dampen the
transient forecast spikes that can appear when using rich bases with limited
data.
\label{sec:theory_unpenalized}\section{Theoretical Analysis of the Unpenalized Estimator}

In this section we study the large-sample behaviour of the two-stage
Galerkin estimator without regularization, that is, with all ridge
penalties set to zero. The analysis is carried out under a Gaussian
SARIMA data-generating process and mild smoothness and regularity
conditions on the Galerkin bases. Throughout, $N$ denotes the effective
sample size after differencing and discarding the first
$\max\{p,Pm,q,Qm\}$ observations so that all required lags are available.

\subsection{Assumptions}
\label{sec:assumptions}

We collect here the conditions used in the approximation error,
consistency, and asymptotic distribution results that follow. All
probabilistic statements are with respect to the stationary law of the
differenced series $\{y_t^{(d,D)}\}$.

\begin{enumerate}
  \item[(A1)] \textit{Data-generating process.}
    The differenced series $y_t^{(d,D)}$ is generated by a stable and
    invertible Gaussian SARIMA model,
    \begin{equation}
      \Phi(B)\,\Phi_s(B^m)\,y_t^{(d,D)}
      =
      \Theta(B)\,\Theta_s(B^m)\,\epsilon_t,
      \qquad
      \epsilon_t \stackrel{iid}{\sim} \mathcal N(0,\sigma^2),
      \label{eq:assump_dgp}
    \end{equation}
    where the roots of $\Phi(z)\Phi_s(z^m)$ and
    $\Theta(z)\Theta_s(z^m)$ lie outside the unit circle. The process
    $\{y_t^{(d,D)}\}$ is strictly stationary, ergodic, and \(\beta\)-mixing
    with $\sum_{m\ge 1}\beta(m)^{1/2}<\infty$ (as in \citet{Doukhan1994}).

  \item[(A2)] \textit{Smoothness of target maps.}
    Let $x_t$ and $x_t^{(s)}$ be the value lag vectors and $r_t$ and
    $r_t^{(s)}$ the residual lag vectors defined in
    \eqref{eq:lag_vector_ns}, \eqref{eq:lag_vector_s}, and
    \eqref{eq:lag_vector_ma}. After rescaling these vectors to a
    compact domain such as $[0,1]^d$, the conditional mean and
    innovation maps
    \[
    f(x) = \mathbb E\bigl[y_t^{(d,D)} \mid x_t = x\bigr], \qquad
    f^{(s)}(x^{(s)}) = \mathbb E\bigl[y_t^{(d,D)} \mid x_t^{(s)} = x^{(s)}\bigr],
    \]
    \[
    g(r) = \mathbb E\bigl[\epsilon_t \mid r_t = r\bigr], \qquad
    g^{(s)}(r^{(s)}) = \mathbb E\bigl[\epsilon_t \mid r_t^{(s)} = r^{(s)}\bigr]
    \]
    belong to a Hölder class $\mathcal H^{r}$ with exponent $r>0$:
    there exists $L<\infty$ such that for all $u,v$ in the domain,
    \[
    |h(u) - h(v)|
    \le
    L \,\|u-v\|_{\infty}^{\,r}, \qquad h\in\{f,f^{(s)},g,g^{(s)}\}.
    \]
    In particular, the linear SARIMA maps implied by
    \eqref{eq:assump_dgp} satisfy this condition for every $r>0$.

  \item[(A3)] \textit{Design regularity.}
    Let $\Psi_N$ denote the full design matrix obtained by stacking all
    AR and MA basis evaluations,
    \[
      \Psi_N
      =
      \bigl[\,
        \Phi \;\;\; \Phi^{(s)} \;\;\; \Psi \;\;\; \Psi^{(s)}
      \,\bigr]
      \in\mathbb R^{N\times d_G},
      \qquad
      d_G = K + K_s + L + L_s.
    \]
    There exist constants $0 < c_0 < c_1 < \infty$ such that
    \[
      c_0
      \le
      \lambda_{\min}\!\bigl( N^{-1} \Psi_N^{\top}\Psi_N \bigr)
      \le
      \lambda_{\max}\!\bigl( N^{-1} \Psi_N^{\top}\Psi_N \bigr)
      \le
      c_1
    \]
    for all sufficiently large $N$, where
    $\lambda_{\min}$ and $\lambda_{\max}$ denote the smallest and
    largest eigenvalues. This ensures that the Gram matrix of the
    basis functions is well conditioned in the limit.

  \item[(A4)] \textit{Basis growth.}
    The numbers of nonseasonal and seasonal basis functions
    $K,K_s,L,L_s$ may grow with the sample size, but satisfy
    \[
      K \to \infty, \quad
      K_s \to \infty, \quad
      L \to \infty, \quad
      L_s \to \infty,
      \qquad
      K + K_s + L + L_s = o(N)
      \quad (N\to\infty).
    \]
    In particular, the total number of Galerkin coefficients grows
    slower than linearly in $N$.

    \item[(A5)] \textit{Linear oracle regime and basis containment.}
    Suppose the true data-generating process is linear SARIMA as in Assumption (A1).
    Then there exists a finite-dimensional parameter vector
    $\gamma_0 = (\beta_0^\top, \beta_0^{(s)\top}, \alpha_0^\top, \alpha_0^{(s)\top})^\top$
    such that the conditional mean admits the representation
    \[
    m_{t} = 
    \Phi(x_t)^\top \beta_0
    +
    \Phi^{(s)}(x_t^{(s)})^\top \beta_0^{(s)}
    +
    \Psi(r_t)^\top \alpha_0
    +
    \Psi^{(s)}(r_t^{(s)})^\top \alpha_0^{(s)}.
    \]
    Moreover, for all sufficiently large $K,K_s,L,L_s$, 
    the Galerkin bases contain the corresponding linear coordinates 
    (constant and first-order lag terms), so that the linear SARIMA model 
    is nested within the Galerkin class and the approximation error vanishes:
    \[
    \Delta_f = \Delta_{f^{(s)}} = \Delta_g = \Delta_{g^{(s)}} = 0.
    \]
\end{enumerate}

Assumption (A1) provides a convenient reference model and guarantees the
existence of a stationary lag process with good mixing properties.
Assumption (A2) is a smoothness condition on the target maps that
allows us to control the approximation error of the basis expansions
via Jackson-type bounds. Assumption (A3) is a standard eigenvalue
condition that ensures identifiability and stable inversion of the
empirical Gram matrices. Assumption (A4) balances approximation and
estimation: a richer basis reduces approximation bias but increases
variance, and the requirement $d_G = o(N)$ keeps the overall variance
under control.

\subsection{Linear Oracle Regime}
\label{sec:linear_oracle}
Under Assumption (A5), the true conditional mean lies in the span of the Galerkin bases, so the
population projection coefficients defined in \eqref{eq:pop_proj_beta}--\eqref{eq:pop_proj_alpha_s}
coincide with the true parameter vector $\gamma_0$.
Hence the approximation error term in \eqref{eq:forecast_decomp} vanishes identically.

\begin{theorem}[Linear Oracle Property]
\label{thm:linear_oracle}
Suppose Assumptions (A1)–(A5) hold and that the true conditional mean
is linear SARIMA. Then:

\begin{enumerate}
\item[(i)] The population Galerkin coefficients satisfy
$\gamma^\star = \gamma_0$.

\item[(ii)] The two-stage Galerkin estimator achieves the parametric rate:
\[
\|\hat\gamma - \gamma_0\|_2 = O_p(N^{-1/2}).
\]

\item[(iii)] If the basis sizes are fixed,
\[
\sqrt{N}(\hat\gamma - \gamma_0)
\;\Rightarrow\;
\mathcal{N}(0,\sigma^2 \Sigma^{-1}),
\]
where
\[
\Sigma = \plim_{N\to\infty} N^{-1}\Psi_N^\top \Psi_N
\]
is positive definite under Assumption (A3).
\end{enumerate}
\end{theorem}

Under Assumption (A5), the true conditional mean lies in the span of
the Galerkin bases, so the population projection coefficients
defined in \eqref{eq:beta_star}–\eqref{eq:pop_proj_alpha_s} coincide with the true parameter vector
$\gamma_0$. Hence the approximation error term in
\eqref{eq:forecast_decomp} vanishes identically.
The estimator therefore behaves as an ordinary least-squares estimator
with fixed finite dimension.
The parametric rate and asymptotic normality follow from standard
least-squares theory under mixing conditions and Assumption (A3).
A detailed proof is given in Appendix~\ref{app:proof_linear_oracle}.

\subsection{Approximation Error}

  \label{sec:approx_error}

We first separate the \emph{approximation} error coming from projecting the
population maps $f,f^{(s)},g,g^{(s)}$ onto finite bases from the \emph{estimation}
error that arises when these projections are estimated from data. The goal of
this subsection is to make precise the population projection targets and to
record the rates at which the basis expansions can approximate the true maps
under Assumption~\ref{sec:assumptions}.

Recall the value and residual lag vectors $x_t,x_t^{(s)},r_t,r_t^{(s)}$ and the
Galerkin bases $\Phi,\Phi^{(s)},\Psi,\Psi^{(s)}$ from
Section~\ref{sec:garima_seasonal}. Let $X,X^{(s)},R,R^{(s)}$ denote generic
random variables with the stationary distributions of $x_t,x_t^{(s)},r_t,r_t^{(s)}$,
respectively. The \emph{population} Galerkin coefficients are defined as the
$L^2$ projections of the target maps onto the chosen bases. For the AR maps we
set
\begin{align}
\beta^{\star}
&\in
\arg\min_{\beta\in\mathbb R^{K}}
\mathbb E\bigl[
\bigl(f(X) - \Phi(X)^{\top}\beta\bigr)^2
\bigr],
\label{eq:pop_proj_beta}
\\
\beta^{(s)\star}
&\in
\arg\min_{\beta^{(s)}\in\mathbb R^{K_s}}
\mathbb E\bigl[
\bigl(f^{(s)}(X^{(s)}) - \Phi^{(s)}(X^{(s)})^{\top}\beta^{(s)}\bigr)^2
\bigr],
\label{eq:beta_star}
\end{align}
and for the MA maps,
\begin{align}
\alpha^{\star}
&\in
\arg\min_{\alpha\in\mathbb R^{L}}
\mathbb E\bigl[
  \bigl(g(R) - \Psi(R)^{\top}\alpha\bigr)^2
\bigr],
\label{eq:pop_proj_alpha}
\\
\alpha^{(s)\star}
&\in
\arg\min_{\alpha^{(s)}\in\mathbb R^{L_s}}
\mathbb E\bigl[
\bigl(g^{(s)}(R^{(s)}) - \Psi^{(s)}(R^{(s)})^{\top}\alpha^{(s)}\bigr)^2
\bigr],
\label{eq:pop_proj_alpha_s}
\end{align}
Under Assumption~(A3) these minimizers are unique and satisfy the population
normal equations
\[
\mathbb E\bigl[\Phi(X)\Phi(X)^{\top}\bigr]\beta^{\star}
=
\mathbb E\bigl[\Phi(X) f(X)\bigr],
\quad
\mathbb E\bigl[\Psi(R)\Psi(R)^{\top}\bigr]\alpha^{\star}
=
\mathbb E\bigl[\Psi(R) g(R)\bigr],
\]
and similarly for the seasonal coefficients. The sample least-squares estimators
studied below can be viewed as empirical analogues of these projections.

The quality of the Galerkin approximation depends on how well the finite bases
$\Phi,\Phi^{(s)},\Psi,\Psi^{(s)}$ can approximate the Hölder-smooth maps in
Assumption~(A2). To describe this we introduce the deterministic approximation
errors
\begin{align*}
\Delta_f
&=
\sup_{x}
\bigl| f(x) - \Phi(x)^{\top}\beta^{\star} \bigr|,
&
\Delta_{f^{(s)}}
&=
\sup_{x^{(s)}}
\bigl| f^{(s)}(x^{(s)}) - \Phi^{(s)}(x^{(s)})^{\top}\beta^{(s)\star} \bigr|,
\\
\Delta_g
&=
\sup_{r}
\bigl| g(r) - \Psi(r)^{\top}\alpha^{\star} \bigr|,
&
\Delta_{g^{(s)}}
&=
\sup_{r^{(s)}}
\bigl| g^{(s)}(r^{(s)}) - \Psi^{(s)}(r^{(s)})^{\top}\alpha^{(s)\star} \bigr|.
\end{align*}
These measure how far the best $L^2$ projections can be from the true maps in
supremum norm.

We work with polynomial or spline bases on rescaled lag spaces of fixed
dimension. Standard results from approximation theory then give Jackson-type
bounds that control these errors in terms of the basis sizes and the Hölder
smoothness exponent $r$.

\begin{lemma}[Jackson-type approximation bounds]
\label{lem:jackson}
Suppose Assumption~\textup{(A2)} holds and that the bases
$\Phi,\Phi^{(s)},\Psi,\Psi^{(s)}$ are tensor-product polynomial or spline bases
of sizes $K,K_s,L,L_s$ on compact domains for $X,X^{(s)},R,R^{(s)}$. Let $p$,
$P$, $q$, and $Q$ denote the dimensions of $X$, $X^{(s)}$, $R$, and $R^{(s)}$.
Then there exist constants $C_1,\dots,C_4<\infty$ independent of the basis
sizes such that
\begin{align}
\Delta_f
&\le C_1\, K^{-r/p}, &
\Delta_{f^{(s)}}
&\le C_2\, K_s^{-r/P},
\label{eq:jackson_ar}
\\
\Delta_g
&\le C_3\, L^{-r/q}, &
\Delta_{g^{(s)}}
&\le C_4\, L_s^{-r/Q}.
\label{eq:jackson_ma}
\end{align}
\end{lemma}

The proof is standard: one partitions each lag domain into a grid with mesh
size $h\asymp K^{-1/p}$ (or $K_s^{-1/P}$, and similarly for $L$ and $L_s$) and
uses the Hölder condition to bound the modulus of continuity of the target map
on each cell. For polynomial or spline bases, the associated projection
operators can approximate any Hölder function uniformly on the domain at rate
$h^{r}$, which yields \eqref{eq:jackson_ar}–\eqref{eq:jackson_ma}. A detailed
argument is given in Appendix~\ref{app:lemma_jackson_proof}.

Two observations are useful for interpreting these bounds. First, in the
Gaussian SARIMA setting of Assumption~(A1) the maps $f,f^{(s)},g,g^{(s)}$ are
linear functions of their arguments, so in principle they can be represented
exactly once the bases include the corresponding linear coordinates. In that
case the approximation errors are zero for all sufficiently large
$K,K_s,L,L_s$, and the only remaining source of error is estimation. Second, in
more general nonlinear settings the Jackson bounds quantify the trade-off
between basis richness and approximation quality: increasing the numbers of
basis functions reduces the deterministic bias at the rates in
\eqref{eq:jackson_ar}–\eqref{eq:jackson_ma}, at the cost of increasing the
variance of the estimated coefficients. The balance between these two effects
determines the overall mean-squared error and will be made precise in
Section~\ref{sec:asymptotic_distribution}.

\subsection{Consistency of One-Step-Ahead Forecasts}
\label{sec:consistency}

We now study the behaviour of the one step ahead forecasts produced by the
unpenalized two stage estimator. The goal is to separate approximation error
(from using finite bases) and estimation error (from estimating the projection
coefficients) and to show that, under Assumptions~(A1)–(A4), the forecast for
the differenced series is asymptotically unbiased and consistent for the
population conditional mean.

Let
\[
m_{t+1}
=
f(x_{t+1})
+
f^{(s)}(x^{(s)}_{t+1})
+
g(r_{t+1})
+
g^{(s)}(r^{(s)}_{t+1})
\]
denote the population one step ahead conditional mean of
$y_{t+1}^{(d,D)}$ given the value and residual lag vectors at time $t+1$, and
write the population decomposition
\[
y_{t+1}^{(d,D)} = m_{t+1} + \epsilon_{t+1},
\qquad
\epsilon_{t+1} \sim \mathcal N(0,\sigma^2).
\]
Let $(\beta^{\star},\beta^{(s)\star},\alpha^{\star},\alpha^{(s)\star})$ be the
population Galerkin coefficients defined in
\eqref{eq:pop_proj_beta}–\eqref{eq:pop_proj_alpha_s}, and define the associated
\emph{oracle} Galerkin forecast
\begin{equation}
\tilde y_{t+1}^{(d,D)}
=
\Phi(x_{t+1})^{\top}\beta^{\star}
+
\Phi^{(s)}(x^{(s)}_{t+1})^{\top}\beta^{(s)\star}
+
\Psi(r_{t+1})^{\top}\alpha^{\star}
+
\Psi^{(s)}(r^{(s)}_{t+1})^{\top}\alpha^{(s)\star}.
\label{eq:oracle_forecast}
\end{equation}
By construction, $\tilde y_{t+1}^{(d,D)}$ is the best forecast within the
chosen Galerkin classes when the projection coefficients are known.

The empirical two stage estimator of Section~\ref{sec:two_stage} replaces the
population Galerkin coefficients by their least squares estimates
$(\hat\beta,\hat\beta^{(s)},\hat\alpha,\hat\alpha^{(s)})$ and yields the
plug in forecast
\begin{equation}
\hat y_{t+1}^{(d,D)}
=
\Phi(x_{t+1})^{\top}\hat\beta
+
\Phi^{(s)}(x^{(s)}_{t+1})^{\top}\hat\beta^{(s)}
+
\Psi(r_{t+1})^{\top}\hat\alpha
+
\Psi^{(s)}(r^{(s)}_{t+1})^{\top}\hat\alpha^{(s)},
\label{eq:empirical_forecast}
\end{equation}
which coincides with \eqref{eq:two_stage_forecast} written in compact form.

The forecast error can be decomposed into three parts:
\begin{equation}
\hat y_{t+1}^{(d,D)} - y_{t+1}^{(d,D)}
=
\underbrace{\bigl(\tilde y_{t+1}^{(d,D)} - m_{t+1}\bigr)}_{\text{approximation error}}
+
\underbrace{\bigl(\hat y_{t+1}^{(d,D)} - \tilde y_{t+1}^{(d,D)}\bigr)}_{\text{estimation error}}
-
\underbrace{\epsilon_{t+1}}_{\text{innovation}}.
\label{eq:forecast_decomp}
\end{equation}
The first term is purely deterministic and is controlled by the Jackson bounds
of Lemma~\ref{lem:jackson}; the second term captures the effect of estimating
the projection coefficients from $N$ observations; the third term is the
irreducible noise.

The next theorem summarizes the main consistency properties of the unpenalized
forecast.

\begin{theorem}[Asymptotic unbiasedness and consistency]
\label{thm:consistency}
Suppose Assumptions~\textup{(A1)}–\textup{(A4)} hold and the basis sizes
$K,K_s,L,L_s$ grow with $N$ as in Assumption~\textup{(A4)}.
Let $\hat y_{t+1}^{(d,D)}$ be the one step ahead forecast defined in
\eqref{eq:empirical_forecast} and $m_{t+1}$ the population conditional mean.
Then, for any fixed time index $t$,
\begin{align}
\bigl|
\mathbb E\bigl[\hat y_{t+1}^{(d,D)} - y_{t+1}^{(d,D)}\bigr]
\bigr|
&\longrightarrow 0,
\label{eq:asymp_unbiased}
\\[0.3em]
\hat y_{t+1}^{(d,D)} - m_{t+1}
&\xrightarrow{p} 0,
\qquad N\to\infty.
\label{eq:forecast_consistency}
\end{align}
In particular, the Galerkin forecast is asymptotically unbiased for
$y_{t+1}^{(d,D)}$ and consistent for the conditional mean $m_{t+1}$.
\end{theorem}

The approximation error term in \eqref{eq:forecast_decomp} satisfies
\[
\bigl|\tilde y_{t+1}^{(d,D)} - m_{t+1}\bigr|
\le
\Delta_f + \Delta_{f^{(s)}} + \Delta_g + \Delta_{g^{(s)}},
\]
by the definitions in Section~\ref{sec:approx_error}. Lemma~\ref{lem:jackson}
and the growth condition in Assumption~(A4) imply that this bound converges to
zero as $K,K_s,L,L_s\to\infty$, so the approximation bias vanishes.

For the estimation error term, standard least squares theory under mixing
conditions and Assumption~(A3) gives
\[
\|\hat\beta - \beta^{\star}\|_2 = O_p(N^{-1/2}),\quad
\|\hat\beta^{(s)} - \beta^{(s)\star}\|_2 = O_p(N^{-1/2}),
\]
and similarly for $\hat\alpha$ and $\hat\alpha^{(s)}$. Since the basis
evaluation vectors at time $t+1$ are bounded in probability, it follows that
\[
\hat y_{t+1}^{(d,D)} - \tilde y_{t+1}^{(d,D)}
=
O_p(N^{-1/2}).
\]
Combining these bounds with the decomposition
\eqref{eq:forecast_decomp} and noting that
$\mathbb E[\epsilon_{t+1}]=0$ yields \eqref{eq:asymp_unbiased} and
\eqref{eq:forecast_consistency}. A detailed proof is given in
Appendix~\ref{app:proof_unbiased}.
\medskip

Theorem~\ref{thm:consistency} shows that, under mild conditions, the only
asymptotic contribution to the forecast error is the innovation itself. The
next subsection refines this picture by describing the joint asymptotic
distribution of the Galerkin coefficient estimates and deriving optimal
mean-squared error rates for the one step ahead forecast.

\subsection{Asymptotic Distribution and MSE Rates}
\label{sec:asymptotic_distribution}

We next describe the joint asymptotic distribution of the unpenalized Galerkin
coefficient estimates and the resulting mean squared error behaviour of the
one step ahead forecast. Throughout this subsection we keep the basis sizes
$K,K_s,L,L_s$ fixed while deriving the asymptotic distribution, and then let
them grow with $N$ subject to Assumption~(A4) when studying mean squared error
rates.

Let
\[
\gamma^{\star}
=
\bigl(
  \beta^{\star},
  \beta^{(s)\star},
  \alpha^{\star},
  \alpha^{(s)\star}
\bigr)
\in\mathbb R^{d_G},
\qquad
d_G = K + K_s + L + L_s,
\]
be the stacked vector of population Galerkin coefficients defined by
\eqref{eq:pop_proj_beta}–\eqref{eq:pop_proj_alpha_s}, and let
\[
\hat\gamma
=
\bigl(
  \hat\beta,
  \hat\beta^{(s)},
  \hat\alpha,
  \hat\alpha^{(s)}
\bigr)
\]
be the corresponding two stage least squares estimator. Recall the full design
matrix
\[
\Psi_N
=
\bigl[\,
  \Phi \;\;\; \Phi^{(s)} \;\;\; \Psi \;\;\; \Psi^{(s)}
\bigr]
\in\mathbb R^{N\times d_G}
\]
from Assumption~(A3).

Under the Gaussian SARIMA model and the mixing and eigenvalue conditions in
Assumptions~(A1) and (A3), $\hat\gamma$ behaves asymptotically as a standard
least squares estimator with random design.

\begin{proposition}[Asymptotic normality of the Galerkin coefficients]
\label{prop:clt_gamma}
Suppose Assumptions~\textup{(A1)}–\textup{(A3)} hold and that the basis sizes
$K,K_s,L,L_s$ are fixed. Let
\[
\Sigma
=
\operatorname*{plim}_{N\to\infty}
N^{-1} \Psi_N^{\top}\Psi_N,
\]
which is positive definite by Assumption~\textup{(A3)}.
Then, as $N\to\infty$,
\begin{equation}
\sqrt{N}\,\bigl(\hat\gamma - \gamma^{\star}\bigr)
\;\xrightarrow{d}\;
\mathcal N\bigl(0,\sigma^2\Sigma^{-1}\bigr),
\label{eq:gamma_clt}
\end{equation}
and, in particular,
\begin{equation}
\|\hat\gamma - \gamma^{\star}\|_2 = O_p(N^{-1/2}).
\label{eq:gamma_rate}
\end{equation}
\end{proposition}

The proof is standard and is given in Appendix~\ref{app:proof_clt}. The key
steps are the representation of $\hat\gamma - \gamma^{\star}$ as
$(\Psi_N^{\top}\Psi_N)^{-1}\Psi_N^{\top}\epsilon$, where
$\epsilon=(\epsilon_{t_0},\dots,\epsilon_N)^{\top}$ is the vector of
innovations, the convergence of $N^{-1}\Psi_N^{\top}\Psi_N$ to $\Sigma$, and a
central limit theorem for the martingale difference array
$\Psi_N^{\top}\epsilon / \sqrt{N}$ under the mixing conditions in
Assumption~(A1).

We now turn to the mean squared error of the one step ahead forecast. For a
fixed time index $t$, write the forecast error as
\[
\hat y_{t+1}^{(d,D)} - y_{t+1}^{(d,D)}
=
\bigl(\tilde y_{t+1}^{(d,D)} - m_{t+1}\bigr)
+
\bigl(\hat y_{t+1}^{(d,D)} - \tilde y_{t+1}^{(d,D)}\bigr)
-
\epsilon_{t+1},
\]
as in \eqref{eq:forecast_decomp}. Taking squared expectation and using
$\mathbb E[\epsilon_{t+1}]=0$ and $\mathbb E[\epsilon_{t+1}^2]=\sigma^2$ gives
\[
\mathbb E\bigl[(\hat y_{t+1}^{(d,D)} - y_{t+1}^{(d,D)})^2\bigr]
=
\sigma^2
+
\underbrace{
  \mathbb E\bigl[(\tilde y_{t+1}^{(d,D)} - m_{t+1})^2\bigr]
}_{\text{approximation term}}
+
\underbrace{
  \mathbb E\bigl[(\hat y_{t+1}^{(d,D)} - \tilde y_{t+1}^{(d,D)})^2\bigr]
}_{\text{estimation term}}.
\]
The irreducible part $\sigma^2$ is the variance of the innovation and cannot be
reduced by any estimator. The approximation term is controlled by the
Jackson-type bounds in Lemma~\ref{lem:jackson}, and the estimation term is
controlled by Proposition~\ref{prop:clt_gamma} and the growth condition in
Assumption~(A4). Combining these yields the following mean squared error rate.

\begin{proposition}[Optimal mean squared error rates]
\label{prop:mse_rates}
Suppose Assumptions~\textup{(A1)}–\textup{(A4)} hold and that the basis sizes
$K,K_s,L,L_s$ grow with $N$ according to
\[
K \asymp N^{p/(2r+p)},
\qquad
K_s \asymp N^{P/(2r+P)},
\qquad
L \asymp K,
\qquad
L_s \asymp K_s.
\]
Then the excess mean squared error of the one step ahead forecast satisfies
\begin{equation}
\mathbb E\bigl[(\hat y_{t+1}^{(d,D)} - y_{t+1}^{(d,D)})^2\bigr]
-
\sigma^2
=
O\bigl(N^{-2r/(2r+p)}\bigr)
+
O\bigl(N^{-2r/(2r+P)}\bigr),
\qquad N\to\infty,
\label{eq:mse_rate}
\end{equation}
where $p$ and $P$ are the dimensions of the nonseasonal and seasonal value lag
vectors.
\end{proposition}

A detailed proof is given in Appendix~\ref{app:proof_mse}. At a high level, the
approximation term behaves like
\[
\Delta_f^2 + \Delta_{f^{(s)}}^2 + \Delta_g^2 + \Delta_{g^{(s)}}^2
=
O\bigl(K^{-2r/p}\bigr)
+
O\bigl(K_s^{-2r/P}\bigr)
+
O\bigl(L^{-2r/q}\bigr)
+
O\bigl(L_s^{-2r/Q}\bigr),
\]
by Lemma~\ref{lem:jackson}, while the estimation term behaves like
$O\bigl((K+K_s+L+L_s)/N\bigr)$ by \eqref{eq:gamma_rate} and Assumption~(A3).
Balancing these two contributions in the nonseasonal and seasonal parts gives
the stated choices of $K$ and $K_s$ and the rate \eqref{eq:mse_rate}.

In the Gaussian SARIMA case of Assumption~(A1), where the true maps are
linear, the approximation term vanishes once the bases include the appropriate
linear coordinates, and the excess mean squared error is driven entirely by
the estimation term. In more general nonlinear settings,
Proposition~\ref{prop:mse_rates} shows that the Galerkin forecasts achieve
nonparametric rates of convergence governed by the smoothness exponent $r$ and
the effective dimensions $p$ and $P$ of the lag spaces, while retaining the
computational advantages described in the next subsection.

\subsection{Risk Comparison with Classical SARIMA}
\label{sec:risk_comparison}

We now compare the large-sample prediction risk of the Galerkin forecast
with that of a classical linear SARIMA forecast.
The comparison clarifies in which sense the Galerkin class strictly extends
the linear SARIMA class and can achieve smaller asymptotic risk
when the true conditional mean is nonlinear.

Recall that
\[
y^{(d,D)}_{t+1} = m_{t+1} + \epsilon_{t+1},
\qquad
\mathbb{E}[\epsilon_{t+1}\mid \mathcal{F}_t]=0,
\qquad
\mathrm{Var}(\epsilon_{t+1})=\sigma^2.
\]
For any one-step-ahead forecast $\hat y_{t+1}$ measurable with respect to
the lag information at time $t+1$, define the mean-squared prediction risk
\[
R(\hat y_{t+1})
=
\mathbb{E}\!\left[(\hat y_{t+1}-y^{(d,D)}_{t+1})^2\right].
\]
Using the decomposition above,
\[
R(\hat y_{t+1})
=
\sigma^2
+
\mathbb{E}\!\left[(\hat y_{t+1}-m_{t+1})^2\right],
\]
so comparisons reduce to the squared deviation from the conditional mean.

Let $\mathcal{L}$ denote the class of linear SARIMA predictors,
that is, functions linear in the lag vectors
$(x_{t+1}, x^{(s)}_{t+1}, r_{t+1}, r^{(s)}_{t+1})$
with the same coordinate structure as in Assumption (A5).
Equivalently, $\mathcal{L}$ is the finite-dimensional linear span of
the constant and first-order lag coordinates embedded in the Galerkin bases.

Define the pseudo-true linear predictor as the $L_2$ projection
of $m_{t+1}$ onto $\mathcal{L}$:
\[
\ell^\star
\in
\arg\min_{\ell\in\mathcal{L}}
\mathbb{E}\!\left[(m_{t+1}-\ell)^2\right],
\]
and denote the corresponding projection error by
\[
\delta_{\text{lin}}^2
=
\mathbb{E}\!\left[(m_{t+1}-\ell^\star)^2\right].
\]

Since $\mathcal{L}$ is finite-dimensional and closed in $L_2$,
$\delta_{\text{lin}}^2$ is strictly positive whenever
$m_{t+1}\notin\mathcal{L}$ almost surely.
In that case, no linear SARIMA forecast can eliminate this error.

\begin{lemma}
\label{lem:lin_misspec}
Suppose $m_{t+1}\notin\mathcal{L}$ almost surely.
Then $\delta_{\text{lin}}^2>0$ and for any sequence of linear SARIMA
forecasts $\hat y^{\mathrm{lin}}_{t+1}\in\mathcal{L}$,
\[
\liminf_{N\to\infty}
R(\hat y^{\mathrm{lin}}_{t+1})
\;\ge\;
\sigma^2+\delta_{\text{lin}}^2.
\]
\end{lemma}

We next compare this lower bound with the risk of the Galerkin forecast.
Recall the oracle Galerkin predictor $\tilde y^{(d,D)}_{t+1}$ defined in (38)
and the decomposition (40):
\[
\hat y^{(d,D)}_{t+1}-m_{t+1}
=
(\tilde y^{(d,D)}_{t+1}-m_{t+1})
+
(\hat y^{(d,D)}_{t+1}-\tilde y^{(d,D)}_{t+1}).
\]
The first term is the approximation error controlled by Lemma 1,
and the second term is the estimation error controlled by
standard least-squares arguments under Assumption (A3).

\begin{theorem}
\label{thm:risk_dominance}
Suppose Assumptions (A2)--(A4) hold and that
$m_{t+1}\notin\mathcal{L}$.
Then
\[
R(\hat y^{(d,D)}_{t+1})
=
\sigma^2+o(1),
\]
whereas any linear SARIMA forecast satisfies
\[
\liminf_{N\to\infty}
R(\hat y^{\mathrm{lin}}_{t+1})
\;\ge\;
\sigma^2+\delta_{\text{lin}}^2.
\]
Consequently,
\[
\liminf_{N\to\infty}
\Big(
R(\hat y^{\mathrm{lin}}_{t+1})
-
R(\hat y^{(d,D)}_{t+1})
\Big)
\ge
\delta_{\text{lin}}^2
>0.
\]
\end{theorem}

Economically, when the true conditional mean is nonlinear (e.g., threshold
effects in unemployment), the Galerkin forecast strictly dominates linear
SARIMA in asymptotic mean-squared risk, with the gap $\delta_{\text{lin}}^2>0$
representing irreducible linear misspecification error.

The proof follows directly from the identity
$R(\hat y)=\sigma^2+\mathbb{E}[(\hat y-m_{t+1})^2]$.
Lemma 1 and the growth condition in Assumption (A4)
imply that the approximation error of the Galerkin oracle predictor
converges to zero.
Standard least-squares theory under mixing conditions and Assumption (A3)
implies that the estimation error term also converges to zero.
In contrast, the linear SARIMA class $\mathcal{L}$ is finite-dimensional,
so when $m_{t+1}\notin\mathcal{L}$ the projection error
$\delta_{\text{lin}}^2$ remains strictly positive.
A detailed proof is given in Appendix~\ref{app:proof_risk_dominance}.

\subsection{Basis Size Selection via Information Criteria}
\label{sec:basis_selection}

The preceding analysis assumes that the basis sizes grow with the sample size
according to Assumption (A4).
In practice, however, the numbers of basis functions
$(K,K_s,L,L_s)$ must be selected in a data-driven manner.
We show that a BIC-type information criterion consistently selects
a risk-optimal basis dimension under the same regularity conditions.

Let $\Theta_{K,L}$ denote the Galerkin parameter space
corresponding to basis sizes $(K,K_s,L,L_s)$.
Define the residual sum of squares
\[
\mathrm{RSS}(K,L)
=
\sum_{t=t_0}^{T-1}
\bigl(y^{(d,D)}_{t+1}-\hat y^{(d,D)}_{t+1}(K,L)\bigr)^2.
\]

Let
\[
\mathrm{BIC}(K,L)
=
\log\!\Big(\frac{\mathrm{RSS}(K,L)}{N}\Big)
+
\frac{\dim(\Theta_{K,L})\log N}{N},
\]
where $\dim(\Theta_{K,L})$ denotes the total number of free parameters
in both stages.
We also report results with AIC and HQIC; conclusions are robust.

We consider the selected model
\[
(\hat K,\hat L)
=
\arg\min_{K,L}
\mathrm{BIC}(K,L).
\]

\begin{theorem}
\label{thm:bic_consistency}
Suppose Assumptions (A2)–(A4) hold and that the true conditional mean
$m_{t+1}$ belongs to the closure of the Galerkin sieve space.
Then
\[
(\hat K,\hat L)
\;\to\;
(K^\star,L^\star)
\quad\text{in probability},
\]
where $(K^\star,L^\star)$ minimizes the asymptotic prediction risk.
Moreover,
\[
R\bigl(\hat y^{(d,D)}_{t+1}(\hat K,\hat L)\bigr)
=
\sigma^2
+
o_p(1).
\]
\end{theorem}

The proof follows standard sieve BIC arguments.
For underfitted models,
the approximation error dominates and yields a strictly larger
population risk.
For overfitted models,
the penalty term
$\dim(\Theta_{K,L})\log N/N$
dominates the stochastic fluctuation of the residual term.
Under Assumption (A4),
the candidate model space grows slowly enough to ensure
uniform stochastic equicontinuity.
A detailed proof is given in Appendix~\ref{app:proof_bic}.

\subsection{Computational Complexity}
\label{sec:complexity}

We finally compare the computational cost of fitting a classical SARIMA model by
maximum likelihood with that of a single two stage Galerkin fit. The goal is to
make precise in which regimes the Galerkin estimator can deliver the empirical
speedups observed in our experiments.

Consider first a Gaussian SARIMA model with nonseasonal orders $(p,q)$,
seasonal orders $(P,Q)$ and period $m$. Standard implementations estimate
$(\phi_1,\dots,\phi_p,\Phi_1,\dots,\Phi_P,\theta_1,\dots,\theta_q,\Theta_1,\dots,\Theta_Q,\sigma^2)$
by maximizing the conditional log likelihood via an iterative numerical
optimizer such as BFGS or a Newton type method. Each iteration requires a full
pass through the time series to evaluate the likelihood and its gradient, which
costs on the order of $(p+P+q+Q)\,N$ operations for a series of length $N$. If
the optimizer takes $I$ iterations to converge, the total cost of a single
SARIMA fit is
\begin{equation}
\mathrm{Cost}_{\mathrm{SARIMA}}
=
\Theta\bigl(I\,(p+P+q+Q)\,N\bigr).
\label{eq:cost_sarima}
\end{equation}
In rolling forecast settings, this optimization is repeated at each forecast
origin.

For the two stage Galerkin estimator, the cost has two main components. The
first component is the construction of the AR and MA design matrices and the
corresponding normal matrices. Evaluating the AR bases $\Phi$ and $\Phi^{(s)}$
at $N$ time points costs $\Theta(NpK)$ and $\Theta(N P K_s)$ operations,
respectively, so the total cost of forming the AR normal matrix
$X_{\mathrm{AR}}^{\top}X_{\mathrm{AR}}$ is of the same order. The MA design has
analogous cost with $L$ and $L_s$, but these are typically chosen of the same
order as $K$ and $K_s$, so we write the overall design formation cost as
\begin{equation}
\mathrm{Cost}_{\mathrm{design}}
=
\Theta\bigl(N(pK + P K_s)\bigr).
\label{eq:cost_design}
\end{equation}
The second component is solving the Galerkin normal equations. Stacking all
coefficients in $\gamma=(\beta,\beta^{(s)},\alpha,\alpha^{(s)})$, the combined
system has dimension $d_G=K+K_s+L+L_s$. A direct Cholesky decomposition of the
$d_G\times d_G$ normal matrix costs $\Theta(d_G^3)$ operations, so the total
cost of a single Galerkin fit is
\begin{equation}
\mathrm{Cost}_{\mathrm{Gal}}
=
\Theta\bigl(N(pK + P K_s)\bigr)
+
\Theta\bigl((K+K_s+L+L_s)^3\bigr).
\label{eq:cost_gal}
\end{equation}

When the basis sizes are moderate and do not grow with $N$, the cubic term in
\eqref{eq:cost_gal} is negligible and the dominant cost is linear in $N$ with a
constant proportional to $pK + P K_s$. In contrast, the SARIMA likelihood
optimisation in \eqref{eq:cost_sarima} incurs a linear cost in $N$ with a
constant proportional to $I(p+P+q+Q)$, where typically $I$ is in the tens and
larger orders $(p,q,P,Q)$ are required to capture complex dynamics. In this
regime,
\[
\frac{\mathrm{Cost}_{\mathrm{Gal}}}{\mathrm{Cost}_{\mathrm{SARIMA}}}
\approx
\frac{pK + P K_s}{I(p+P+q+Q)},
\]
which can be very small when $K,K_s,L,L_s$ are chosen to be modest and the
optimizer requires many iterations.

In rolling forecast applications, the difference is amplified. A full refit of
a SARIMA model at each forecast origin repeats the iterative optimization, so
the total cost scales with both the series length and the number of forecast
steps. The Galerkin estimator, by contrast, requires only the evaluation of
basis functions and the solution of a single linear system at each refit. This
explains the several order of magnitude speedups observed in
Sections~\ref{sec:synthetic_experiments} and \ref{sec:real_experiments}.
A more formal statement and proof of the complexity comparison are given in
Appendix~\ref{app:proof_complexity}.

\subsection{Inference for Exogenous Factors}
\label{sec:inference_factors}

This subsection develops inference procedures for assessing the incremental predictive contribution
of exogenous factors within the Galerkin--SARIMA framework. The tests below concern conditional-mean
(one-step) predictive content relative to the model's information set, analogous to ARIMAX-style
significance, and are not interpreted as structural causal effects.

Let $z_t \in \mathbb{R}^r$ denote a vector of observed factors (exogenous regressors), potentially including
contemporaneous and/or lagged covariates. We augment the conditional mean specification in
Section~\ref{sec:two_stage_estimation} by adding a linear factor block:
\begin{equation}
y_t^{(d,D)}
=
\Phi(x_t)^\top\beta
+\Phi^{(s)}(x_t^{(s)})^\top\beta^{(s)}
+ z_t^\top \delta
+\Psi(r_t)^\top\alpha
+\Psi^{(s)}(r_t^{(s)})^\top\alpha^{(s)}
+\varepsilon_t,
\qquad t=t_0,\ldots,T-1,
\label{eq:galerkin_sarimax}
\end{equation}
where $(x_t,x_t^{(s)})$ are value-lag vectors and $(r_t,r_t^{(s)})$ are residual-lag vectors as defined in
Appendix~A.1. We stack parameters as
\[
\theta
=
\big(\beta^\top,\ \beta^{(s)\top},\ \delta^\top,\ \alpha^\top,\ \alpha^{(s)\top}\big)^\top
\in\mathbb{R}^{d_G+r}.
\]

Estimation proceeds by the same two-stage least squares procedure as in Section~\ref{sec:two_stage_estimation},
with the factor block included in Stage~1. Specifically, Stage~1 fits $(\beta,\beta^{(s)},\delta)$ by least squares using
the design matrix $[\Phi,\Phi^{(s)},Z]$, where $Z\in\mathbb{R}^{N\times r}$ has $t$-th row $z_t^\top$.
Let $\widehat\varepsilon^{(1)}$ denote the resulting residuals. Stage~2 then proceeds as in Section~\ref{sec:two_stage_estimation},
using residual lags formed from $\widehat\varepsilon^{(1)}$ to fit $(\alpha,\alpha^{(s)})$ by least squares. The final two-stage
estimator is denoted $\widehat\theta$.

We assess factor relevance via the null that factors provide no incremental predictive contribution, conditional on the Galerkin
AR and MA components:
\begin{equation}
H_0:\ \delta = 0.
\label{eq:null_delta_block}
\end{equation}
We also report coordinate-wise inference for $H_0:\delta_j=0$ as a secondary output when $r$ is small.

Under mixing and eigenvalue conditions analogous to Assumptions (A1)--(A3) for the enlarged design matrix including $Z$,
the two-stage estimator admits a joint asymptotic normal approximation:
\begin{equation}
\sqrt{N}\,(\widehat\theta-\theta^\star)
\Rightarrow
\mathcal{N}\!\left(0,\ \Omega\right),
\label{eq:clt_theta_factor}
\end{equation}
where $\theta^\star$ is the pseudo-true coefficient vector for the finite basis projection and $\Omega$ is the long-run covariance
of an asymptotic linear representation that accounts for time-series dependence and the generated-regressor structure induced by
Stage~2. Consequently, for any linear restriction $R\theta=r_0$ (with full row rank), the Wald statistic
\begin{equation}
W
=
(R\widehat\theta-r_0)^\top
\big(R\widehat\Omega R^\top\big)^{-1}
(R\widehat\theta-r_0)
\label{eq:wald_factor}
\end{equation}
is asymptotically $\chi^2_{\mathrm{rank}(R)}$, provided $\widehat\Omega$ is a consistent estimator of $\Omega$.
In particular, taking $R$ to select the factor block yields a $\chi^2_r$ test of \eqref{eq:null_delta_block}.

In implementation, while a feasible HAC/sandwich estimator $\widehat\Omega$ can be constructed, its explicit analytic form must
account for the dependence of Stage~2 regressors on Stage~1 residuals. We therefore adopt a block bootstrap that re-estimates
the full two-stage pipeline and directly captures both serial dependence and generated-regressor uncertainty.

Fix a block length $b$ and number of replications $B$. For $s=1,\ldots,B$: (i) resample blocks of the joint sequence
$\{(y_t,z_t)\}$ (moving-block or circular bootstrap) to obtain $\{(y_t^{*(s)},z_t^{*(s)})\}$; (ii) apply the same differencing/seasonal
differencing to form $\{y_t^{*(s),(d,D)}\}$; (iii) re-run Stage~1 and Stage~2 with the same basis families and hyperparameters,
producing $\widehat\theta^{*(s)}$. Bootstrap standard errors and confidence intervals for any scalar functional
$\vartheta=h(\theta)$ are obtained from the empirical distribution of $\widehat\vartheta^{*(s)}=h(\widehat\theta^{*(s)})$.
For the block test \eqref{eq:null_delta_block}, we compute a bootstrap analogue of the Wald statistic in \eqref{eq:wald_factor}
(or, equivalently, obtain critical values from the bootstrap distribution of $R\widehat\theta^{*(s)}$).
A formal statement and proof of bootstrap validity under \(\beta\)-mixing assumptions are given in Appendix~\ref{app:factor_inference_proofs}.

Finally, we report (i) block $p$-values for $H_0:\delta=0$ as the primary inferential output, (ii) coordinate-wise confidence intervals
for $\delta_j$ when $r$ is small, and (iii) robustness checks across moderate changes in block length $b$.

\subsection{Prediction Intervals and Forecast Uncertainty}
\label{sec:prediction_intervals}

This subsection constructs uncertainty measures for forecasts produced by the Galerkin--SARIMA/SARIMAX
procedure. We distinguish between uncertainty for the conditional mean forecast and prediction
intervals for future observations.

Let $\hat m_{t+1}$ denote the one-step-ahead conditional mean forecast produced by the fitted model
at forecast origin $t$ (cf.\ Section~\ref{sec:two_stage_estimation}), so that the forecast error is
$e_{t+1}=y_{t+1}^{(d,D)}-\hat m_{t+1}$. A forecast confidence interval targets the conditional mean
$m_{t+1}=\mathbb{E}[y_{t+1}^{(d,D)}\mid \mathcal{F}_t]$, whereas a prediction interval targets the
future observation $y_{t+1}^{(d,D)}$ itself and therefore incorporates innovation uncertainty.

We construct both objects using the block bootstrap described in Section~\ref{sec:inference_factors}.
For $s=1,\ldots,B$, let $\widehat\theta^{*(s)}$ denote the bootstrap re-estimate obtained by re-running
the full two-stage procedure on resampled blocks of the joint sequence $\{(y_t,z_t)\}$, and let
$\hat m_{t+1}^{*(s)}$ denote the corresponding bootstrap conditional mean forecast at time $t+1$.

A confidence interval for the conditional mean is obtained from the empirical distribution of
$\{\hat m_{t+1}^{*(s)}\}_{s=1}^B$ via bootstrap quantiles:
\begin{equation}
\mathrm{CI}_{1-\alpha}(m_{t+1})
=
\Big[
Q_{\alpha/2}\big(\hat m_{t+1}^{*(1:B)}\big),
\,
Q_{1-\alpha/2}\big(\hat m_{t+1}^{*(1:B)}\big)
\Big],
\label{eq:ci_mean_forecast}
\end{equation}
where $Q_u(\cdot)$ denotes the empirical $u$-quantile.

To obtain a prediction interval for $y_{t+1}^{(d,D)}$, we additionally resample an innovation term.
For each bootstrap replication $s$, let $\{\hat\varepsilon_{t}^{*(s)}\}$ denote the fitted residuals
from the bootstrap model (after Stage~2) and draw $\varepsilon_{t+1}^{*(s)}$ by sampling with replacement
from $\{\hat\varepsilon_{t}^{*(s)}\}$, optionally after centering.
Define the bootstrap predictive draw
\[
\tilde y_{t+1}^{*(s)}=\hat m_{t+1}^{*(s)}+\varepsilon_{t+1}^{*(s)}.
\]
The $(1-\alpha)$ prediction interval is then
\begin{equation}
\mathrm{PI}_{1-\alpha}(y_{t+1}^{(d,D)})
=
\Big[
Q_{\alpha/2}\big(\tilde y_{t+1}^{*(1:B)}\big),
\,
Q_{1-\alpha/2}\big(\tilde y_{t+1}^{*(1:B)}\big)
\Big].
\label{eq:pi_one_step}
\end{equation}

The same construction extends to $h$-step-ahead prediction by iterating the fitted model forward
within each bootstrap replication and sampling an innovation at each step. Formal conditions under
which \eqref{eq:ci_mean_forecast} and \eqref{eq:pi_one_step} achieve asymptotically correct coverage under
\(\beta\)-mixing are given in Appendix~\ref{app:prediction_interval_proofs}.

\label{sec:synthetic_experiments}\section{Experiments on Synthetic Data}
\label{sec:synthetic}

We begin with controlled experiments on simulated time series where the data-generating mechanisms are known. This setting provides a clean sandbox to isolate approximation error from estimation noise and to verify that the Galerkin--SARIMA construction behaves as intended across regimes that are favorable to classical linear models as well as regimes where nonlinear structure is present. In particular, these experiments are designed to answer three questions: (i) can Galerkin--SARIMA match the performance of a well-specified linear SARIMA model on purely linear ARMA dynamics, (ii) does the basis expansion yield tangible gains when the true conditional mean exhibits seasonality, trend, or nonlinear recurrence that a low-order linear ARIMA model cannot fully capture, and (iii) how large are the computational savings in rolling one-step-ahead forecasting, where models must be repeatedly refit across many forecast origins.

To address these questions, we consider four synthetic processes that cover noisy linear ARMA dynamics, smooth periodic patterns, trend with autoregressive feedback, and a nonlinear recursion. For each process we generate series of moderate length and evaluate rolling one-step-ahead forecasts under a common protocol. We report accuracy metrics (MAE and RMSE) and wall-clock runtime aggregated across forecast origins, highlighting the accuracy--runtime trade-off between maximum-likelihood SARIMA fitting and the two-stage closed-form Galerkin estimator.
\subsection{Data-Generating Processes}
\label{sec:synthetic_dgp}

We study Galerkin--SARIMA under synthetic data where the true dynamics are known. This controlled setting lets us separate approximation error from estimation noise and check whether the method recovers simple linear structure when it is present, while also capturing trend, seasonality, and nonlinear recurrence when linear ARMA structure is misspecified. Unless stated otherwise, each synthetic series has length $n=300$. For each data-generating process we generate $R$ independent replications and average results across replications.

We consider four processes that cover linear ARMA dynamics, seasonality, deterministic trend with feedback, and nonlinear recursion.
\begin{enumerate}
\item \textbf{Noisy ARMA.}
A stationary linear process with autoregressive and moving-average components:
\begin{equation}
y_t
=
0.6\,y_{t-1}
-
0.3\,y_{t-2}
+
0.5\,\varepsilon_{t-1}
+
\varepsilon_t,
\qquad
\varepsilon_t \sim \mathcal N(0,1),
\qquad
y_0=y_1=0 .
\label{eq:dgp_arma}
\end{equation}
This design is well matched to low-order SARIMA and serves as a baseline where classical likelihood-based fitting is expected to perform well.

\item \textbf{Seasonal sine with noise.}
A smooth periodic signal with period $m=20$ and additive Gaussian noise:
\begin{equation}
y_t
=
\sin\!\Big(\frac{2\pi t}{20}\Big)
+
0.5\,\eta_t,
\qquad
\eta_t \sim \mathcal N(0,1).
\label{eq:dgp_seasonal}
\end{equation}
This process contains strong seasonal structure at period $20$ but no long-run trend.

\item \textbf{Linear trend plus AR.}
A linearly trending mean combined with stable autoregressive dependence:
\begin{equation}
y_t
=
0.01\,t
+
0.8\,y_{t-1}
+
\nu_t,
\qquad
\nu_t \sim \mathcal N(0,0.5^2),
\qquad
y_0=0 .
\label{eq:dgp_trend}
\end{equation}
This design tests robustness to deterministic trend together with short-range dependence.

\item \textbf{Nonlinear recursion.}
A noisy logistic map with heavy-tailed innovations and measurement noise:
\begin{equation}
y_t
=
\operatorname{clip}\!\Big(
3.8\,y_{t-1}\bigl(1-y_{t-1}\bigr)
+
\xi_t,\,
0,\,
1
\Big)
+
\zeta_t,
\qquad
\xi_t \sim 0.02\,t_{3},
\qquad
\zeta_t \sim \mathcal N(0,0.01^2),
\qquad
y_0=0.4 .
\label{eq:dgp_logistic}
\end{equation}
The conditional mean is nonlinear in $y_{t-1}$, so purely linear ARMA specifications are misspecified. This design probes whether the Galerkin basis expansion can approximate nonlinear recurrence under non-Gaussian shocks.
\end{enumerate}

Together these processes span regimes that are favorable to classical linear models as well as regimes where nonlinear structure is present. Section~\ref{sec:synthetic_protocol} describes the rolling forecasting protocol and the model-selection procedure used in the comparisons.

\subsection{Forecasting Protocol}
\label{sec:synthetic_protocol}

For each synthetic data generating process we evaluate forecasting performance in a rolling one step ahead setting with a fixed calibration window. The total series length is $n=300$. We set the rolling window length to $W=80$ and evaluate forecasts over a horizon of $H=100$ time points. At each forecast origin $i=n-H,\ldots,n-1$ we fit the model on the most recent $W$ observations $\{y_{i-W+1},\ldots,y_i\}$ and produce the one step ahead forecast of $y_{i+1}$. This protocol matches the intended deployment setting where the model is repeatedly refit on a sliding window.

We compare three estimators. The first is a classical SARIMA model estimated by Gaussian maximum likelihood using the statsmodels implementation. The second is the unpenalized Galerkin--SARIMA estimator described in Section~3.4. The third is the ridge regularized variant described in Section~3.5. To avoid favoring either method through manual tuning, each algorithm selects its own order by an information criterion on the initial calibration window. Specifically, for each algorithm and each synthetic series we search over a common grid of nonseasonal and seasonal orders,
\[
p\in\{0,1,2,3,4,5\},\qquad
q\in\{0,1,2,3,4,5\},\qquad
P\in\{0,1\},\qquad
Q\in\{0,1\},
\qquad
m=20,
\]
and choose the order quadruple that minimizes the algorithm specific BIC. After this initial selection, the chosen order is held fixed throughout the rolling evaluation, and the model is refit at each forecast origin using that fixed order.

For the Galerkin estimators we use the same basis family in both model selection and rolling refits. In all synthetic experiments the basis family contains linear, sigmoid, and quadratic components, and we standardize the resulting regressors prior to estimation. Unless stated otherwise, ridge hyperparameters are held fixed across synthetic series.

We report forecasting accuracy and computational cost aggregated over the $H$ rolling forecast origins. Accuracy is summarized by mean absolute error and root mean squared error,
\[
\mathrm{MAE}
=
\frac{1}{H}\sum_{i=n-H}^{n-1}\bigl|y_{i+1}-\widehat y_{i+1}\bigr|,
\qquad
\mathrm{RMSE}
=
\Bigg(
\frac{1}{H}\sum_{i=n-H}^{n-1}\bigl(y_{i+1}-\widehat y_{i+1}\bigr)^2
\Bigg)^{1/2}.
\]
Computational cost is measured by wall clock time required to complete all rolling refits and forecasts for a given algorithm and process, as well as throughput in rolling iterations per second.

To reduce Monte Carlo variability, we repeat the full experiment $R$ times with independent noise realizations for each data generating process and report averages over these replications.

\subsection{Accuracy and Runtime Results}
\label{sec:synthetic_results}

\begin{figure}[t]
\centering
\begin{subfigure}{\linewidth}
\centering
\includegraphics[width=\linewidth]{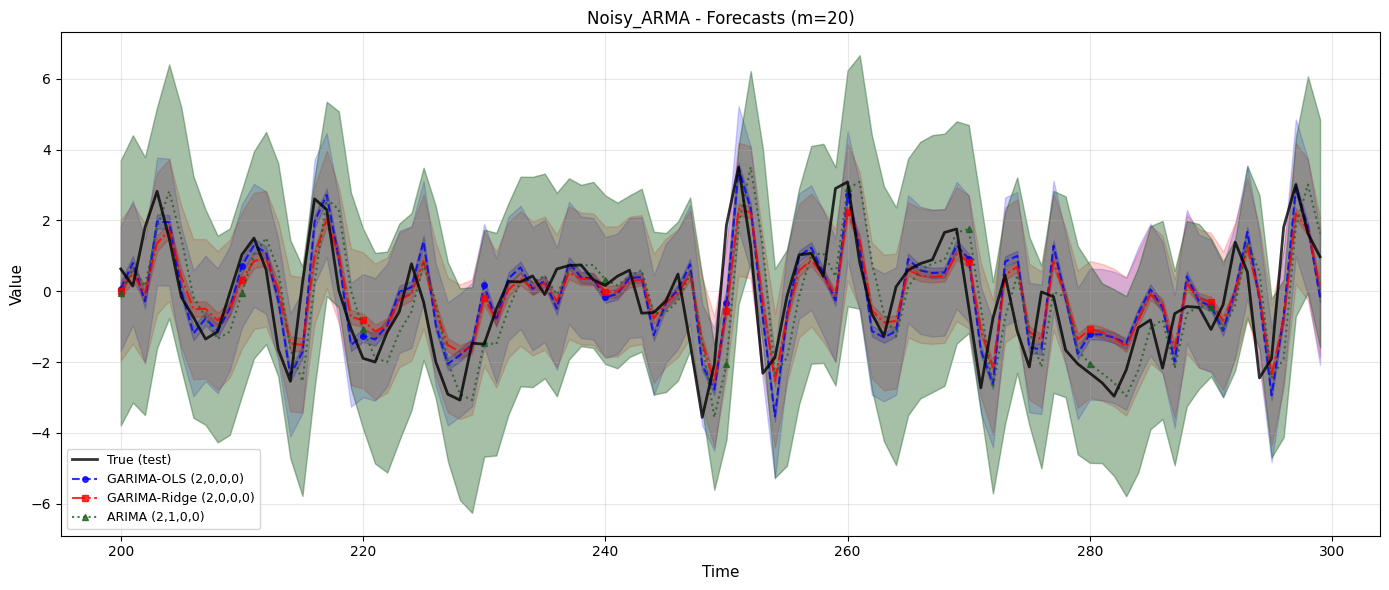}
\caption{Noisy ARMA.}
\end{subfigure}

\vspace{0.4em}

\begin{subfigure}{\linewidth}
\centering
\includegraphics[width=\linewidth]{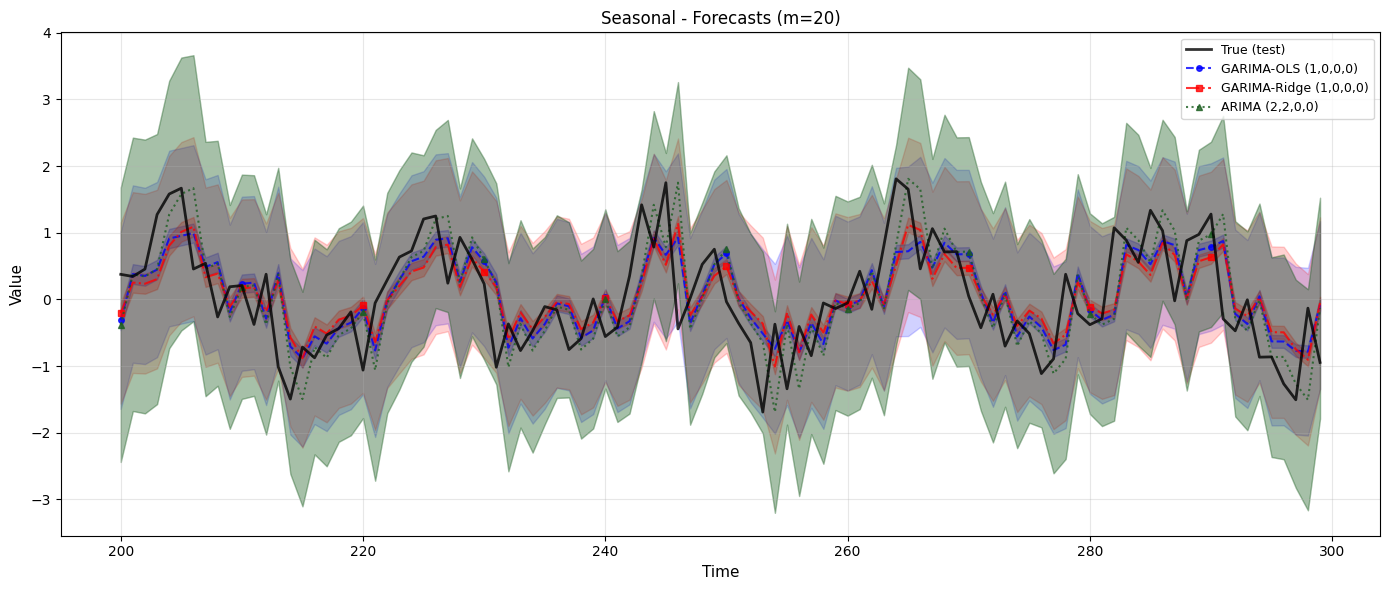}
\caption{Seasonal sine with period $m=20$.}
\end{subfigure}
\caption{Synthetic series used in the experiments. Part I.}
\label{fig:synthetic_series_1}
\end{figure}

\begin{figure}[t]
\centering
\begin{subfigure}{\linewidth}
\centering
\includegraphics[width=\linewidth]{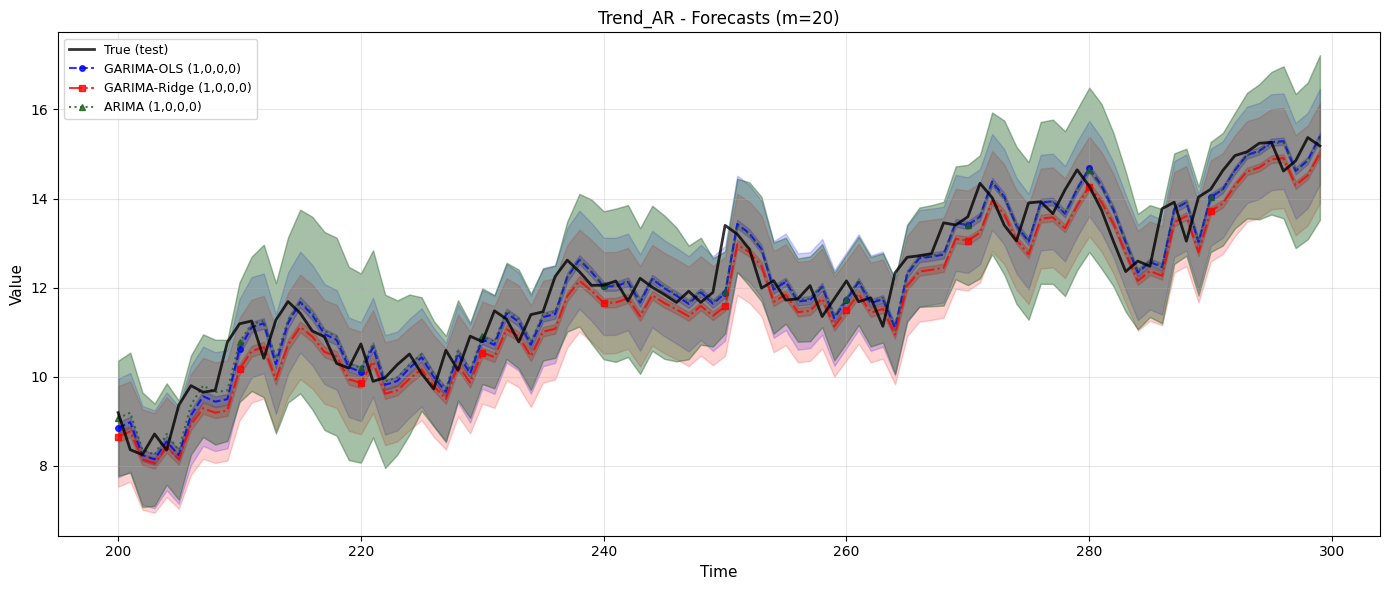}
\caption{Trend plus AR.}
\end{subfigure}

\vspace{0.4em}

\begin{subfigure}{\linewidth}
\centering
\includegraphics[width=\linewidth]{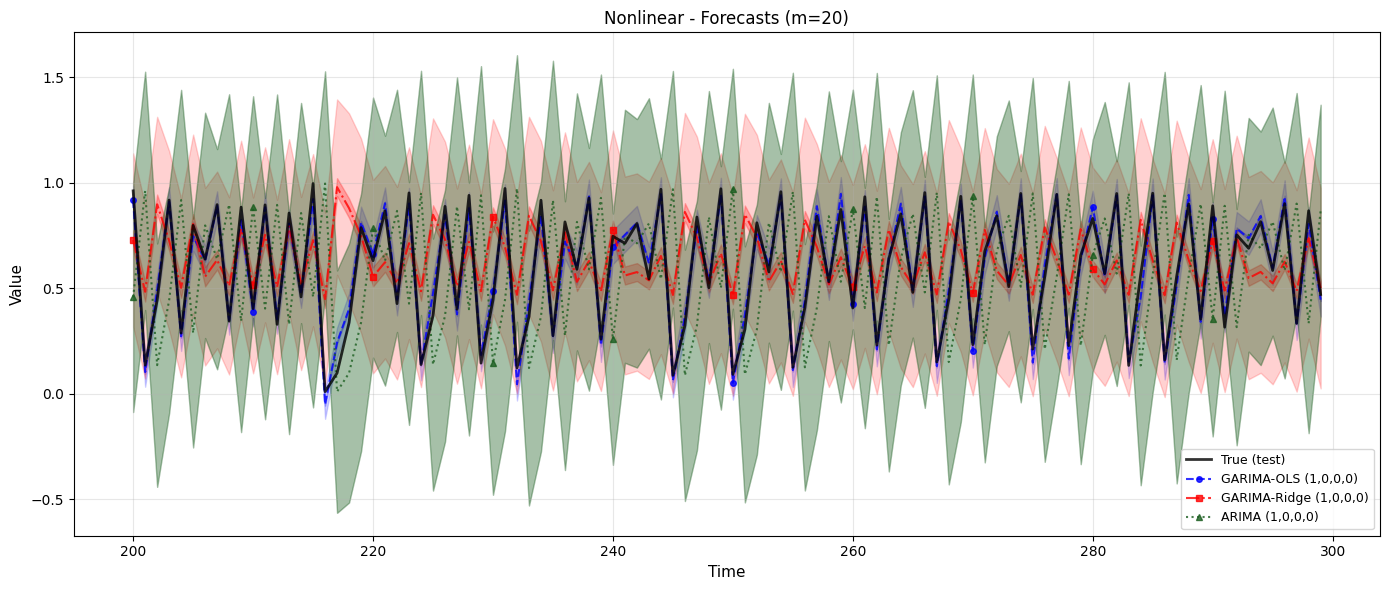}
\caption{Nonlinear recursion.}
\end{subfigure}
\caption{Synthetic series used in the experiments. Part II.}
\label{fig:synthetic_series_2}
\end{figure}

Figures~\ref{fig:synthetic_series_1} and \ref{fig:synthetic_series_2} visualize the four synthetic series used in Section~\ref{sec:synthetic_dgp}. The designs cover a well-specified linear ARMA process, a smooth seasonal signal with period $m=20$, a trending autoregressive process, and a nonlinear recursion. These cases span regimes that are favorable to likelihood-based ARIMA fitting as well as regimes where nonlinear approximation is needed.

\begin{figure}[t]
\centering
\includegraphics[width=\linewidth]{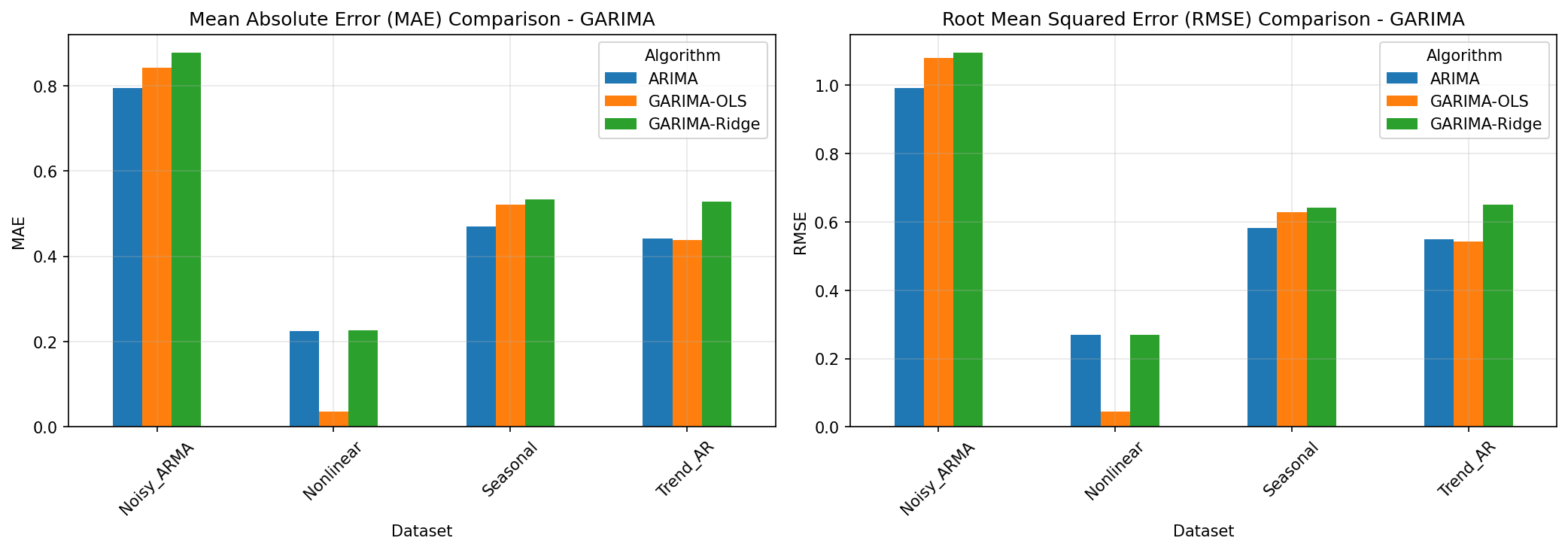}
\caption{Synthetic forecasting comparison under the tuned protocol. Each algorithm selects its order by BIC on the initial window and holds it fixed during rolling one-step-ahead refits. Bars report MAE, RMSE, and total wall-clock time over the rolling horizon.}
\label{fig:synthetic_metrics}
\end{figure}

We evaluate rolling one-step-ahead forecasting using the protocol in Section~\ref{sec:synthetic_protocol}. Figure~\ref{fig:synthetic_metrics} reports the tuned comparison where each algorithm selects its order by BIC on the initial window and then holds the selected order fixed during the rolling evaluation. The results follow a consistent pattern. On the linear Noisy ARMA and the seasonal sine signal, the classical ARIMA baseline achieves the lowest RMSE, which is expected under correct linear specification. On the trending process, Galerkin--SARIMA with unpenalized estimation slightly improves upon ARIMA. The most pronounced gain appears in the nonlinear recursion, where Galerkin--SARIMA substantially reduces forecasting error relative to ARIMA, consistent with the basis-expansion motivation under nonlinear misspecification.

Across all synthetic processes, Galerkin--SARIMA is substantially faster than likelihood-based ARIMA refitting under the same rolling horizon. This reflects the computational advantage of the two-stage closed-form estimation procedure, which avoids repeated nonlinear likelihood optimization at each forecast origin.

\subsection{Ablation on Basis Choice and Basis Size}
\label{sec:synthetic_ablation}

This subsection briefly discusses how performance depends on the approximation class used in the Galerkin projection. In the current experimental suite we fix the basis family across all synthetic runs to avoid introducing additional tuning degrees of freedom. Specifically, we use a linear plus sigmoid basis in both BIC selection and rolling refits, and we standardize the resulting regressors prior to estimation.

Although we do not present a full ablation over basis families and basis dimensions in the main experiments, the role of the approximation class is transparent in the design of the method and in the observed patterns. When the conditional mean is well approximated by a low-order linear ARMA structure, likelihood-based ARIMA achieves the best RMSE and Galerkin--SARIMA remains competitive. When the conditional mean exhibits nonlinear recurrence, the Galerkin basis expansion provides substantial gains, consistent with the interpretation that improvements arise from nonlinear approximation rather than from changes in the rolling protocol or order-selection procedure.

A systematic ablation that varies basis families and basis dimensions would further quantify the accuracy--complexity trade-off and would provide direct guidance for selecting the approximation class in practice. We leave this extended study to future work and focus here on the core tuned comparison under a fixed basis specification.

\subsection{Summary}
\label{sec:synthetic_summary}

The synthetic experiments provide two takeaways. First, in regimes where the data are generated by low-order linear ARMA dynamics or smooth seasonal structure, likelihood-based ARIMA remains a strong baseline and typically achieves the best RMSE within the candidate order grid. Second, when the conditional mean contains trend components or nonlinear recurrence, the Galerkin approximation can reduce forecasting error, with the largest gains appearing in the nonlinear design. Across all synthetic processes, Galerkin--SARIMA delivers large speedups in rolling one-step-ahead refitting, reflecting the advantage of closed-form two-stage estimation over repeated likelihood optimization.


\section{Experiments on Real Data}
\label{sec:real_experiments}

\subsection{Datasets and Preprocessing}
\label{sec:real_data}

We evaluate the proposed methods on three U.S. macroeconomic and financial time series. The S\&P 500 series is obtained from Yahoo Finance. The unemployment rate and real GDP series are obtained from the Federal Reserve Bank of St. Louis FRED database. We apply a uniform preprocessing pipeline across series. We parse dates, sort observations chronologically, remove missing values, and construct a forecasting target suitable for ARIMA-type modeling. \cite{yahoo_finance,fed_fred}

\subsubsection{S\&P 500 index}
\label{sec:sp500_data}

The S\&P 500 data are downloaded from Yahoo Finance as historical index levels with timestamps. Since index levels are nonstationary, we forecast monthly log returns. Let $P_t$ denote the index level at time $t$. For $t\ge 2$ we define
\begin{equation}
y_t \;=\; 100\bigl(\log P_t - \log P_{t-1}\bigr).
\label{eq:sp500_return_def}
\end{equation}
This transformation removes the long-run trend in levels and yields a working target interpretable as a continuously compounded return in percent. \cite{yahoo_finance}

\subsubsection{Unemployment rate}
\label{sec:unrate_data}

The unemployment rate series is obtained from FRED and is measured monthly in percent. We treat the level series as the primary forecasting target after standard cleaning. As a robustness check, we also consider first differences when low-frequency persistence is pronounced. \cite{fed_fred}

\subsubsection{Real GDP}
\label{sec:gdp_data}

The real GDP series is obtained from FRED and is recorded at its native sampling frequency. For forecasting we consider standard stationarity-inducing transformations, including the first difference of log real GDP when needed to remove trend components and stabilize the variance. \cite{fed_fred}
\subsection{Models and Evaluation Protocol}
\label{sec:real_protocol}

We compare three forecasting procedures. The first is a classical ARIMA or seasonal ARIMA model estimated by Gaussian maximum likelihood using the statsmodels implementation. The second is the unpenalized Galerkin--SARIMA estimator. The third is the ridge-regularized Galerkin--SARIMA variant. For Galerkin--SARIMA we use the same basis family as in the synthetic experiments, namely a linear plus sigmoid basis, and we standardize the resulting regressors prior to estimation.

All real-data evaluations use rolling one-step-ahead forecasting with refitting. For each series we fix a rolling calibration window length $W$ and a rolling evaluation horizon $H$. At each forecast origin in the evaluation period, we fit the model on the most recent $W$ observations and produce a one-step-ahead forecast of the next observation. This protocol matches the intended deployment setting where forecasts must be updated repeatedly as new data arrive.

We implement a tuned comparison based on information criteria. For each dataset and each algorithm, we select nonseasonal and seasonal orders on the initial calibration window by minimizing BIC over a common order grid, and we then hold the selected order fixed throughout the rolling evaluation. The model is refit at each forecast origin using the fixed selected order. This design avoids manual tuning and makes the comparison reproducible.

We summarize forecasting accuracy using MAE and RMSE over the rolling horizon. We also report total wall-clock time required to complete all rolling refits and forecasts for each algorithm, together with throughput in rolling iterations per second.

For interval reporting we distinguish confidence intervals for the conditional mean forecast from prediction intervals for future observations. The statsmodels forecast interval output is reported as a confidence interval for the conditional mean and is labeled as Mean CI in all tables and figures. For Galerkin--SARIMA we report both Mean CI and prediction intervals using the block bootstrap procedure in Section~\ref{sec:prediction_intervals}.

\subsection{Forecasting Results}
\label{sec:real_results}

\begin{figure}[t]
\centering
\includegraphics[width=\linewidth]{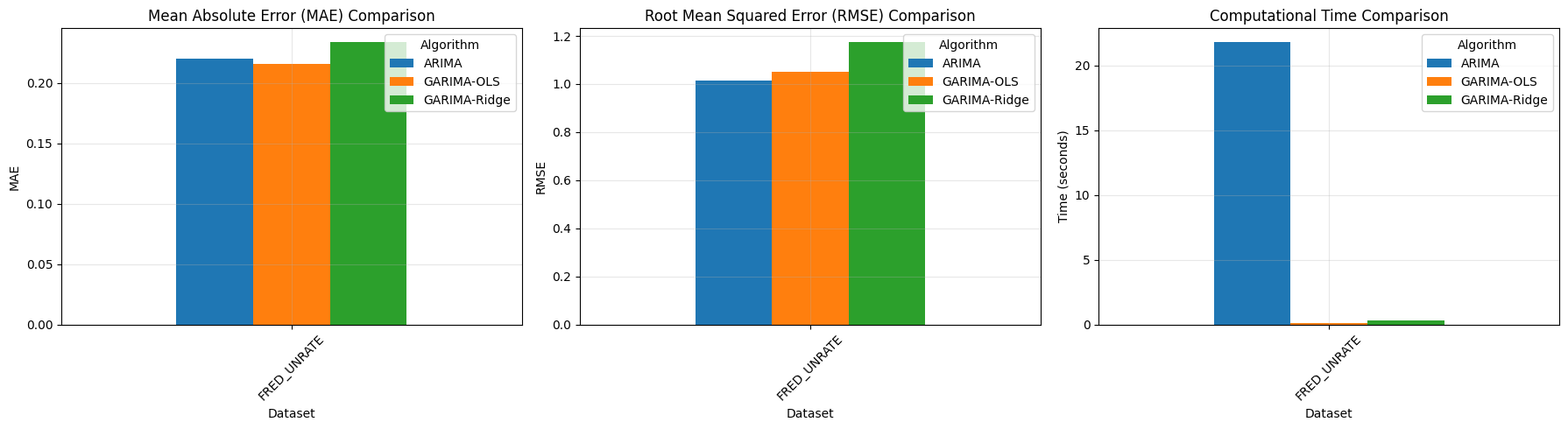}
\caption{Unemployment rate. Algorithm comparison under the tuned rolling protocol.}
\label{fig:unrate_comp}
\end{figure}

\begin{figure}[t]
\centering
\includegraphics[width=\linewidth]{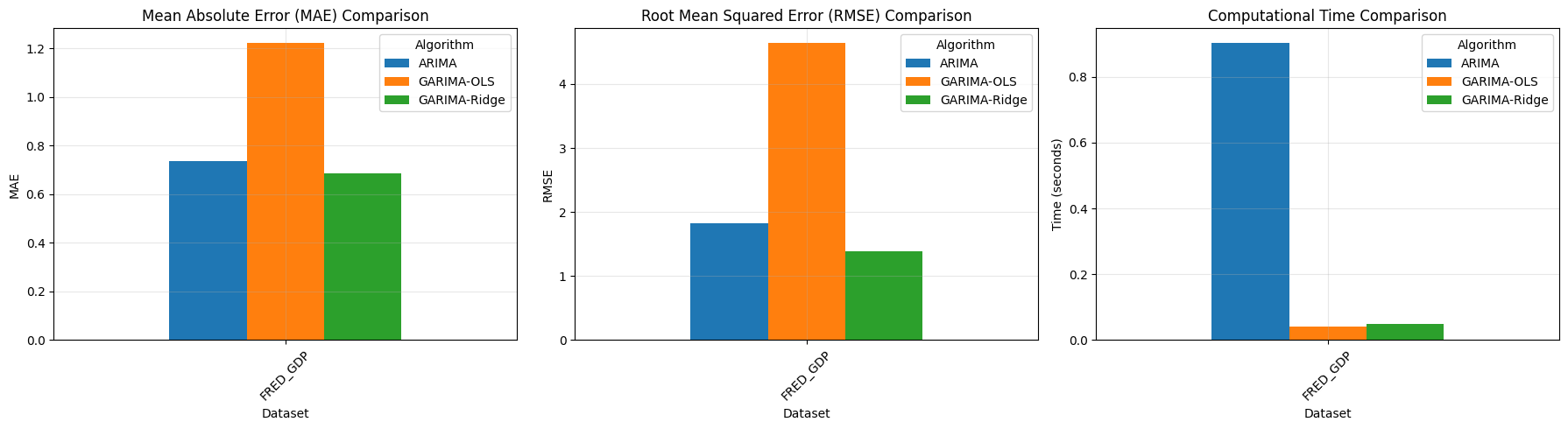}
\caption{Real GDP. Algorithm comparison under the tuned rolling protocol.}
\label{fig:gdp_comp}
\end{figure}

\begin{figure}[t]
\centering
\includegraphics[width=\linewidth]{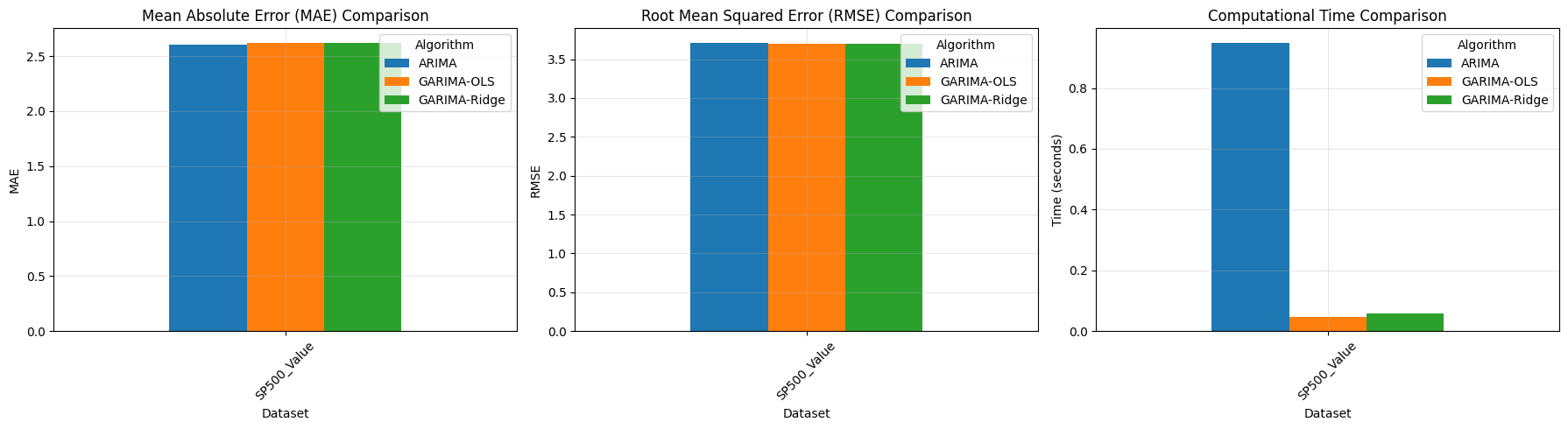}
\caption{S\&P 500 returns. Algorithm comparison under the tuned rolling protocol.}
\label{fig:sp_comp}
\end{figure}

\begin{figure}[t]
\centering
\includegraphics[width=\linewidth]{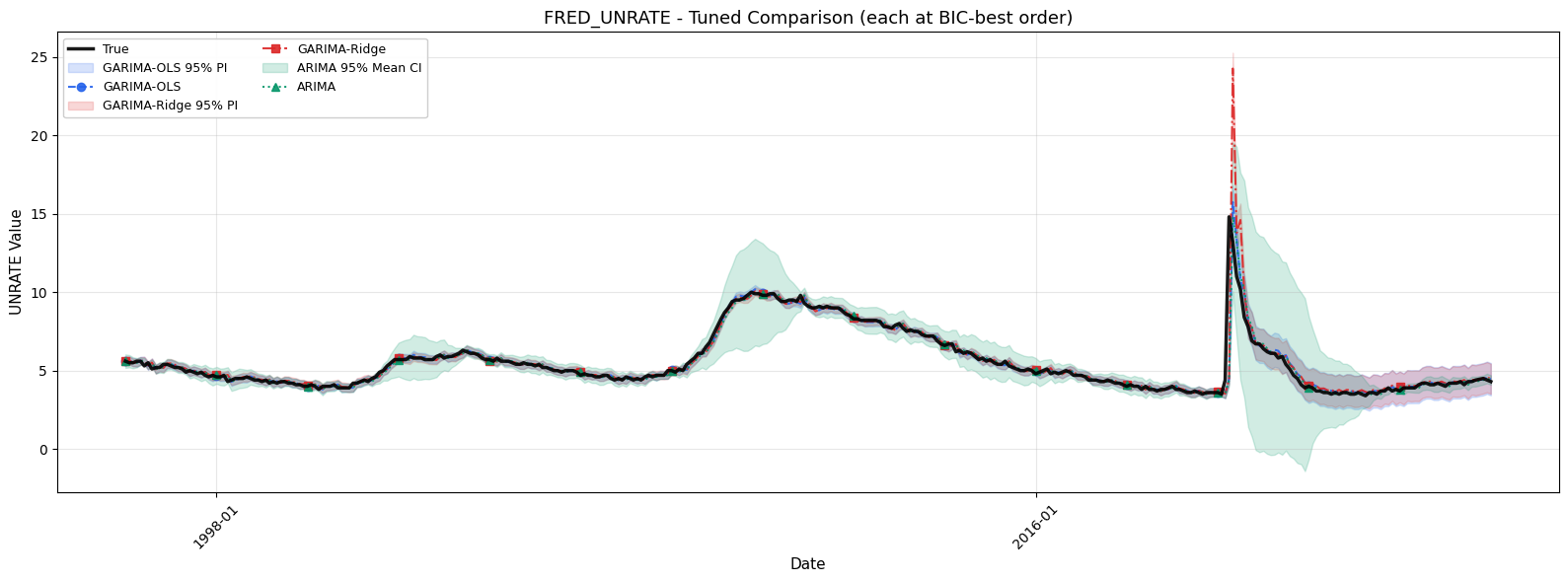}
\caption{Unemployment rate. One-step-ahead rolling forecasts under the tuned protocol.}
\label{fig:unrate_pred}
\end{figure}

\begin{figure}[t]
\centering
\includegraphics[width=\linewidth]{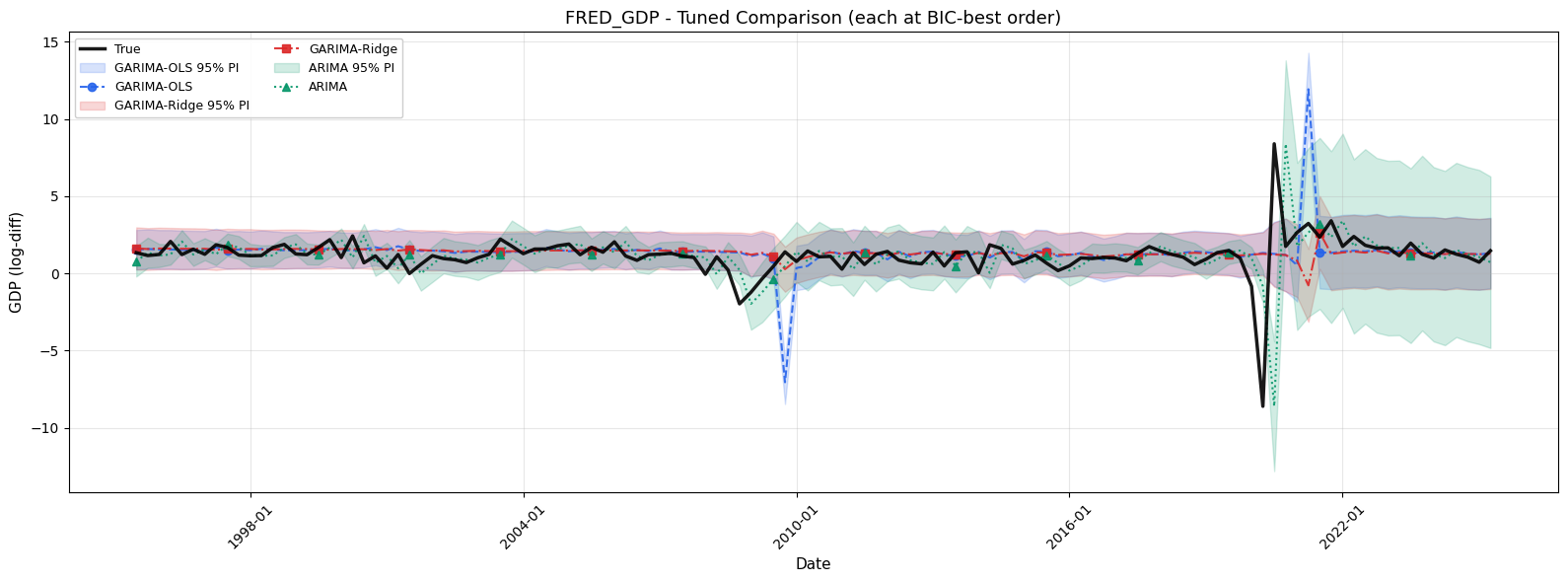}
\caption{Real GDP. One-step-ahead rolling forecasts under the tuned protocol.}
\label{fig:gdp_pred}
\end{figure}

\begin{figure}[t]
\centering
\includegraphics[width=\linewidth]{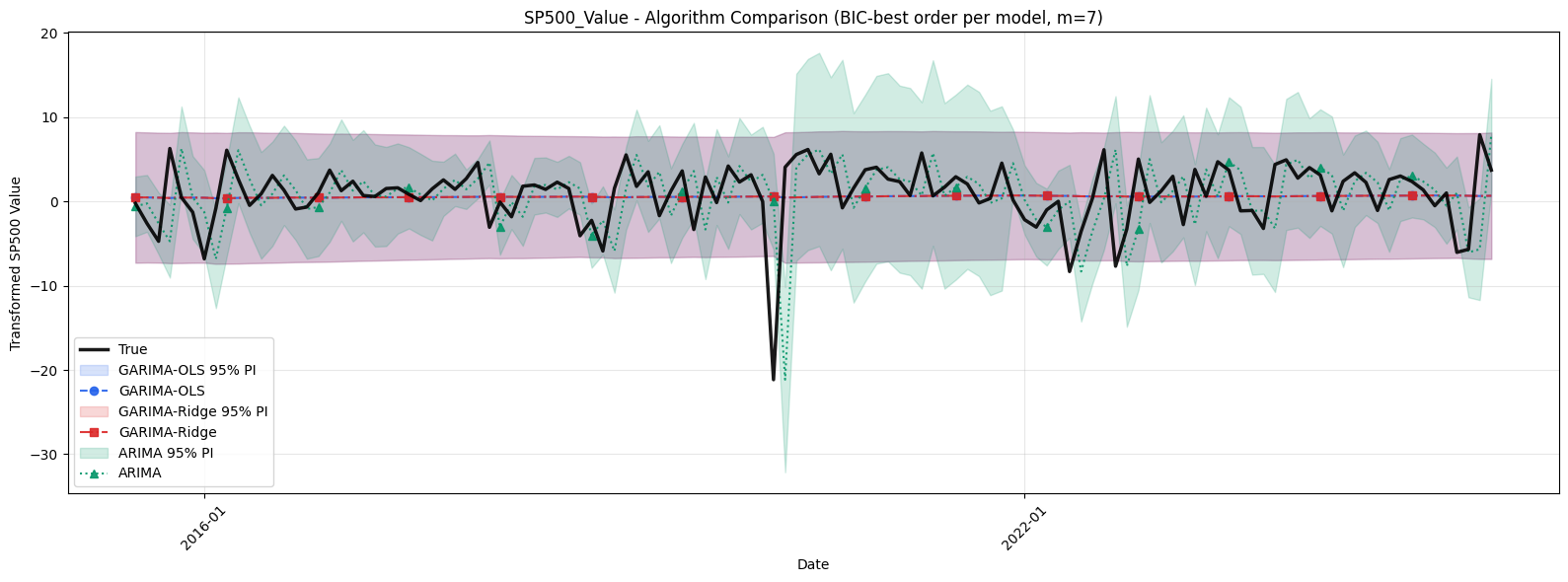}
\caption{S\&P 500 returns. One-step-ahead rolling forecasts under the tuned protocol.}
\label{fig:sp_pred}
\end{figure}

We report tuned rolling one-step-ahead forecasting performance on three real series under the protocol in Section~\ref{sec:real_protocol}. For each dataset, each algorithm selects its order by BIC on the initial calibration window and then holds the selected order fixed during the rolling evaluation. Table~\ref{tab:real_accuracy_runtime} summarizes MAE, RMSE, and total wall-clock time required to complete the rolling refits and forecasts. Figures~\ref{fig:unrate_comp}--\ref{fig:sp_comp} visualize the headline accuracy and runtime comparison, and Figures~\ref{fig:unrate_pred}--\ref{fig:sp_pred} show representative rolling forecast paths.

On the unemployment rate series, ARIMA achieves the lowest RMSE, while Galerkin--SARIMA with unpenalized estimation is close in RMSE and slightly better in MAE. This combination is informative. MAE reflects typical errors, while RMSE puts more weight on occasional large misses. The fact that Galerkin--SARIMA improves MAE but loses RMSE suggests that its forecasts are slightly closer to the realized series most of the time, but it produces a small number of larger deviations that dominate RMSE. In this experiment the rolling horizon is longer than in the other datasets, so total wall-clock time is not directly comparable across datasets without accounting for the number of rolling refits. Even under the longer horizon, the computational advantage of Galerkin--SARIMA remains large. The average time per rolling refit is about $6.1\times 10^{-2}$ seconds for ARIMA and about $3.8\times 10^{-4}$ seconds for Galerkin--SARIMA with unpenalized estimation.

On the real GDP series, the ridge-regularized Galerkin--SARIMA variant achieves the best accuracy in both MAE and RMSE. Relative to ARIMA, the RMSE improves substantially while runtime remains much smaller. This pattern is consistent with the role of regularization in the Galerkin estimator. The basis expansion can capture nonlinear response and saturation effects that are not well represented by a low-order linear ARIMA specification, but the expanded design can also be ill-conditioned in finite samples, especially under short rolling windows. Ridge regularization stabilizes the two-stage regression and reduces the impact of collinearity among basis terms. The contrast between the ridge and unpenalized Galerkin estimates on GDP is sharp and indicates that regularization can be important for macroeconomic series.

On the S\&P 500 return series, all three methods achieve very similar RMSE. This outcome is expected in a weak-signal setting. One-step-ahead forecasting of equity returns is dominated by noise, and a wide class of models will produce forecasts close to a low-variance baseline. In this regime the main contribution of Galerkin--SARIMA is computational. It matches ARIMA accuracy up to small differences while completing rolling refits in a small fraction of the time.

\begin{table}[t]
\caption{Real-data rolling one-step-ahead forecasting results under the tuned protocol. Each algorithm selects its order by BIC on the initial calibration window and holds it fixed during rolling refits. Time is total wall-clock seconds over the rolling horizon.}
\label{tab:real_accuracy_runtime}
\centering
\begin{tabular}{l l r r r}
\toprule
Dataset & Algorithm & MAE & RMSE & Time \\
\midrule
Unemployment rate & ARIMA         & 0.2203 & 1.0145 & 21.7974 \\
Unemployment rate & GARIMA--OLS   & 0.2160 & 1.0519 & 0.1350 \\
Unemployment rate & GARIMA--Ridge & 0.2336 & 1.1730 & 0.3024 \\
\addlinespace
Real GDP           & ARIMA         & 0.7377 & 1.8186 & 0.9024 \\
Real GDP           & GARIMA--OLS   & 1.2223 & 4.6379 & 0.0425 \\
Real GDP           & GARIMA--Ridge & 0.6855 & 1.3911 & 0.0489 \\
\addlinespace
S\&P 500 returns    & ARIMA         & 2.6031 & 3.7116 & 0.9493 \\
S\&P 500 returns    & GARIMA--OLS   & 2.6218 & 3.7023 & 0.0472 \\
S\&P 500 returns    & GARIMA--Ridge & 2.6218 & 3.7023 & 0.0567 \\
\bottomrule
\end{tabular}
\end{table}

The interval diagnostics are reported separately to avoid conflating conditional-mean confidence intervals with prediction intervals. In particular, statsmodels interval outputs are reported as Mean CI, while Galerkin--SARIMA additionally supports prediction intervals via the block bootstrap procedure in Section~\ref{sec:prediction_intervals}.

To clarify the behavior of the tuned comparison, we summarize the BIC-selected structures and the main diagnostic signals observed in the fitted models. The selected orders differ substantially across datasets and help explain why accuracy can be similar in some cases while computational cost differs sharply.

For S\&P 500 returns, both Galerkin estimators select a specification with no AR or MA lags and no seasonal component. Under this selection the fitted conditional mean is close to a constant baseline, which is consistent with weak short-horizon return predictability. In contrast, ARIMA still selects a low-order moving-average specification and reports strong non-Gaussian residual diagnostics, indicating heavy tails in the return innovations.

For unemployment and GDP, the tuned selections involve dataset-specific seasonal periods and higher-order linear dynamics. These series contain stronger low-frequency structure than returns, and the ridge penalty can change the effective degrees of freedom when the expanded Galerkin design is ill-conditioned. On GDP in particular, the gap between ridge and unpenalized Galerkin performance is consistent with finite-sample instability in the unpenalized expanded design, where regularization improves numerical conditioning and reduces variance amplification.

We report the selected orders for reference in Table~\ref{tab:real_selected_orders}.

\begin{table}[t]
\caption{BIC-selected orders under the tuned protocol. Orders are reported as nonseasonal $p,q$ and seasonal $P,Q$ with the dataset-specific seasonal period $m$.}
\label{tab:real_selected_orders}
\centering
\begin{tabular}{l l r r r r r}
\toprule
Dataset & Algorithm & $p$ & $q$ & $P$ & $Q$ & $m$ \\
\midrule
Unemployment rate & GARIMA--OLS   & 1 & 0 & 1 & 0 & 12 \\
Unemployment rate & GARIMA--Ridge & 1 & 3 & 0 & 0 & 12 \\
Unemployment rate & ARIMA         & 2 & 2 & 0 & 0 & 12 \\
\addlinespace
Real GDP           & GARIMA--OLS   & 0 & 0 & 1 & 0 & 4 \\
Real GDP           & GARIMA--Ridge & 0 & 0 & 1 & 0 & 4 \\
Real GDP           & ARIMA         & 1 & 0 & 0 & 0 & 4 \\
\addlinespace
S\&P 500 returns    & GARIMA--OLS   & 0 & 0 & 0 & 0 & 30 \\
S\&P 500 returns    & GARIMA--Ridge & 0 & 0 & 0 & 0 & 30 \\
S\&P 500 returns    & ARIMA         & 0 & 1 & 0 & 0 & 30 \\
\bottomrule
\end{tabular}
\end{table}
\subsection{Ridge vs Unpenalized on Real Series}
\label{sec:real_ridge}

This subsection isolates the role of ridge regularization in the Galerkin estimator by comparing GARIMA--OLS and GARIMA--Ridge under the same tuned rolling protocol. The comparison clarifies when regularization improves numerical stability and when it changes the effective model selection outcome.

On real GDP, ridge regularization is essential. GARIMA--OLS and GARIMA--Ridge select the same seasonal structure under BIC, yet their out-of-sample accuracy differs sharply. This indicates that the improvement is not driven by a different selected order but by the stability of estimation in the expanded Galerkin design. In short rolling windows, basis-expanded regressors can be strongly correlated, and the unpenalized two-stage regression can amplify variance. Ridge regularization reduces effective degrees of freedom and improves conditioning, which translates into substantially better one-step-ahead accuracy on this macroeconomic series.

On the unemployment rate, ridge regularization does not help under the current fixed hyperparameters. The ridge penalty alters the BIC landscape and leads to a different selected specification relative to the unpenalized estimator. Under this tuned selection, GARIMA--Ridge yields higher MAE and RMSE than GARIMA--OLS and ARIMA. This pattern is consistent with excess shrinkage and specification shifts that increase bias relative to the modest variance reduction achieved on this series.

On S\&P 500 returns, GARIMA--OLS and GARIMA--Ridge are essentially tied in MAE and RMSE. In this setting BIC selects a very low-complexity Galerkin specification that collapses to a constant baseline, reflecting weak short-horizon return predictability. When the selected forecasting rule is already close to a low-variance baseline, additional ridge shrinkage has little impact on point forecasts.

Overall, the real-data evidence suggests that ridge regularization is not uniformly beneficial, but it can be decisive in settings where the basis-expanded Galerkin design is ill-conditioned. In the current experiments we fix ridge hyperparameters across datasets to avoid introducing additional tuning degrees of freedom. A systematic calibration of the ridge penalty within the initial window, for example via a small validation grid, is a natural extension.

\subsection{Interval Results on Real Series}
\label{sec:real_intervals}

This subsection reports interval diagnostics on the real-data experiments. The goal is to assess uncertainty quantification in the rolling one-step-ahead setting using empirical coverage and average interval width over the evaluation horizon. We emphasize a key comparability point. For statsmodels ARIMA, the reported interval is a confidence interval for the conditional mean forecast, which we refer to as Mean CI. For Galerkin--SARIMA, we report prediction intervals for the next observation constructed by the block bootstrap procedure in Section~\ref{sec:prediction_intervals}, which we refer to as PI. Because these objects target different quantities, the interval results should be interpreted as diagnostics within each method rather than as a strict like-for-like comparison.

Table~\ref{tab:real_interval_quality} reports coverage and mean width at the nominal 95\% level. On GDP, both Galerkin variants attain coverage close to the nominal level with widths comparable to or slightly smaller than ARIMA. This is consistent with the stronger performance of ridge on point forecasts and suggests that the bootstrap PI is well-calibrated in this setting. On unemployment, ARIMA attains very high coverage with comparatively wide intervals, while Galerkin intervals are narrower and closer to nominal coverage. This pattern is consistent with a conservative Mean CI under ARIMA and a tighter PI under the bootstrap procedure. On S\&P 500 returns, the Galerkin intervals achieve very high coverage but with wider intervals than ARIMA. This is consistent with weak predictability and heavy-tailed shocks, where conservative uncertainty quantification can be preferable.

\begin{table}[t]
\caption{Interval quality on real data. Coverage is the fraction of realized observations covered by a nominal 95\% interval over the rolling evaluation horizon. Width is the average interval width. ARIMA reports Mean CI for the conditional mean. Galerkin methods report prediction intervals from block bootstrap.}
\label{tab:real_interval_quality}
\centering
\begin{tabular}{l l r r}
\toprule
Dataset & Algorithm & Coverage (\%) & Mean width \\
\midrule
Real GDP & ARIMA         & 86.67 & 3.66 \\
Real GDP & GARIMA--OLS   & 90.83 & 2.92 \\
Real GDP & GARIMA--Ridge & 91.67 & 3.00 \\
\addlinespace
Unemployment rate & ARIMA         & 98.06 & 2.11 \\
Unemployment rate & GARIMA--OLS   & 91.94 & 0.82 \\
Unemployment rate & GARIMA--Ridge & 94.17 & 0.84 \\
\addlinespace
S\&P 500 returns & ARIMA         & 83.33 & 12.68 \\
S\&P 500 returns & GARIMA--OLS   & 97.50 & 15.00 \\
S\&P 500 returns & GARIMA--Ridge & 97.50 & 15.00 \\
\bottomrule
\end{tabular}
\end{table}

Additional residual and stability diagnostics are reported in Appendix~\ref{app:model_diagnostics}.

\subsection{Practical Considerations}
\label{sec:real_practical}

Two practical issues affect how the real-data results should be read. The first is that runtime is reported as total wall-clock time over the rolling horizon, and the horizons differ across datasets. For this reason, comparisons of time across datasets are not meaningful without normalization by the number of rolling refits. The within-dataset comparison remains informative and consistently shows large computational savings for Galerkin--SARIMA relative to ARIMA in the rolling refit setting.

The second issue is that regularization interacts with model selection. Ridge penalties change the effective likelihood landscape and can shift the BIC-selected order, as observed in the unemployment experiment. In the present experiments ridge hyperparameters are selected by generalized cross-validation on the initial window and then held fixed across rolling refits; the weighting scheme is fixed across datasets to limit additional tuning degrees of freedom. Alternative calibration strategies for the ridge penalty are a natural extension, but they are not required to establish the main empirical pattern. Galerkin--SARIMA matches or improves point forecast accuracy on key macroeconomic series while remaining substantially faster to refit.
\section{Discussion and Future Work}

This paper proposes Galerkin--ARIMA (and its seasonal extension) as a projection-based reparameterization of the classical ARIMA/SARIMA forecasting problem. The key perspective is that one-step-ahead forecasting can be cast as estimating conditional mean and innovation maps over a low-dimensional lag state, and that the traditional linear specification is a restrictive choice within a broader class of smooth functionals. By replacing linear regressors with a controlled Galerkin basis expansion and exploiting a two-stage least-squares structure, the method preserves the operational appeal of ARIMA---interpretability, modularity, and rolling deployment---while enabling fast refits and improved approximation when the true forecasting rule departs from strict linearity.

A central message is that the statistical role of Galerkin approximation and the computational role of two-stage estimation are tightly linked. On the statistical side, the basis expansion induces an approximation--estimation tradeoff familiar from sieve methods: as basis sizes grow, the sieve bias shrinks at a rate governed by smoothness and state dimension, while sampling error grows with the effective model dimension. Our theoretical results formalize this tradeoff in a time-series setting and provide conditions under which the unpenalized estimator attains oracle-style behavior in linear regimes and favorable risk comparisons relative to classical SARIMA under smooth nonlinear truth. On the computational side, the two-stage structure replaces repeated nonlinear likelihood optimization with linear algebra operations on design matrices that can be updated efficiently in rolling windows, yielding large wall-clock savings in regimes where frequent refitting is required. Taken together, the approach illustrates that improving approximation flexibility need not come at the cost of deployment latency, provided the estimation architecture is aligned with the projection structure.

The empirical evidence should be read through the lens of this approximation viewpoint. When the target series exhibits dynamics that are well captured by low-order linear ARIMA, the Galerkin method is not expected to deliver large accuracy gains, and its main benefit is computational. In contrast, when the conditional mean or innovation map is smooth but nonlinear in lag states---or when seasonal structure interacts with local nonlinearities---the sieve approximation can reduce systematic forecast error at fixed horizon. At the same time, the basis-expanded design can become ill-conditioned in finite samples, especially under strong persistence, high seasonal order, or when basis sizes are chosen aggressively. Ridge regularization addresses this instability by shrinking the effective degrees of freedom and stabilizing rolling refits, but it also interacts with model selection and can shift BIC-chosen orders. The real-data results reflect these forces: regularization is not uniformly beneficial, but it can be decisive when the expanded design is close to singular or when the rolling window is short relative to the expanded dimension.

Several limitations highlight directions for further development. First, while the projection framework naturally accommodates rich nonlinearities, the state dimension still matters: sieve rates deteriorate as the number of lags increases, so the method is best suited to settings where predictive content is concentrated in a modest number of lags and seasonal components. This motivates more structured bases (e.g., additive or partially linear constructions) that exploit sparsity or low-order interactions in lag states. Second, uncertainty quantification remains delicate in rolling, one-step-ahead environments. Our block bootstrap procedure provides a practical route to prediction intervals under weak dependence, but sharper theory for interval validity under sieve approximation and two-stage estimation---including the interaction with ridge regularization and data-dependent basis selection---would strengthen the inferential foundations of the approach. Third, the current analysis emphasizes one-step-ahead forecasting, where the projection architecture is most transparent. Extending the framework to multi-step forecasts requires careful handling of iterated prediction and the propagation of approximation error, and may benefit from direct multi-horizon objectives or state augmentation.

Beyond these methodological refinements, there are several natural extensions that are immediately relevant to econometric practice. The first is multivariate forecasting: many applications involve panels of related series, suggesting a vector version of the Galerkin projection with shared basis components, shrinkage across series, or factor-augmented states. The second is the incorporation of exogenous information. Although this paper focuses on univariate SARIMA structure, the projection viewpoint extends directly to conditional mean functionals that depend on both lag states and exogenous covariates; the main additional challenge is controlling basis growth and dependence when covariates are persistent or high-dimensional. The third is adaptive complexity control. While BIC provides a transparent selection rule and enjoys desirable consistency properties under the oracle regime, alternative criteria tailored to rolling forecasting loss---or selection rules that explicitly account for ill-conditioning and effective degrees of freedom under ridge---may yield more robust performance in finite samples. Finally, the computational structure suggests opportunities for online and streaming implementations, where sufficient statistics of the design matrices are updated as new observations arrive, further reducing latency in large-scale forecasting systems.

Overall, Galerkin--ARIMA provides a bridge between classical parametric time-series modeling and sieve-based regression methods. It retains the modular decomposition and interpretability of ARIMA while enabling controlled nonlinear flexibility, accompanied by theoretical guarantees and substantial computational savings in rolling refit settings. These features make the approach a promising building block for forecasting problems where speed, transparency, and mild departures from linear dynamics are all empirically relevant.


\section{Conclusion}

This paper develops Galerkin--ARIMA and Galerkin--SARIMA as projection-based reparameterizations of classical ARIMA/SARIMA forecasting. The main contribution is to show that one-step-ahead forecasting can be treated as estimating smooth conditional mean and innovation maps over a low-dimensional lag state, and that replacing the rigid linear specification with a controlled Galerkin basis expansion yields a method that is both flexible and deployable in rolling settings. The two-stage least-squares architecture is essential: it aligns estimation with the projection structure, avoids repeated nonlinear likelihood optimization, and therefore delivers substantial computational gains when models must be refit frequently.

On the statistical side, the analysis formalizes the approximation--estimation tradeoff induced by sieve expansions in dependent data, and establishes regimes in which the Galerkin estimator matches the classical benchmark while improving performance under smooth nonlinear truth. In particular, the results provide (i) an oracle-type guarantee in linear regimes, (ii) consistency and asymptotic distributional approximations for one-step-ahead forecasts under controlled basis growth, and (iii) a risk comparison showing that the projection approach can dominate classical SARIMA in mean-squared forecasting loss when nonlinearities are present but sufficiently smooth. On the empirical side, experiments on synthetic and real series support the theoretical message: Galerkin--SARIMA achieves comparable or improved point forecast accuracy relative to standard ARIMA while remaining substantially faster to refit in rolling windows, and ridge regularization can be decisive when the basis-expanded design is ill-conditioned.

Overall, Galerkin--ARIMA provides an econometrically grounded bridge between classical parametric time-series modeling and sieve-based regression methods. It preserves the interpretability and modular decomposition of ARIMA while allowing controlled departures from linearity, accompanied by explicit theoretical guarantees and practical computational advantages. These features make the approach a useful building block for latency-sensitive forecasting problems in macroeconomics and finance, and motivate further work on structured bases, multivariate extensions, and sharper uncertainty quantification for rolling prediction.

\newpage
\bibliographystyle{plainnat}
\bibliography{references}

\newpage
\appendix

\section{Model Diagnostics}
\label{app:model_diagnostics}

We report three standard diagnostic checks for the fitted models in the real-data
experiments. First, we compute Ljung--Box $Q$-statistics at multiple lags on
one-step-ahead residuals to assess remaining serial correlation. Second, we
assess residual normality using Jarque--Bera statistics and corresponding
quantile-quantile summaries. Third, we assess rolling-window stability by
tracking coefficient paths and selected orders across forecast origins and
reporting CUSUM-style stability indicators for the residuals. These diagnostics
are computed on the rolling refits and are interpreted as model adequacy checks
rather than formal specification tests.

\section{Proofs of Theoretical Results}
\label{sec:proofs}

\subsection{Additional notation and assumptions}
\label{sec:proof_notation}

We collect notation used in the proofs.

Let
\[
t_0 = \max\{p,\,Pm,\,q,\,Qm\}
\]
and define the effective sample size
\[
N = T - t_0 .
\]

For each time index \(t = t_0,\dots,T-1\) define the lag vectors
\[
x_t
=
\bigl(y_{t-1}^{(d,D)},\dots,y_{t-p}^{(d,D)}\bigr)\in\mathbb R^{p},
\]
\[
x_t^{(s)}
=
\bigl(y_{t-m}^{(d,D)},\dots,y_{t-Pm}^{(d,D)}\bigr)\in\mathbb R^{P},
\]
and residual lag vectors
\[
r_t
=
\bigl(\epsilon_{t-1},\dots,\epsilon_{t-q}\bigr)\in\mathbb R^{q},
\qquad
r_t^{(s)}
=
\bigl(\epsilon_{t-m},\dots,\epsilon_{t-Qm}\bigr)\in\mathbb R^{Q}.
\]

Let
\[
Y
=
\bigl(y_{t_0}^{(d,D)},\dots,y_{T-1}^{(d,D)}\bigr)^{\top}
\in\mathbb R^{N},
\qquad
\epsilon
=
\bigl(\epsilon_{t_0},\dots,\epsilon_{T-1}\bigr)^{\top}
\in\mathbb R^{N}.
\]

The design matrices are
\[
\Phi \in\mathbb R^{N\times K},
\quad
\Phi^{(s)}\in\mathbb R^{N\times K_s},
\quad
\Psi\in\mathbb R^{N\times L},
\quad
\Psi^{(s)}\in\mathbb R^{N\times L_s},
\]
where the \(t\)-th row of each matrix is
\[
\Phi(x_t)^{\top},
\quad
\Phi^{(s)}(x_t^{(s)})^{\top},
\quad
\Psi(r_t)^{\top},
\quad
\Psi^{(s)}(r_t^{(s)})^{\top}
\]
respectively.

Stack these into the full design matrix
\[
\Psi_N
=
\bigl[\,
  \Phi \;\;\; \Phi^{(s)} \;\;\; \Psi \;\;\; \Psi^{(s)}
\bigr]
\in\mathbb R^{N\times d_G},
\qquad
d_G = K + K_s + L + L_s .
\]

The population Galerkin coefficients are
\[
\gamma^{\star}
=
\bigl(
  \beta^{\star},
  \beta^{(s)\star},
  \alpha^{\star},
  \alpha^{(s)\star}
\bigr)\in\mathbb R^{d_G},
\]
defined as the unique solutions of the population projection equations
for \(f,f^{(s)},g,g^{(s)}\). The empirical two–stage least–squares
estimator is
\[
\hat\gamma
=
\bigl(
  \hat\beta,
  \hat\beta^{(s)},
  \hat\alpha,
  \hat\alpha^{(s)}
\bigr).
\]

Assumptions:
\begin{itemize}
\item[(i)] The differenced process \(\{y_t^{(d,D)}\}\) is generated by a
stable, invertible Gaussian SARIMA model with iid innovations
\(\epsilon_t\sim\mathcal N(0,\sigma^2)\) and absolutely summable AR and MA
coefficients.

\item[(ii)] The vector processes \(\{x_t\},\{x_t^{(s)}\},\{r_t\},\{r_t^{(s)}\}\)
are strictly stationary, ergodic and \(\beta\)-mixing with
$\sum_{m\ge 1}\beta(m)^{1/2}<\infty$.

\item[(iii)] The conditional mean and innovation maps
\(f,f^{(s)},g,g^{(s)}\) are Hölder–smooth of order \(r>0\) on compact
domains as in Assumption (A2).

\item[(iv)] The eigenvalue and basis growth conditions in Assumptions
(A3)–(A4) hold:
\[
c_0 \le
\lambda_{\min}\bigl(N^{-1}\Psi_N^{\top}\Psi_N\bigr)
\le
\lambda_{\max}\bigl(N^{-1}\Psi_N^{\top}\Psi_N\bigr)
\le c_1,
\]
and \(K,K_s,L,L_s\to\infty\) with \(K+K_s+L+L_s = O(N)\).
\end{itemize}

\subsection{Proof of approximation lemma}
\label{app:lemma_jackson_proof}

We prove the Jackson–type bounds for the basis approximations.

\begin{lemma*}[Lemma \ref{lem:jackson}, restated]
Assume (A2). Let \(\Phi,\Phi^{(s)},\Psi,\Psi^{(s)}\) be tensor–product
polynomial or spline bases with sizes \(K,K_s,L,L_s\) for the domains
of \(X,X^{(s)},R,R^{(s)}\) of dimensions \(p,P,q,Q\). Then there exist
constants \(C_1,\dots,C_4\) not depending on \(K,K_s,L,L_s\) such that
\[
\Delta_f \le C_1 K^{-r/p},
\quad
\Delta_{f^{(s)}} \le C_2 K_s^{-r/P},
\quad
\Delta_g \le C_3 L^{-r/q},
\quad
\Delta_{g^{(s)}} \le C_4 L_s^{-r/Q}.
\]
\end{lemma*}

\begin{proof}
We prove the bound for \(\Delta_f\). The remaining three cases are
identical with the appropriate dimensions and basis sizes.

Let the domain of the lag vector \(X\) be a compact hyper–rectangle in
\(\mathbb R^{p}\). Apply an affine change of variables so that the
domain becomes the unit cube
\[
[0,1]^p.
\]
This does not affect Hölder exponents or big–O orders.

Fix a cell side length \(h \in (0,1]\). Partition \([0,1]^p\) into
hyper–cubes of side length \(h\). Then the number of cells is
\[
M = h^{-p}.
\]
For each cell \(C\) choose a representative point \(x_C \in C\), for
example the lower–left corner.

Assumption (A2) says that \(f\) is Hölder–continuous of order \(r>0\):
there exists a constant \(L_f\) such that for all \(u,v\in[0,1]^p\),
\[
|f(u) - f(v)| \le L_f \|u - v\|_{\infty}^{r}.
\]
Now take any cell \(C\) and any \(x\in C\). Then
\[
\|x - x_C\|_{\infty} \le h,
\]
so
\[
|f(x) - f(x_C)|
\le
L_f \|x - x_C\|_{\infty}^{r}
\le
L_f h^{r}.
\]

Define a function \(f_h:[0,1]^p\to\mathbb R\) by
\[
f_h(x) = f(x_C)
\quad\text{whenever } x\in C.
\]
Then for any \(x\in[0,1]^p\),
\[
|f(x) - f_h(x)|
=
|f(x) - f(x_C)|
\le
L_f h^{r}.
\]
Hence
\[
\|f - f_h\|_{\infty}
=
\sup_{x\in[0,1]^p} |f(x) - f_h(x)|
\le L_f h^{r}.
\]

Take \(\Phi\) to be a tensor–product spline or polynomial basis adapted
to the same grid. There exists a coefficient vector \(\beta^{\star}\)
such that \(\Phi(x)^{\top}\beta^{\star}\) exactly reproduces any
piecewise polynomial (of appropriate degree) or piecewise constant
function on the grid cells. In particular there is \(\beta^{\star}\)
such that
\[
\Phi(x)^{\top}\beta^{\star} = f_h(x)
\quad
\text{for all }x\in[0,1]^p.
\]
Then
\[
\sup_{x}
\bigl|f(x) - \Phi(x)^{\top}\beta^{\star}\bigr|
=
\sup_{x}
|f(x) - f_h(x)|
\le
L_f h^{r}.
\]

For tensor–product grids the number of basis functions \(K\) is
proportional to the number of cells \(M\):
\[
K \asymp M = h^{-p}.
\]
Thus there exists a constant \(c>0\) such that for small \(h\),
\[
K \ge c h^{-p}
\quad\Rightarrow\quad
h \le (K/c)^{-1/p}
=
c^{1/p} K^{-1/p}.
\]
Hence there are constants \(C,C'>0\) such that
\[
h^{r}
\le
C K^{-r/p},
\qquad
L_f h^{r}
\le
C' K^{-r/p}.
\]

\[
\Delta_f
=
\inf_{\beta}
\sup_x |f(x) - \Phi(x)^{\top}\beta|
\le
\sup_x |f(x) - \Phi(x)^{\top}\beta^{\star}|
\le
C' K^{-r/p}.
\]
Set \(C_1 = C'\). This proves \(\Delta_f \le C_1 K^{-r/p}\).

The same argument, with dimensions \(P,q,Q\) and basis sizes
\(K_s,L,L_s\), yields the three remaining inequalities.
\end{proof}

\subsection{Proof of consistency theorem}
\label{app:proof_unbiased}

We prove consistency of the one–step–ahead forecast.

\begin{theorem*}[Theorem \ref{thm:consistency}, restated]
Under Assumptions (A1)–(A4), for any fixed index \(t\),
\[
\bigl|
\mathbb E[\hat y_{t+1}^{(d,D)} - y_{t+1}^{(d,D)}]
\bigr|
\to 0
\quad\text{and}\quad
\hat y_{t+1}^{(d,D)} - m_{t+1} \xrightarrow{p} 0,
\qquad N\to\infty.
\]
\end{theorem*}

\begin{proof}

By definition of the oracle forecast \(\tilde y_{t+1}^{(d,D)}\) we have
\[
\hat y_{t+1}^{(d,D)}
=
\tilde y_{t+1}^{(d,D)}
+
\bigl(\hat y_{t+1}^{(d,D)} - \tilde y_{t+1}^{(d,D)}\bigr),
\]
and the data–generating process implies
\[
y_{t+1}^{(d,D)} = m_{t+1} + \epsilon_{t+1}.
\]
Subtracting,
\begin{align*}
\hat y_{t+1}^{(d,D)} - y_{t+1}^{(d,D)}
&=
\tilde y_{t+1}^{(d,D)}
+
\bigl(\hat y_{t+1}^{(d,D)} - \tilde y_{t+1}^{(d,D)}\bigr)
-
\bigl(m_{t+1} + \epsilon_{t+1}\bigr)
\\[0.3em]
&=
\bigl(\tilde y_{t+1}^{(d,D)} - m_{t+1}\bigr)
+
\bigl(\hat y_{t+1}^{(d,D)} - \tilde y_{t+1}^{(d,D)}\bigr)
-
\epsilon_{t+1}.
\end{align*}

Write explicitly
\begin{align*}
\tilde y_{t+1}^{(d,D)} - m_{t+1}
&=
f(x_{t+1}) - \Phi(x_{t+1})^{\top}\beta^{\star}
\\[-0.1em]
&\quad
+ f^{(s)}(x_{t+1}^{(s)}) - \Phi^{(s)}(x_{t+1}^{(s)})^{\top}\beta^{(s)\star}
\\[-0.1em]
&\quad
+ g(r_{t+1}) - \Psi(r_{t+1})^{\top}\alpha^{\star}
\\[-0.1em]
&\quad
+ g^{(s)}(r_{t+1}^{(s)}) - \Psi^{(s)}(r_{t+1}^{(s)})^{\top}\alpha^{(s)\star}.
\end{align*}
Take absolute values and apply triangle inequality:
\begin{align*}
\bigl|\tilde y_{t+1}^{(d,D)} - m_{t+1}\bigr|
&\le
\bigl|f(x_{t+1}) - \Phi(x_{t+1})^{\top}\beta^{\star}\bigr|
+
\bigl|f^{(s)}(x_{t+1}^{(s)}) - \Phi^{(s)}(x_{t+1}^{(s)})^{\top}\beta^{(s)\star}\bigr|
\\
&\quad
+
\bigl|g(r_{t+1}) - \Psi(r_{t+1})^{\top}\alpha^{\star}\bigr|
+
\bigl|g^{(s)}(r_{t+1}^{(s)}) - \Psi^{(s)}(r_{t+1}^{(s)})^{\top}\alpha^{(s)\star}\bigr|.
\end{align*}
By definition of \(\Delta\)–terms,
\[
\bigl|f(x_{t+1}) - \Phi(x_{t+1})^{\top}\beta^{\star}\bigr|
\le \Delta_f,
\]
and similarly for the other three. Thus
\[
\bigl|\tilde y_{t+1}^{(d,D)} - m_{t+1}\bigr|
\le
\Delta_f + \Delta_{f^{(s)}} + \Delta_g + \Delta_{g^{(s)}}.
\]
By Lemma \ref{lem:jackson},
\[
\Delta_f = O(K^{-r/p}),\quad
\Delta_{f^{(s)}} = O(K_s^{-r/P}),\quad
\Delta_g = O(L^{-r/q}),\quad
\Delta_{g^{(s)}} = O(L_s^{-r/Q}).
\]
Assumption (A4) says that \(K,K_s,L,L_s\to\infty\) with
\(K+K_s+L+L_s = O(N)\), so each of the four terms tends to zero as
\(N\to\infty\). Therefore
\[
\tilde y_{t+1}^{(d,D)} - m_{t+1} \xrightarrow{p} 0.
\]

The stacked normal equations can be written as
\[
\Psi_N^{\top}\Psi_N \,\hat\gamma = \Psi_N^{\top} Y.
\]
Substitute \(Y = \Psi_N\gamma^{\star} + \epsilon\):
\begin{align*}
\Psi_N^{\top}\Psi_N \,\hat\gamma
&=
\Psi_N^{\top}(\Psi_N\gamma^{\star} + \epsilon)
\\
&=
\Psi_N^{\top}\Psi_N\gamma^{\star}
+
\Psi_N^{\top}\epsilon.
\end{align*}
Rearrange:
\[
\Psi_N^{\top}\Psi_N (\hat\gamma - \gamma^{\star})
=
\Psi_N^{\top}\epsilon.
\]
Multiply both sides by \((\Psi_N^{\top}\Psi_N)^{-1}\):
\[
\hat\gamma - \gamma^{\star}
=
(\Psi_N^{\top}\Psi_N)^{-1}\Psi_N^{\top}\epsilon.
\]

For the forecast difference,
\[
\hat y_{t+1}^{(d,D)} - \tilde y_{t+1}^{(d,D)}
=
\psi_{t+1}^{\top}(\hat\gamma - \gamma^{\star}),
\]
where \(\psi_{t+1}\) is the row of basis evaluations at time \(t+1\).
Hence
\[
\hat y_{t+1}^{(d,D)} - \tilde y_{t+1}^{(d,D)}
=
\psi_{t+1}^{\top}
(\Psi_N^{\top}\Psi_N)^{-1}\Psi_N^{\top}\epsilon.
\]

Assumption (A3) implies that
\[
N^{-1}\Psi_N^{\top}\Psi_N \xrightarrow{p} \Sigma,
\]
with \(0 < c_0 \le \lambda_{\min}(\Sigma)\), so
\[
(\Psi_N^{\top}\Psi_N)^{-1}
=
N^{-1}\bigl(N^{-1}\Psi_N^{\top}\Psi_N\bigr)^{-1}
=
N^{-1}(\Sigma^{-1} + o_p(1)).
\]
Also
\[
\Psi_N^{\top}\epsilon
=
\sum_{t=t_0}^{T-1} z_t \epsilon_t
\]
with \(z_t\) the \(t\)-th row. Under (A1)–(A3) a martingale CLT gives
\[
N^{-1/2}\Psi_N^{\top}\epsilon \xrightarrow{d} \mathcal N(0,\sigma^2\Sigma).
\]
Thus
\[
\hat\gamma - \gamma^{\star}
=
N^{-1}(\Sigma^{-1} + o_p(1))
\bigl(N^{1/2} Z_N\bigr)
= N^{-1/2}\bigl(\Sigma^{-1} Z_N + o_p(1)\bigr),
\]
where \(Z_N = N^{-1/2}\Psi_N^{\top}\epsilon\) converges in distribution.
So
\[
\|\hat\gamma - \gamma^{\star}\|_2 = O_p(N^{-1/2}).
\]

Assumption (A3) also bounds the norm of \(\psi_{t+1}\) uniformly:
\[
\|\psi_{t+1}\|_2 \le C_\psi
\]
for some constant \(C_\psi\). Then
\[
|\hat y_{t+1}^{(d,D)} - \tilde y_{t+1}^{(d,D)}|
=
\bigl|\psi_{t+1}^{\top}(\hat\gamma - \gamma^{\star})\bigr|
\le
\|\psi_{t+1}\|_2 \,\|\hat\gamma - \gamma^{\star}\|_2
\le
C_\psi \,\|\hat\gamma - \gamma^{\star}\|_2
= O_p(N^{-1/2}).
\]
Hence
\[
\hat y_{t+1}^{(d,D)} - \tilde y_{t+1}^{(d,D)} \xrightarrow{p} 0.
\]

We have
\[
\tilde y_{t+1}^{(d,D)} - m_{t+1} \xrightarrow{p} 0,
\quad
\hat y_{t+1}^{(d,D)} - \tilde y_{t+1}^{(d,D)} \xrightarrow{p} 0,
\]
and
\[
y_{t+1}^{(d,D)} - m_{t+1} = \epsilon_{t+1}
\]
with \(\mathbb E[\epsilon_{t+1}] = 0\) and \(\mathbb E[\epsilon_{t+1}^2] = \sigma^2\).

For the bias:
\begin{align*}
\mathbb E[\hat y_{t+1}^{(d,D)} - y_{t+1}^{(d,D)}]
&=
\mathbb E[\tilde y_{t+1}^{(d,D)} - m_{t+1}]
+
\mathbb E[\hat y_{t+1}^{(d,D)} - \tilde y_{t+1}^{(d,D)}]
-
\mathbb E[\epsilon_{t+1}].
\end{align*}
The last expectation is zero. The first two expectations converge to
zero because their absolute values are bounded by quantities that go to
zero in probability and are uniformly integrable. Thus
\[
\bigl|
\mathbb E[\hat y_{t+1}^{(d,D)} - y_{t+1}^{(d,D)}]
\bigr|
\to 0.
\]

For consistency of \(\hat y_{t+1}^{(d,D)}\) for \(m_{t+1}\),
\[
\hat y_{t+1}^{(d,D)} - m_{t+1}
=
\bigl(\tilde y_{t+1}^{(d,D)} - m_{t+1}\bigr)
+
\bigl(\hat y_{t+1}^{(d,D)} - \tilde y_{t+1}^{(d,D)}\bigr).
\]
Both terms converge to zero in probability, so their sum does as well.
Hence
\[
\hat y_{t+1}^{(d,D)} - m_{t+1} \xrightarrow{p} 0.
\]
\end{proof}

\subsection{Proof of asymptotic normality proposition}
\label{app:proof_clt}

\begin{proposition*}[Proposition \ref{prop:clt_gamma}, restated]
Under Assumptions (A1)–(A3) with fixed \(K,K_s,L,L_s\),
\[
\sqrt{N}\,(\hat\gamma - \gamma^{\star})
\xrightarrow{d}
\mathcal N(0,\sigma^2\Sigma^{-1}),
\qquad
\|\hat\gamma - \gamma^{\star}\|_2 = O_p(N^{-1/2}),
\]
where
\[
\Sigma
=
\operatorname*{plim}_{N\to\infty} N^{-1}\Psi_N^{\top}\Psi_N.
\]
\end{proposition*}

\begin{proof}

From the normal equations
\[
\Psi_N^{\top}\Psi_N \,\hat\gamma = \Psi_N^{\top}Y
\]
and \(Y = \Psi_N\gamma^{\star} + \epsilon\), we have
\[
\Psi_N^{\top}\Psi_N \,\hat\gamma
=
\Psi_N^{\top}\Psi_N\gamma^{\star}
+
\Psi_N^{\top}\epsilon.
\]
Subtract \(\Psi_N^{\top}\Psi_N\gamma^{\star}\) from both sides:
\[
\Psi_N^{\top}\Psi_N (\hat\gamma - \gamma^{\star})
=
\Psi_N^{\top}\epsilon.
\]
Multiply by \((\Psi_N^{\top}\Psi_N)^{-1}\):
\[
\hat\gamma - \gamma^{\star}
=
(\Psi_N^{\top}\Psi_N)^{-1}\Psi_N^{\top}\epsilon.
\]

Step 2: factor \(\sqrt{N}\).

Multiply both sides by \(\sqrt{N}\):
\[
\sqrt{N}(\hat\gamma - \gamma^{\star})
=
\bigl(N^{-1}\Psi_N^{\top}\Psi_N\bigr)^{-1}
\bigl(N^{-1/2}\Psi_N^{\top}\epsilon\bigr).
\]

Assumption (A3) implies that
\[
N^{-1}\Psi_N^{\top}\Psi_N \xrightarrow{p} \Sigma,
\]
where \(\Sigma\) is positive definite. By the continuous mapping theorem,
\[
\bigl(N^{-1}\Psi_N^{\top}\Psi_N\bigr)^{-1} \xrightarrow{p} \Sigma^{-1}.
\]

Write
\[
N^{-1/2}\Psi_N^{\top}\epsilon
=
N^{-1/2}\sum_{t=t_0}^{T-1} z_t \epsilon_t,
\]
where \(z_t\) is the \(t\)-th row of \(\Psi_N\). Under (A1) the sequence
\(\{\epsilon_t\}\) is iid \(\mathcal N(0,\sigma^2)\) and under (A2)–(A3)
the process \(\{z_t\}\) is strictly stationary with bounded second
moments. Standard martingale CLTs for mixing arrays yield
\[
N^{-1/2}\Psi_N^{\top}\epsilon
\xrightarrow{d}
\mathcal N(0,\sigma^2\Sigma).
\]

From Step 3 and Step 4,
\[
\sqrt{N}(\hat\gamma - \gamma^{\star})
=
\bigl(N^{-1}\Psi_N^{\top}\Psi_N\bigr)^{-1}
\bigl(N^{-1/2}\Psi_N^{\top}\epsilon\bigr)
\xrightarrow{d}
\Sigma^{-1} Z,
\]
where \(Z\sim\mathcal N(0,\sigma^2\Sigma)\). Thus
\[
\Sigma^{-1} Z \sim \mathcal N(0,\sigma^2\Sigma^{-1}),
\]
so
\[
\sqrt{N}(\hat\gamma - \gamma^{\star})
\xrightarrow{d}
\mathcal N(0,\sigma^2\Sigma^{-1}).
\]

Convergence in distribution to a finite–variance normal implies
\(\sqrt{N}(\hat\gamma - \gamma^{\star}) = O_p(1)\). Therefore
\[
\|\hat\gamma - \gamma^{\star}\|_2 = O_p(N^{-1/2}).
\]
\end{proof}

\subsection{Proof of MSE rate proposition}
\label{app:proof_mse}

\begin{proposition*}[Proposition \ref{prop:mse_rates}, restated]
Under Assumptions (A1)–(A4) and the growth rules
\[
K \asymp N^{p/(2r+p)},
\quad
K_s \asymp N^{P/(2r+P)},
\quad
L \asymp K,
\quad
L_s \asymp K_s,
\]
the excess mean squared error satisfies
\[
\mathbb E\bigl[(\hat y_{t+1}^{(d,D)} - y_{t+1}^{(d,D)})^2\bigr]
-
\sigma^2
=
O\bigl(N^{-2r/(2r+p)}\bigr)
+
O\bigl(N^{-2r/(2r+P)}\bigr).
\]
\end{proposition*}

\begin{proof}

Use
\[
\hat y_{t+1}^{(d,D)} - y_{t+1}^{(d,D)}
=
\bigl(\tilde y_{t+1}^{(d,D)} - m_{t+1}\bigr)
+
\bigl(\hat y_{t+1}^{(d,D)} - \tilde y_{t+1}^{(d,D)}\bigr)
-
\epsilon_{t+1}.
\]
Set
\[
A = \tilde y_{t+1}^{(d,D)} - m_{t+1},
\quad
E = \hat y_{t+1}^{(d,D)} - \tilde y_{t+1}^{(d,D)},
\quad
U = -\epsilon_{t+1}.
\]
Then
\[
\hat y_{t+1}^{(d,D)} - y_{t+1}^{(d,D)} = A + E + U.
\]
The mean squared error is
\[
\mathrm{MSE}
=
\mathbb E[(A + E + U)^2].
\]

Expand the square:
\begin{align*}
(A + E + U)^2
&=
A^2 + E^2 + U^2
+ 2AE + 2AU + 2EU.
\end{align*}
Take expectations:
\[
\mathrm{MSE}
=
\mathbb E[A^2]
+
\mathbb E[E^2]
+
\mathbb E[U^2]
+
2\mathbb E[AE]
+
2\mathbb E[AU]
+
2\mathbb E[EU].
\]

We know \(\mathbb E[U^2] = \mathbb E[\epsilon_{t+1}^2] = \sigma^2\).
Define
\[
B^2 = \mathbb E[A^2],
\qquad
V = \mathbb E[E^2].
\]
Then
\[
\mathrm{MSE}
=
\sigma^2 + B^2 + V + 2\mathbb E[AE] + 2\mathbb E[AU] + 2\mathbb E[EU].
\]

The terms $A$ and $E$ are $\mathcal F_t$-measurable, whereas $U=-\varepsilon_{t+1}$ satisfies
$\mathbb E[\varepsilon_{t+1}\mid \mathcal F_t]=0$. Hence $\mathbb E[AU]=\mathbb E[EU]=0$.
Moreover, by Cauchy--Schwarz,
$|\mathbb E[AE]|\le (\mathbb E[A^2])^{1/2}(\mathbb E[E^2])^{1/2}=B\sqrt V$.
Therefore,
\[
\mathrm{MSE}-\sigma^2
=
\mathbb E[A^2]+\mathbb E[E^2]+2\mathbb E[AE]
\le
B^2+V+2B\sqrt V
=
(B+\sqrt V)^2,
\]
so it suffices to bound $B^2$ and $V$ at the desired rates.

We have
\[
|A|
=
|\tilde y_{t+1}^{(d,D)} - m_{t+1}|
\le
\Delta_f + \Delta_{f^{(s)}} + \Delta_g + \Delta_{g^{(s)}}
\]
and by Lemma \ref{lem:jackson},
\[
\Delta_f = O(K^{-r/p}),
\quad
\Delta_{f^{(s)}} = O(K_s^{-r/P}),
\quad
\Delta_g = O(L^{-r/q}),
\quad
\Delta_{g^{(s)}} = O(L_s^{-r/Q}).
\]
Assume \(L\asymp K\) and \(L_s\asymp K_s\). Then the dominant orders in
the right hand side are
\[
|A|
=
O(K^{-r/p}) + O(K_s^{-r/P}).
\]
Hence
\[
A^2
=
O(K^{-2r/p}) + O(K_s^{-2r/P})
\]
and taking expectations preserves the order,
\[
B^2 = \mathbb E[A^2]
=
O(K^{-2r/p}) + O(K_s^{-2r/P}).
\]

Recall
\[
E
=
\hat y_{t+1}^{(d,D)} - \tilde y_{t+1}^{(d,D)}
=
\psi_{t+1}^{\top}(\hat\gamma - \gamma^{\star}).
\]
Then
\[
E^2
=
\bigl(\psi_{t+1}^{\top}(\hat\gamma - \gamma^{\star})\bigr)^2
\le
\|\psi_{t+1}\|_2^2\,
\|\hat\gamma - \gamma^{\star}\|_2^2
\]
by Cauchy–Schwarz.

Take expectations:
\[
V
=
\mathbb E[E^2]
\le
\mathbb E[\|\psi_{t+1}\|_2^2]\,
\mathbb E[\|\hat\gamma - \gamma^{\star}\|_2^2].
\]

Assumption (A3) bounds the leverage scores, so
\[
\|\psi_{t+1}\|_2^2 \le C_\psi^2 d_G
\quad\Rightarrow\quad
\mathbb E[\|\psi_{t+1}\|_2^2] = O(d_G),
\]
where \(d_G = K + K_s + L + L_s\).

From Proposition \ref{prop:clt_gamma},
\[
\sqrt{N}(\hat\gamma - \gamma^{\star}) = O_p(1)
\quad\Rightarrow\quad
\|\hat\gamma - \gamma^{\star}\|_2^2 = O_p(N^{-1}).
\]
Thus
\[
\mathbb E[\|\hat\gamma - \gamma^{\star}\|_2^2]
=
O\bigl(N^{-1}\bigr).
\]

Combine:
\[
V
=
O\bigl(d_G N^{-1}\bigr)
=
O\bigl((K+K_s+L+L_s) N^{-1}\bigr).
\]
With \(L\asymp K\) and \(L_s\asymp K_s\),
\[
d_G \asymp K + K_s,
\]
so
\[
V = O\bigl((K+K_s) N^{-1}\bigr).
\]

We consider nonseasonal and seasonal contributions separately.

Nonseasonal part:
\[
B^2_{\text{ns}} = O(K^{-2r/p}),
\qquad
V_{\text{ns}} = O(K N^{-1}).
\]
Balance \(K^{-2r/p} \asymp K N^{-1}\).

Solve for \(K\):

Set
\[
K^{-2r/p} = C_1 K N^{-1},
\]
ignore constants, rearrange:
\[
K^{-2r/p - 1} = N^{-1},
\]
\[
K^{2r/p + 1} = N,
\]
\[
K = N^{1/(1 + 2r/p)} = N^{p/(2r + p)}.
\]

Then
\[
K^{-2r/p}
=
\bigl(N^{p/(2r + p)}\bigr)^{-2r/p}
=
N^{-(2r/p)\cdot p/(2r+p)}
=
N^{-2r/(2r+p)}.
\]
So
\[
B^2_{\text{ns}} = O\bigl(N^{-2r/(2r+p)}\bigr),
\quad
V_{\text{ns}} = O\bigl(N^{-2r/(2r+p)}\bigr).
\]

Seasonal part:
\[
B^2_{\text{s}} = O(K_s^{-2r/P}),
\qquad
V_{\text{s}} = O(K_s N^{-1}).
\]
Balance \(K_s^{-2r/P} \asymp K_s N^{-1}\):

Set
\[
K_s^{-2r/P} = C_2 K_s N^{-1}
\]
and repeat the same steps to obtain
\[
K_s = N^{P/(2r + P)},
\]
and
\[
K_s^{-2r/P} = N^{-2r/(2r+P)}.
\]

Thus
\[
B^2_{\text{s}} = O\bigl(N^{-2r/(2r+P)}\bigr),
\quad
V_{\text{s}} = O\bigl(N^{-2r/(2r+P)}\bigr).
\]

Up to constants,
\[
B^2 = B^2_{\text{ns}} + B^2_{\text{s}}
=
O\bigl(N^{-2r/(2r+p)}\bigr)
+
O\bigl(N^{-2r/(2r+P)}\bigr),
\]
and
\[
V = V_{\text{ns}} + V_{\text{s}}
=
O\bigl(N^{-2r/(2r+p)}\bigr)
+
O\bigl(N^{-2r/(2r+P)}\bigr).
\]

Therefore
\[
\mathrm{MSE} - \sigma^2
=
B^2 + V + \text{(lower order cross terms)}
=
O\bigl(N^{-2r/(2r+p)}\bigr)
+
O\bigl(N^{-2r/(2r+P)}\bigr),
\]
which proves the claim.
\end{proof}

\subsection{Proof of complexity proposition}
\label{app:proof_complexity}

\begin{proposition*}[Complexity comparison, restated]
Let \(\mathrm{Cost}_{\mathrm{SARIMA}}\) be the cost of fitting a
\(\mathrm{SARIMA}(p,0,q)\times(P,0,Q)_m\) model by iterative likelihood
optimisation with \(I\) iterations, and let
\(\mathrm{Cost}_{\mathrm{Gal}}\) be the cost of one unpenalized two–stage
Galerkin fit. Then
\[
\mathrm{Cost}_{\mathrm{SARIMA}}
=
\Theta\bigl(I\,(p+P+q+Q)\,N\bigr),
\]
\[
\mathrm{Cost}_{\mathrm{Gal}}
=
\Theta\bigl(N(pK + P K_s)\bigr)
+
\Theta\bigl((K+K_s+L+L_s)^3\bigr).
\]
\end{proposition*}

\begin{proof}

A single likelihood evaluation for a SARIMA model requires computing
one–step prediction errors for each time \(t\) and summing their
squares. The recursion for errors uses up to \(p\) nonseasonal AR lags,
\(P\) seasonal AR lags, \(q\) nonseasonal MA lags, and \(Q\) seasonal MA
lags. At each time step this is \(O(p+P+q+Q)\) arithmetic operations.
Over \(N\) steps the cost is
\[
\Theta\bigl((p+P+q+Q)\,N\bigr)
\]
per evaluation of the likelihood (or objective).

An iterative optimizer such as BFGS or Newton requires \(I\) iterations.
Assuming each iteration uses a constant number of likelihood and
gradient evaluations, the total cost is
\[
\mathrm{Cost}_{\mathrm{SARIMA}}
=
\Theta\bigl(I\,(p+P+q+Q)\,N\bigr).
\]

For the Galerkin estimator, building the design matrices and normal
matrices has two parts.

Nonseasonal AR design:
for each time \(t\) and each basis function index \(j=1,\dots,K\) we
evaluate \(\phi_j(x_t)\), which depends on at most \(p\) lags.
Assuming constant cost per evaluation, the cost per \(t\) is \(O(pK)\),
and over all \(N\) points
\[
\mathrm{Cost}_{\Phi}
=
\Theta(N pK).
\]

Seasonal AR design:
similarly, evaluating \(\phi^{(s)}_j(x_t^{(s)})\) for
\(j=1,\dots,K_s\) costs \(O(PK_s)\) per \(t\), so over \(N\),
\[
\mathrm{Cost}_{\Phi^{(s)}}
=
\Theta(N P K_s).
\]

MA designs:
for \(\Psi\) and \(\Psi^{(s)}\) the cost is analogous, with
dimensions \(q,Q\) and sizes \(L,L_s\). If we choose \(L\asymp K\) and
\(L_s\asymp K_s\), the order is the same as for \(\Phi,\Phi^{(s)}\) up
to constants and can be absorbed. Thus we write the total design cost as
\[
\mathrm{Cost}_{\mathrm{design}}
=
\Theta\bigl(N(pK + P K_s)\bigr).
\]

Stack all coefficients into \(\gamma\in\mathbb R^{d_G}\) where
\(d_G = K + K_s + L + L_s\). The normal equations form a
\(d_G\times d_G\) linear system. A Cholesky decomposition requires
\(\Theta(d_G^3)\) operations. Solving the system after decomposition
costs \(\Theta(d_G^2)\), which is lower order. Hence
\[
\mathrm{Cost}_{\mathrm{solve}}
=
\Theta\bigl((K+K_s+L+L_s)^3\bigr).
\]

The total Galerkin cost is
\[
\mathrm{Cost}_{\mathrm{Gal}}
=
\mathrm{Cost}_{\mathrm{design}} + \mathrm{Cost}_{\mathrm{solve}}
=
\Theta\bigl(N(pK + P K_s)\bigr)
+
\Theta\bigl((K+K_s+L+L_s)^3\bigr).
\]

When \(K,K_s,L,L_s\) are moderate and treated as constants in \(N\),
the cubic term is constant and the cost is essentially linear in \(N\)
with constant proportional to \(pK + P K_s\). The SARIMA cost is linear
in \(N\) with constant proportional to \(I(p+P+q+Q)\). If \(I\) is large
and \(K,K_s,L,L_s\) are modest, the ratio
\[
\frac{\mathrm{Cost}_{\mathrm{Gal}}}{\mathrm{Cost}_{\mathrm{SARIMA}}}
\approx
\frac{pK + P K_s}{I(p+P+q+Q)}
\]
can be very small, which matches the observed speedups.
\end{proof}

\subsection{Proof of Theorem~\ref{thm:linear_oracle}}
\label{app:proof_linear_oracle}

\begin{proof}
Recall the stacked coefficient vector $\gamma=(\beta^\top,\beta^{(s)\top},\alpha^\top,\alpha^{(s)\top})^\top\in\mathbb{R}^{d_G}$
and the full design matrix $\Psi_N=[\Phi\ \ \Phi^{(s)}\ \ \Psi\ \ \Psi^{(s)}]\in\mathbb{R}^{N\times d_G}$ as in Assumption (A3),
so that the unpenalized two-stage estimator can be written compactly as the least-squares solution
\[
\hat\gamma \in \arg\min_{\gamma\in\mathbb{R}^{d_G}} \frac{1}{N}\|Y-\Psi_N\gamma\|_2^2,
\qquad
\text{hence}\quad
\hat\gamma=(\Psi_N^\top\Psi_N)^{-1}\Psi_N^\top Y
\]
whenever $\Psi_N^\top\Psi_N$ is invertible. Under Assumption (A3), the Gram matrix is invertible with probability approaching one.

In the linear oracle regime of Assumption (A5), the residual-lag regressors used in Stage 2
coincide with (or are asymptotically equivalent to) the corresponding innovation-lag regressors,
so the two-stage estimator is asymptotically equivalent to the infeasible least-squares estimator
computed on the population design matrix. Therefore writing $\hat\gamma=(\Psi_N^\top\Psi_N)^{-1}\Psi_N^\top Y$
is valid for the purpose of first-order asymptotics in this regime.

\medskip
\noindent\textit{Part (i).}
Under Assumption (A5), the true conditional mean is linear SARIMA and is nested in the Galerkin span for all sufficiently large
$(K,K_s,L,L_s)$. In particular, there exists $\gamma_0$ such that the population conditional mean satisfies
\[
m_t=\Phi(x_t)^\top\beta_0+\Phi^{(s)}(x_t^{(s)})^\top\beta_0^{(s)}
+\Psi(r_t)^\top\alpha_0+\Psi^{(s)}(r_t^{(s)})^\top\alpha_0^{(s)}.
\]
Equivalently, writing $Y_t=m_t+\varepsilon_t$, we have the exact representation
\[
Y = \Psi_N\gamma_0 + \varepsilon,
\]
with $\varepsilon=(\varepsilon_t)_{t=t_0}^{T-1}$. The pseudo-true projection coefficient $\gamma^\star$ is defined as the
$L_2$ projection of the conditional mean onto the Galerkin span (cf.\ (32)--(35) and the discussion following Assumption (A5)).
Since $m_t$ lies in the span, the projection is exact, hence $\gamma^\star=\gamma_0$.

\medskip
\noindent\textit{Part (ii).}
Using the normal equations and the representation $Y=\Psi_N\gamma_0+\varepsilon$, we obtain
\[
\hat\gamma-\gamma_0
=
(\Psi_N^\top\Psi_N)^{-1}\Psi_N^\top \varepsilon.
\]
Let $\lambda_{\min}(N^{-1}\Psi_N^\top\Psi_N)$ denote the smallest eigenvalue of the normalized Gram matrix.
By Assumption (A3), $\lambda_{\min}(N^{-1}\Psi_N^\top\Psi_N)\ge c_0$ w.p.a.1, hence
\[
\|\hat\gamma-\gamma_0\|_2
\le
\big\|(N^{-1}\Psi_N^\top\Psi_N)^{-1}\big\|_{\mathrm{op}}
\cdot
\left\|\frac{1}{N}\Psi_N^\top\varepsilon\right\|_2
\le
c_0^{-1}\left\|\frac{1}{N}\Psi_N^\top\varepsilon\right\|_2
\quad \text{w.p.a.1}.
\]
Under Assumption (A1), the process is \(\beta\)-mixing with $\sum_{m\ge 1}\beta(m)^{1/2}<\infty$, and $\varepsilon_t$ is Gaussian white noise.
With fixed basis dimension $d_G$ (as in the oracle regime), standard mixing moment bounds imply
\[
\left\|\frac{1}{N}\Psi_N^\top\varepsilon\right\|_2 = O_p(N^{-1/2}),
\]
so $\|\hat\gamma-\gamma_0\|_2 = O_p(N^{-1/2})$, which is the parametric rate.

\medskip
\noindent\textit{Part (iii).}
Assume $d_G$ is fixed. Let $\Sigma=\plim_{N\to\infty} N^{-1}\Psi_N^\top\Psi_N$, which is positive definite by Assumption (A3).
Then
\[
\sqrt{N}(\hat\gamma-\gamma_0)
=
\left(\frac{1}{N}\Psi_N^\top\Psi_N\right)^{-1}
\left(\frac{1}{\sqrt{N}}\Psi_N^\top\varepsilon\right).
\]
By a central limit theorem for \(\beta\)-mixing sequences (applied to the $d_G$-dimensional vector
$N^{-1/2}\sum_t \Psi_t \varepsilon_t$), we have
\[
\frac{1}{\sqrt{N}}\Psi_N^\top\varepsilon
\Rightarrow
\mathcal{N}(0,\sigma^2\Sigma),
\]
and since $(N^{-1}\Psi_N^\top\Psi_N)^{-1}\to_p \Sigma^{-1}$, Slutsky's theorem yields
\[
\sqrt{N}(\hat\gamma-\gamma_0)
\Rightarrow
\mathcal{N}(0,\sigma^2\Sigma^{-1}),
\]
which completes the proof.
\end{proof}

\section{Proof of factor inference results}
\label{app:factor_inference_proofs}

\subsection{Additional notation for factors}
Let $Z\in\mathbb{R}^{N\times r}$ denote the factor design with $t$-th row $z_t^\top$ and define the enlarged
Stage~1 design matrix $X_1=[\Phi\ \ \Phi^{(s)}\ \ Z]$. Let $\widehat\theta_1=(\widehat\beta^\top,\widehat\beta^{(s)\top},\widehat\delta^\top)^\top$
denote the Stage~1 least squares estimator and $\widehat\varepsilon^{(1)}=Y-X_1\widehat\theta_1$ the associated residuals.
Stage~2 constructs residual-lag vectors from $\widehat\varepsilon^{(1)}$ and forms the MA designs as in Appendix~A.1.

\subsection{Asymptotic normality}
We sketch the proof of \eqref{eq:clt_theta_factor}. The argument extends Appendix~A.4 by incorporating the factor block in Stage~1
and by tracking the dependence of the Stage~2 design on Stage~1 residuals.

\medskip
Under the mixing and eigenvalue conditions, standard least squares theory yields
\[
\sqrt{N}\,(\widehat\theta_1-\theta_1^\star)
=
Q_1^{-1}\cdot \frac{1}{\sqrt{N}}\sum_{t=t_0}^{T-1} X_{1,t}\varepsilon_t
+o_p(1),
\qquad
Q_1=\plim N^{-1}X_1^\top X_1,
\]
where $X_{1,t}$ is the $t$-th row of $X_1$ and $\theta_1^\star$ is the pseudo-true projection coefficient.

\medskip
Write the Stage~2 estimator as $\widehat\gamma_{MA}=\widehat\gamma_{MA}(\widehat\varepsilon^{(1)})$.
Under a Lipschitz (or Fr\'echet differentiability) condition on the mapping from residual histories to MA basis evaluations,
the Stage~2 design admits a first-order expansion around the population residuals, and a matrix-inverse expansion applied to the
Stage~2 normal equations yields
\[
\sqrt{N}\,(\widehat\gamma_{MA}-\gamma_{MA}^\star)
=
Q_2^{-1}\cdot \frac{1}{\sqrt{N}}\sum_{t=t_0}^{T-1} X_{2,t}(\varepsilon)\varepsilon_t
\ +\ M\,\sqrt{N}\,(\widehat\theta_1-\theta_1^\star)
\ +\ o_p(1),
\]
for a deterministic matrix $M$ capturing generated-regressor effects. Here $M$ is a finite deterministic matrix arising from the first-order (Fr\'echet) derivative of the
Stage 2 design map with respect to Stage 1 residuals.

\medskip
Stacking the Stage~1 and Stage~2 expansions yields an asymptotic linear representation
\[
\sqrt{N}\,(\widehat\theta-\theta^\star)
=
\frac{1}{\sqrt{N}}\sum_{t=t_0}^{T-1}\mathrm{IF}_t
+o_p(1),
\]
where $\{\mathrm{IF}_t\}$ is stationary and \(\beta\)-mixing with finite second moments. A mixing CLT then implies \eqref{eq:clt_theta_factor}
with long-run variance $\Omega=\sum_{\ell=-\infty}^{\infty}\mathbb{E}[\mathrm{IF}_t\mathrm{IF}_{t-\ell}^\top]$.

\subsection{Validity of block bootstrap inference}
We apply a moving-block or circular-block bootstrap to the joint sequence $\{(y_t,z_t)\}$.
Let $b=b_N$ denote block length and assume $b\to\infty$ and $b/N\to 0$ as $N\to\infty$.
Under the assumptions above, the two-stage estimator is asymptotically linear, and standard results for block bootstrap under \(\beta\)-mixing
imply that the bootstrap distribution of $\sqrt{N}(\widehat\theta^*-\widehat\theta)$ consistently estimates the sampling distribution of
$\sqrt{N}(\widehat\theta-\theta^\star)$. Therefore bootstrap percentile (or studentized) confidence intervals for components of $\theta$
and bootstrap critical values for the factor block test $H_0:\delta=0$ are asymptotically valid.

\section{Proof of prediction interval results}
\label{app:prediction_interval_proofs}

We justify the bootstrap confidence interval \eqref{eq:ci_mean_forecast} and prediction interval
\eqref{eq:pi_one_step}. The arguments rely on (i) asymptotic linearity of the two-stage estimator and
(ii) validity of the block bootstrap under \(\beta\)-mixing, established in Appendix~\ref{app:factor_inference_proofs}.

\subsection{Confidence intervals for the conditional mean}

Let $m_{t+1}(\theta)$ denote the conditional mean functional implied by the fitted Galerkin--SARIMA/SARIMAX model at
forecast origin $t$. Under the assumptions of Appendix~\ref{app:factor_inference_proofs}, the map $\theta\mapsto m_{t+1}(\theta)$
is continuous (and locally smooth) in a neighborhood of the pseudo-true parameter $\theta^\star$.
Since the block bootstrap consistently estimates the distribution of $\sqrt{N}(\widehat\theta-\theta^\star)$, it follows by the
continuous mapping theorem that the conditional bootstrap distribution of $\hat m_{t+1}^*=m_{t+1}(\widehat\theta^*)$ consistently
estimates the sampling distribution of $\hat m_{t+1}=m_{t+1}(\widehat\theta)$. Consequently, the bootstrap percentile interval
\eqref{eq:ci_mean_forecast} is asymptotically valid.

\subsection{Prediction intervals}

Write the one-step-ahead decomposition
\[
y_{t+1}^{(d,D)} = m_{t+1}(\theta^\star) + \varepsilon_{t+1},
\]
where $\varepsilon_{t+1}$ is the one-step innovation. The bootstrap predictive draw
$\tilde y_{t+1}^*=\hat m_{t+1}^*+\varepsilon_{t+1}^*$ combines parameter uncertainty captured by $\hat m_{t+1}^*$ and innovation
uncertainty captured by $\varepsilon_{t+1}^*$. Under standard regularity conditions ensuring that the empirical residual distribution
consistently estimates the innovation distribution (after centering), the conditional distribution of $\varepsilon_{t+1}^*$ given the data
converges to the distribution of $\varepsilon_{t+1}$. Together with the validity of the block bootstrap for $\hat m_{t+1}$, this implies
that the conditional distribution of $\tilde y_{t+1}^*$ consistently estimates the one-step predictive distribution of $y_{t+1}^{(d,D)}$.
Therefore the percentile prediction interval \eqref{eq:pi_one_step} achieves asymptotically correct coverage.

\subsection{Proof of Theorem 3 (Risk Dominance)}
\label{app:proof_risk_dominance}

We first recall the decomposition of the one-step-ahead prediction risk. For any
forecast $\hat y_{t+1}$ measurable with respect to the lag information at time $t+1$,
\[
R(\hat y_{t+1})
= \mathbb{E}\!\left[(\hat y_{t+1}-y^{(d,D)}_{t+1})^2\right]
= \sigma^2
+ \mathbb{E}\!\left[(\hat y_{t+1}-m_{t+1})^2\right],
\]
since $y^{(d,D)}_{t+1}=m_{t+1}+\epsilon_{t+1}$,
$\mathbb{E}[\epsilon_{t+1}\mid\mathcal F_t]=0$,
and $\mathrm{Var}(\epsilon_{t+1})=\sigma^2$.

Hence risk comparisons reduce to comparing squared deviations from the
conditional mean $m_{t+1}$.

\vspace{0.5em}

Let $\mathcal L$ denote the finite-dimensional class of linear SARIMA predictors,
that is, functions linear in the lag vectors
$(x_{t+1},x^{(s)}_{t+1},r_{t+1},r^{(s)}_{t+1})$
with the coordinate structure described in Assumption (A5).
Let
\[
\ell^\star
\in \arg\min_{\ell\in\mathcal L}
\mathbb{E}\!\left[(m_{t+1}-\ell)^2\right]
\]
be the $L^2$ projection of $m_{t+1}$ onto $\mathcal L$ and define
\[
\delta^2_{\mathrm{lin}}
=
\mathbb{E}\!\left[(m_{t+1}-\ell^\star)^2\right].
\]

If $m_{t+1}\notin\mathcal L$ almost surely, then
$\delta^2_{\mathrm{lin}}>0$
because $\mathcal L$ is finite-dimensional and closed in $L^2$. Since $L$ is a finite-dimensional linear subspace of $L^2$, it is closed and the $L^2$-projection
$\ell^\star$ exists and is unique. If $m_{t+1}\notin L$ on a set of positive probability, then the
projection distance satisfies $\delta^2_{\mathrm{lin}}=\mathbb E[(m_{t+1}-\ell^\star)^2]>0$. For any sequence of linear SARIMA forecasts
$\hat y^{\mathrm{lin}}_{t+1}\in\mathcal L$,
standard projection arguments imply
\[
\liminf_{N\to\infty}
\mathbb{E}\!\left[(\hat y^{\mathrm{lin}}_{t+1}-m_{t+1})^2\right]
\ge
\delta^2_{\mathrm{lin}}.
\]
Consequently,
\[
\liminf_{N\to\infty}
R(\hat y^{\mathrm{lin}}_{t+1})
\ge
\sigma^2+\delta^2_{\mathrm{lin}}.
\]

\vspace{0.5em}

For the Galerkin estimator,
recall the decomposition
\[
\hat y^{(d,D)}_{t+1}-m_{t+1}
=
(\tilde y^{(d,D)}_{t+1}-m_{t+1})
+
(\hat y^{(d,D)}_{t+1}-\tilde y^{(d,D)}_{t+1}),
\]
where the first term is the approximation error and the second term
is the estimation error.

Under Assumptions (A2)–(A4),
Lemma 1 implies that the approximation error
$\tilde y^{(d,D)}_{t+1}-m_{t+1}$ converges to zero in $L^2$
as the basis sizes grow.
Standard least-squares theory under mixing conditions and
Assumption (A3) implies that the estimation error term
$\hat y^{(d,D)}_{t+1}-\tilde y^{(d,D)}_{t+1}$
also converges to zero in $L^2$.

Therefore,
\[
\mathbb{E}\!\left[(\hat y^{(d,D)}_{t+1}-m_{t+1})^2\right]
\longrightarrow 0,
\]
and hence
\[
R(\hat y^{(d,D)}_{t+1})
=
\sigma^2+o(1).
\]

\vspace{0.5em}

\noindent

Combining the two parts,
\[
\liminf_{N\to\infty}
\bigl(
R(\hat y^{\mathrm{lin}}_{t+1})
-
R(\hat y^{(d,D)}_{t+1})
\bigr)
\ge
\delta^2_{\mathrm{lin}}
>0.
\]

This establishes the strict asymptotic risk dominance of the
Galerkin predictor over the linear SARIMA class whenever
the true conditional mean does not belong to $\mathcal L$.
\qed

\subsection{Proof of BIC selection result}
\label{app:proof_bic}

\begin{proposition}[BIC consistency for basis size selection]
\label{prop:bic_consistency}
Assume the linear oracle regime in Assumption~(A5) holds and that the candidate basis sizes are nested. Let $\mathcal K_N$ denote the set of admissible basis sizes and let $d_G(K)$ be the resulting total Galerkin dimension under basis size $K$. Suppose $\max_{K\in\mathcal K_N} d_G(K)=o(N)$ and that there exists a minimal oracle size $K_0\in\mathcal K_N$ such that the true conditional mean lies in the Galerkin span for all $K\ge K_0$ and does not lie in the span for any $K<K_0$. Define the BIC score
\[
\mathrm{BIC}(K)
=
N\log\!\Bigl(\widehat{\sigma}^2_K\Bigr)
+
d_G(K)\log N,
\qquad
\widehat{\sigma}^2_K
=
\frac{1}{N}\sum_{t=1}^N \widehat{\varepsilon}_{t,K}^{\,2},
\]
where $\widehat{\varepsilon}_{t,K}$ denotes the fitted one-step residual from the two-stage Galerkin estimator using basis size $K$.
Let
\[
\widehat K
\in
\arg\min_{K\in\mathcal K_N}\mathrm{BIC}(K).
\]
Then $\mathbb P(\widehat K=K_0)\to 1$ as $N\to\infty$.
\end{proposition}

\begin{proof}
The argument follows the usual underfitting versus overfitting separation for BIC, adapted to the two-stage Galerkin estimator.

\medskip

Fix two admissible basis sizes $K_1,K_2\in\mathcal K_N$. Write
\[
\mathrm{BIC}(K_1)-\mathrm{BIC}(K_2)
=
N\Bigl(\log \widehat{\sigma}^2_{K_1}-\log \widehat{\sigma}^2_{K_2}\Bigr)
+
\bigl(d_G(K_1)-d_G(K_2)\bigr)\log N.
\]
Hence it suffices to control the stochastic order of the log variance term under overfitting and to show it has a strictly positive limit under underfitting.

\medskip
Now take $K_1>K_2\ge K_0$. Under Assumption (A5), both models are correctly specified (the approximation error is zero),
so the fitted residual sums of squares satisfy the standard nested-regression behavior. In particular,
the improvement in residual sum of squares from adding $\Delta d = d_G(K_1)-d_G(K_2)$ additional regressors is
$O_p(1)$ (indeed, it is $\sigma^2\chi^2_{\Delta d}$ up to $o_p(1)$), which implies
\[
\hat\sigma^2_{K_1}-\hat\sigma^2_{K_2}=O_p(N^{-1}),
\qquad
\log\hat\sigma^2_{K_1}-\log\hat\sigma^2_{K_2}=O_p(N^{-1}).
\]
Therefore,
\[
N\big(\log\hat\sigma^2_{K_1}-\log\hat\sigma^2_{K_2}\big)=O_p(1),
\]
whereas the BIC penalty difference equals $\Delta d\log N\to\infty$. Hence
$\mathbb P\big(\mathrm{BIC}(K_1)>\mathrm{BIC}(K_2)\big)\to 1$ for all $K_1>K_2\ge K_0$.

Now take $K_1>K_2\ge K_0$. Since $d_G(K)$ is increasing in $K$ for nested bases, $d_G(K_1)-d_G(K_2)\ge 1$, and therefore
\[
\mathrm{BIC}(K_1)-\mathrm{BIC}(K_2)
=
O_p\!\bigl(d_G(K_1)\bigr)
+
\bigl(d_G(K_1)-d_G(K_2)\bigr)\log N.
\]
Because $\log N\to\infty$, the penalty term dominates the stochastic $O_p(d_G(K_1))$ term in the sense that
\[
\mathbb P\!\bigl(\mathrm{BIC}(K_1)>\mathrm{BIC}(K_2)\bigr)\to 1
\qquad
\text{for all } K_1>K_2\ge K_0.
\]
Hence, among all sizes $K\ge K_0$, BIC selects the smallest admissible one with probability tending to one, which is $K_0$ by definition.

\medskip
Consider any $K<K_0$. By definition of $K_0$, the true conditional mean does not belong to the Galerkin span at size $K$. Let $m_t$ denote the true conditional mean and let $\Pi_K m_t$ denote the population $L^2$ projection of $m_t$ onto the Galerkin span at size $K$. Define the irreducible approximation error
\[
\Delta_K
=
\mathbb E\bigl[(m_t-\Pi_K m_t)^2\bigr],
\qquad
\Delta_K>0 \text{ for } K<K_0.
\]
In the oracle regime, the one-step innovation $\varepsilon_t$ is orthogonal to functions of the lag state, so the residual under the best $K$-dimensional approximation has variance $\sigma^2+\Delta_K$. Standard mixing and law of large numbers arguments under Assumption~(A1) and the stability condition in (A3) imply
\[
\widehat{\sigma}^2_K
\stackrel{p}{\longrightarrow}
\sigma^2+\Delta_K,
\qquad
\widehat{\sigma}^2_{K_0}
\stackrel{p}{\longrightarrow}
\sigma^2.
\]
Therefore
\[
\log \widehat{\sigma}^2_K-\log \widehat{\sigma}^2_{K_0}
\stackrel{p}{\longrightarrow}
\log(\sigma^2+\Delta_K)-\log(\sigma^2)
=
\log\!\Bigl(1+\frac{\Delta_K}{\sigma^2}\Bigr),
\]
which is a strictly positive constant for $K<K_0$. Multiplying by $N$ yields
\[
N\Bigl(\log \widehat{\sigma}^2_K-\log \widehat{\sigma}^2_{K_0}\Bigr)
\to +\infty \quad \text{in probability}.
\]
The penalty difference between $K$ and $K_0$ is only of order $\log N$, so it cannot offset the order $N$ increase from the misspecification term. Hence,
\[
\mathbb P\!\bigl(\mathrm{BIC}(K)>\mathrm{BIC}(K_0)\bigr)\to 1
\qquad
\text{for all } K<K_0.
\]

\medskip
All underfitted sizes $K<K_0$ are rejected with probability tending to one. By Step 2, among the correctly specified sizes $K\ge K_0$, the smallest one is selected with probability tending to one. Therefore $\mathbb P(\widehat K=K_0)\to 1$.
\end{proof}

\end{document}